\documentclass{article}

\PassOptionsToPackage{numbers,sort}{natbib}

\usepackage[preprint]{neurips_data_2024}

\usepackage[utf8]{inputenc} %
\usepackage[T1]{fontenc}    %
\usepackage[colorlinks,citecolor=linkColor]{hyperref}       %
\usepackage{url}            %
\usepackage{booktabs}       %
\usepackage{amsfonts}       %
\usepackage{nicefrac}       %
\usepackage{microtype}      %
\usepackage{xcolor}         %
\usepackage{booktabs}
\usepackage{utfsym}
\usepackage{amsmath} 

\usepackage{wrapfig}
\usepackage{colortbl}
\usepackage{multirow}
\usepackage{caption}
\usepackage{tcolorbox}
\usepackage[export]{adjustbox}
\usepackage{longtable}

\usepackage{CJKutf8} %
\usepackage{pifont}
\usepackage{algorithm}
\usepackage{algorithmic}
\usepackage{subcaption}
\usepackage{enumerate}
\usepackage{enumitem}
\usepackage{graphicx}
\usepackage{float}

\usepackage{marvosym} %

\definecolor{linkColor}{rgb}{0.18,0.39,0.62}

\definecolor{mygray}{gray}{.92}

\definecolor{mygreen}{HTML}{3cb44b}
\newcommand{\Xmat}[0]{{{\bf X}}}
\newcommand{\Ymat}[0]{{{\bf Y}}}

\newcommand{\Vmat}[0]{{{\bf V}}}
\newcommand{\PredSty}[1]{\textnormal{\ttfamily\color{mygreen!90!black}#1}\unskip}

\definecolor{gray9}{gray}{.9}
\definecolor{gray95}{gray}{.95}

\definecolor{gray8}{gray}{.8}
\definecolor{gray85}{gray}{.85}
\def\ie{\emph{i.e.}}
\def\eg{\emph{e.g.}}

\newcommand\blfootnote[1]{%
\begingroup
\renewcommand\thefootnote{}\footnote{#1}%
\addtocounter{footnote}{-1}%
\endgroup
}

\title{{\dsname}: A Unified Multimodal Corpus \\ of 10 Billion-Level Images Interleaved with Text}

\author{
\textbf{
    Qingyun Li$^{2,1*}$,
    Zhe Chen$^{3,1*}$,
    Weiyun Wang$^{4,1*}$,
    Wenhai Wang$^{5,1*}$,
    Shenglong Ye$^{1*}$,
}
\\
\textbf{
    Zhenjiang Jin$^{1*}$,
    Guanzhou Chen$^{1*}$,
    Yinan He$^{1*}$,
    Zhangwei Gao$^{1*}$,
    Erfei Cui$^{1*}$,
}
\\
\textbf{
    Jiashuo Yu$^{1*}$,
    Hao Tian$^{6*}$,
    Jiasheng Zhou$^{6*}$,
    Chao Xu$^{1*}$,
    Bin Wang$^{1*}$,
    Xingjian Wei$^{1*}$,
}
\\
\textbf{
    Wei Li$^{1*}$,
    Wenjian Zhang$^{1*}$,
    Bo Zhang$^{1*}$,
    Pinlong Cai$^{1*}$,
    Licheng Wen$^{1*}$,
    Xiangchao Yan$^{1*}$,
}
\\
\textbf{
    Zhenxiang Li$^{1*}$,
    Pei Chu$^{1*}$,
    Yi Wang$^{1*}$,
    Min Dou$^{1}$,
    Changyao Tian$^{5,1}$,
    Xizhou Zhu$^{6,1,7}$,
}
\\
\textbf{
    Lewei Lu$^{6}$,
    Yushi Chen$^{2}$,
    Junjun He$^{1}$,
    Zhongying Tu$^{1*}$,
    Tong Lu$^{3}$,
    Yali Wang$^{1}$,
}
\\
\textbf{
    Limin Wang$^{3,1}$,
    Dahua Lin$^{1}$,
    Yu Qiao$^{1}$,
    Botian Shi$^{1}$,
    Conghui He$^{1}$\textsuperscript{\Letter},
    Jifeng Dai$^{7,1}$\textsuperscript{\Letter}
}
\\
\\
$^1$Shanghai AI Laboratory,
$^2$Harbin Institute of Technology,
$^3$Nanjing University,
\\
$^4$Fudan University,
$^5$The Chinese University of Hong Kong,
\\
$^6$SenseTime Research,
$^7$Tsinghua University,
}

\begin{document}

\def\dsname{OmniCorpus}  %

\def\subsetnamecc{OmniCorpus-CC}
\def\subsetnamevideo{OmniCorpus-YT}
\def\subsetnamecn{OmniCorpus-CW}

\def\onenumspace{\phantom{0}}
\def\twonumspace{\phantom{00}}
\def\threenumspace{\phantom{000}}

\maketitle
\blfootnote{{*} Equal contribution; {\Letter} Corresponding Authors: daijifeng@tsinghua.edu.cn; heconghui@pjlab.org.cn}

\begin{abstract}

  Image-text interleaved data, consisting of multiple images and texts arranged in a natural document format, aligns with the presentation paradigm of internet data and closely resembles human reading habits.
  Recent studies have shown that such data aids multimodal in-context learning and maintains the capabilities of large language models during multimodal fine-tuning.
  However, the limited scale and diversity of current image-text interleaved data restrict the development of multimodal large language models.
  In this paper, we introduce \dsname, a 10 billion-level image-text interleaved dataset. Using an efficient data engine, we filter and extract large-scale high-quality documents, which contain 8.6 billion images and 1,696 billion text tokens. Compared to counterparts (\eg, MMC4, OBELICS), our dataset 1) has 15 times larger scales while maintaining good data quality; 2) features more diverse sources, including both English and non-English websites as well as video-centric websites; 3) is more flexible, easily degradable from an image-text interleaved format to pure text corpus and image-text pairs.
  Through comprehensive analysis and experiments, we validate the quality, usability, and effectiveness of the proposed dataset. We hope this could provide a solid data foundation for future multimodal model research.
  Code and data are released at \url{https://github.com/OpenGVLab/OmniCorpus}.

\end{abstract}

\section{Introduction}

With the rise of large language models (LLMs)~\cite{zheng2023vicuna, 2023internlm, cai2024internlm2, bai2023qwen, touvron2023llama, touvron2023llama2, bi2024deepseekllm, brown2020gpt3, openai2023gpt4, zeng2022glm}, multimodal large language models (MLLMs)~\cite{gpt4v, liu2023llava, liu2023llava1_5, chen2023internvl, chen2024internvl1_5, bai2023qwenvl, team2023gemini, reid2024gemini1_5, zhu2023minigpt-4, alayrac2022flamingo, sun2023emu1, ge2024convllava} have also made significant progress. These MLLMs typically integrate pre-trained LLMs with vision foundation models (VFMs)~\cite{radford2021clip, openclip, chen2023internvl, zhai2023siglip, sun2023evaclip}, aligning them through extensive image-text pairing datasets (\eg, LAION~\cite{schuhmann2022laion5b} and COYO~\cite{kakaobrain2022coyo700m}), thereby enabling the comprehension of visual cues within language models. 
These datasets, collected by web scraping to match images with their descriptive captions, establish robust links between visual and linguistic elements.
Nonetheless, they neglect the original structure of documents, leading to a loss of contextual details and resulting in lower text quality and lack of contextual richness compared to the training corpus of LLMs.
Therefore, there is an imperative need to \emph{investigate more natural and flexible multimodal data that go beyond naive image-text pairings, with the aim of enhancing the training efficacy of MLLMs.}

Pioneering studies~\cite{zhu2024mmc4, laurenccon2024obelics, mckinzie2024mm1, alayrac2022flamingo} have introduced image-text interleaved data, demonstrating their promise in preserving the linguistic prowess of LLMs and boosting few-shot capabilities in tasks such as image captioning and visual question answering (VQA). Despite this progress, the scale of these datasets remains relatively limited, with the most extensive containing approximately 140 million documents, significantly smaller than well-established text or image-text pair datasets. Moreover, their primary data sources, mostly English websites from Common Crawl (CC)~\cite{commoncrawl}, restrict content variety.
These constraints hinder the datasets' capacity to fully unleash the potential of MLLMs, restricting their advancement and performance.

Given these considerations, constructing large-scale high-quality image-text interleaved data for MLLMs involves addressing several key challenges: (1) \emph{Diverse data sources:} existing sources like CC are relatively homogeneous, which are mainly text-centric with few images. In addition, the availability of CC images is nearing exhaustion, making it difficult to support the scaling up of future multimodal models. (2) \emph{Large-scale data processing:} An efficient, scalable, and parallelizable data engine is required to handle the massive volumes of multimodal data involved in this task. (3) \emph{High-quality multimodal data:} Comprehensive image and text filters are also crucial to ensure that the generated text corpus maintains the same high quality as the original training data of LLMs while interleaving high-quality images.

In this work, to establish a solid data foundation for MLLM research, we introduce \dsname, 
a 10 billion-level image-text interleaved dataset.
To expand data sources and address the exhaustion of CC images, we supplement our dataset with data from non-English websites and high-quality image content from video platforms.
We propose a unified data format, termed streaming data format, which is not only flexible to store image and text data from different sources, but also facilitates subsequent data reading, visualization, and data cleaning.
To efficiently leverage the large-scale data from multiple sources, we develop \emph{an efficient data pipeline capable of scaling to thousands of CPU cores}. We carefully review the overall pipeline of the data engine and optimize each component (\eg, main body extraction, preliminary text filtering) for higher efficiency and speedup ratio in a parallel framework.
To enhance data quality, we implement a \emph{human-feedback text filter} to reduce the noise within the texts, such as advertisements and other irrelevant content.

\begin{wrapfigure}{r}{0.6\textwidth}
    \centering
    \includegraphics[width=0.6\textwidth]{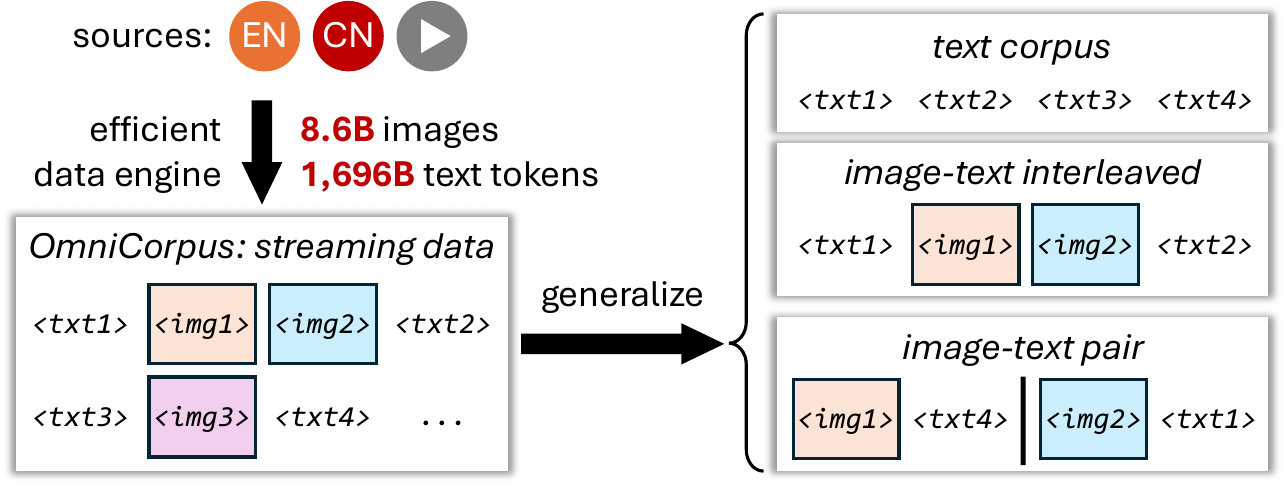}
    \caption{\textbf{Overview of our \dsname\ dataset.} It comprises 8.6 billion images and 1,696 billion text tokens sourced from diverse origins.
    Additionally, our efficient data engine generalizes the data into various formats, such as text corpus, image-text interleaved, and image-text pairs.
    }
    \label{fig:moti}
\end{wrapfigure}

As shown in Figure~\ref{fig:moti} and Table~\ref{tab:dataset_statistics}, our \dsname\ dataset demonstrates several advantages over its counterparts: 
(1) \emph{Larger data scale:} Our dataset stands as the largest multimodal dataset to date, containing 8.6 billion images, 1,696 billion text tokens, and 2.2 billion documents. It is 1.7 times larger in images and 12.5 times larger in texts compared to the previously largest multimodal dataset, LAION-5B~\cite{schuhmann2022laion5b}, while maintaining excellent data quality.
(2) \emph{Richer data diversity}: Drawing from a broader range of data sources, our dataset is more diverse than other image-text interleaved datasets. It includes bilingual multimodal data in both Chinese and English, and encompasses text-centric and vision-centric documents extracted from common websites and video platforms.
(3) \emph{More flexible format}: The streaming data format of our dataset offers exceptional flexibility, allowing adaptation to various data structures, including pure text corpora, image-text pairs, and interleaved data formats.

In summary, our contributions are threefold:

(1) We introduce the \dsname\ dataset, the largest multimodal dataset to date, which pushes the boundaries of scale and diversity by encompassing 8.6 billion images interleaved with 1,696 text tokens from diverse sources, significantly surpassing previous datasets.

(2) We propose a comprehensive set of tools and algorithms, including a streaming data format that unifies multimodal data from various sources, an efficient and scalable data engine capable of processing large-scale data, and human feedback filters to ensure high-quality data.

(3) Through extensive experiments, we validate the quality and effectiveness of our dataset. 
We show that image-text interleaved data enhances few-shot capabilities and maintains the language abilities of multimodal models. Additionally, we also gained some new findings that differ from prior findings.

\begin{table}[t]
\centering
\caption{
    \textbf{Comparison with large-scale image-text pre-training datasets}. 
    ``\#Avg.'' denotes ``\#Images per sample | \#Tokens per sample''. 
    The concept of ``\#Docs'' applies only to interleaved image-text datasets and is not relevant to paired image-text datasets.
    The proposed {\dsname} dataset features a significantly larger scale and a broader range of sources compared to previous image-text datasets.
}
\vspace{3px}
\setlength\tabcolsep{7.6pt}
\renewcommand{\arraystretch}{0.92}
\label{tab:dataset_statistics}
\scriptsize
\begin{tabular}{l|cccc|cc}
\toprule
Dataset                                                  & \#Images             & \#Tokens             & \#Docs               & \#Avg.                          & Language              & Source                \\ \midrule
\textcolor{gray}{\emph{Image-text Paired Datasets}}      & \multicolumn{1}{l}{} & \multicolumn{1}{l}{} & \multicolumn{1}{l}{} & \multicolumn{1}{l|}{}           & \multicolumn{1}{l}{}  & \multicolumn{1}{l}{}   \\
COYO-700M~\cite{kakaobrain2022coyo700m}                  & 747M                 & 12.9B                & $-$                    & \phantom{0}1 | 17               & English               & Common Crawl          \\
LAION-5B~\cite{schuhmann2022laion5b}                     & 5B                   & 135B                 & $-$                    & \phantom{0}1 | 27               & multilingual          & Common Crawl          \\ \midrule
\textcolor{gray}{\emph{Image-text Interleaved Datasets}} & \multicolumn{1}{l}{} & \multicolumn{1}{l}{} & \multicolumn{1}{l}{} & \multicolumn{1}{l|}{}           & \multicolumn{1}{l}{}  & \multicolumn{1}{l}{}   \\
KOSMOS-1 data~\cite{huang2023kosmos-1}                   & $-$                    & $-$                    & 71M                  & $-$ | $-$                           & English               & Common Crawl          \\
M3W (Flamingo)~\cite{alayrac2022flamingo}                & 185M                 & $-$                    & 43M                  & \phantom{-}4.3 | $-$\phantom{...} & English               & English Websites      \\
Web Interleaved (MM1)~\cite{mckinzie2024mm1}             & 1B                   & 500B                 & 500M                 & \phantom{K}2 | 1K               & English               & English Websites      \\
MMC4-Core~\cite{zhu2024mmc4}                             & 29.9M                & 2.4B                 & 7.3M                 & \phantom{0}4.1 | 329\phantom{.} & English               & Common Crawl          \\
MMC4~\cite{zhu2024mmc4}                                  & 585M                 & 43B                  & 103M                 & \phantom{0}5.7 | 417\phantom{.} & English               & Common Crawl          \\
OBELICS~\cite{laurenccon2024obelics}                     & 353M                 & 115B                 & 141M                 & \phantom{0}2.5 | 816\phantom{.} & English               & Common Crawl          \\ 
{\subsetnamevideo} (ours)                                       & 2.1B                 & 7.7B                 & 10M                  & 210 | 770                       & English               & YouTube Videos (YT)   \\
{\subsetnamecn} (ours)                                         & 3.2B                 & 940B               & 1196M                 & \twonumspace2 | 330               & Chinese               & Chinese Websites (CW) \\
{\subsetnamecc}  (ours)                                        & 3.3B                 & 748B                 & 988M                 & \phantom{0}3.3 | 757\phantom{.}                       & English               & Common Crawl (CC)     \\
\rowcolor[HTML]{EBEBEB} 
\textbf{{\dsname}} (ours)                                      & \textbf{8.6B}        & \textbf{1696B}     & \textbf{2.2B}        & \textbf{\phantom{0}3.9 | 574\phantom{.}}               & \textbf{Bilingual} & \textbf{CC, CW, YT}     \\ \bottomrule
\end{tabular}%
\vspace{-8px}
\end{table}

\section{Related Works}

\subsection{Image-Text Datasets}

As one of the three pillars of deep learning, datasets play a critical role in advancing deep learning models, especially in vision-language models (VLMs).
Prior to the era of large-scale models, image-text datasets~\citep{chen2015coco-caption,flickr30k,goyal2017vqav2,singh2019textvqa,marino2019okvqa,schwenk2022aokvqa,masry2022chartqa,mishra2019ocrvqa,wang2020estvqa,clark2017docqa,mathew2022infographicvqa} are primarily human-annotated and have limited data scale.
For example, VQAv2~\cite{goyal2017vqav2} annotated each image with several question-answer pairs, while Visual Genome~\cite{krishna2017visualgenome} further provided region-level annotations.
However, these datasets have limited data scales and fail to encompass diverse scenarios in the open world, hindering models' generalization ability.
To achieve open-world capability, CLIP~\cite{radford2021clip} and ALIGN~\cite{jia2021align} proposed training models using web-scale image-text pairs collected from the internet. 
Subsequent works~\cite{schuhmann2021laion400m,schuhmann2022laion5b,laioncoco,gadre2023datacomp,kakaobrain2022coyo700m,sharma2018cc3m,changpinyo2021cc12m,yfcc15m,thomee2016yfcc100m,wang2023allseeing,wang2024allseeingv2,peng2023kosmos2} have also been introduced for open-source research.
Among them, LAION-5B~\cite{schuhmann2022laion5b} is the pioneering dataset offering billion-scale image-text pairs, whereas AS-1B~\cite{wang2023allseeing} is the first extensive dataset to provide region-level image-text pairs.
However, these datasets contain limited world knowledge in each sample, affecting the performance of the underlying language model of VLMs.
Recently, a series of interleaved datasets~\cite{zhu2024mmc4,laurenccon2024obelics} have been proposed to address these issues. Nonetheless, the data source and the languages involved in these datasets are limited.
In this work, we propose the {\dsname}, the first 10 billion-level image-text interleaved dataset comprising multiple data sources and languages.

\subsection{Vision-Language Models}

Significant advancements have been made in the field of vision-language models (VLMs) in recent years.
Previous methods~\cite{bao2022vlbert,wang2022beit3} mainly focused on specific downstream tasks within predefined closed sets, while recent works have shifted towards understanding the open world.
Models trained with contrastive learning-based methods~\cite{radford2021clip,jia2021align,fang2022eva,chen2023internvl} are capable of recognizing and understanding open-world semantics through an image-text matching framework, although their lack of generative ability limits their applicability.
In recent years, the advancement of large language models (LLMs)~\cite{brown2020gpt3,openai2023gpt4,touvron2023llama} has led to the emergency of many LLM-based VLMs~\cite{zhu2022uni_p,li2023blip2,zhu2023minigpt-4, wang2023visionllm,liu2023interngpt,li2023videochat,wang2023allseeing,chen2024internvl1_5}.
As one of the representative works, InternVL-1.5 \cite{chen2024internvl1_5} achieves performance comparable to GPT-4V~\cite{gpt4v}.
Additionally, models like Kosmos-2 \cite{peng2023kosmos2} and ASMv2 \cite{wang2024allseeingv2} enable LLMs to comprehend specific regions within images.
Recently, a series of works \cite{sun2023emu1,sun2023emu2,tian2024mminterleaved,zhu2023vl_gpt,jin2023lavit,dong2023dreamllm,laurenccon2024idefics2} have explored the use of image-text interleaved data to enhance VLM capabilities.
However, the training corpora for these models remain limited to English data from Common Crawl. 
The effectiveness of image-text interleaved data from other sources or languages is still unexplored. In this work, we provide more empirical insights into the use of interleaved data.

\section{Data Engine}

\subsection{Overall Pipeline}
\label{sec:overall-pipeline}

\begin{figure}[t]
    \centering
    \includegraphics[width=\textwidth]{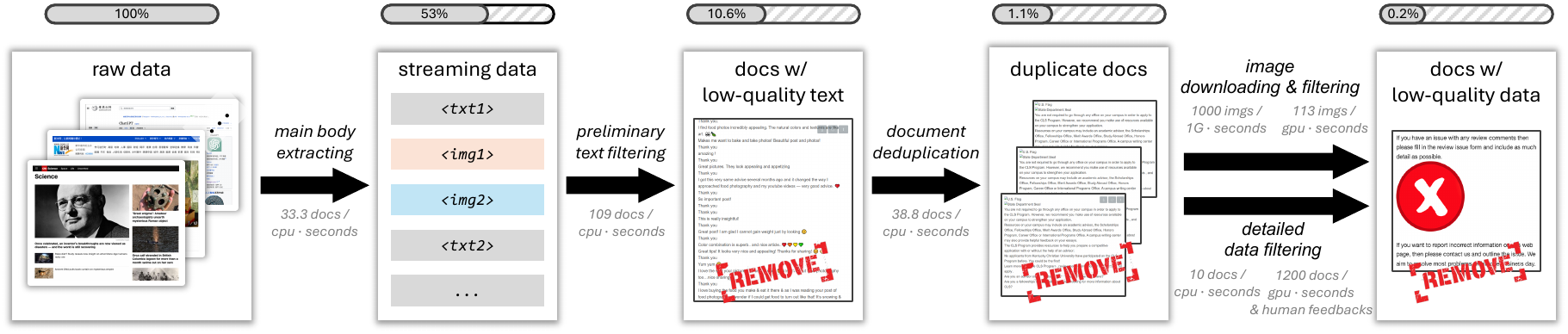}
    \caption{\textbf{Overview of the data processing pipeline.} 
    It contains five key stages: main body extraction, preliminary text filtering, document deduplication, image downloading \& filtering, and detailed text filtering. Each stage efficiently reduces the dataset to retain only high-quality data.}
    \label{fig:overall pipeline}
\end{figure}

Figure \ref{fig:overall pipeline} illustrates the overall pipeline of our data engine, which consists of five key stages as follows:

\noindent\textbf{Main Body Extraction.}
We extract primary content from each web document using an improved version of Trafilatura \cite{barbaresi-2021-Trafilatura}, which can more accurately and efficiently extract main content and images while handling a broader range of languages (see Section~\ref{sec:tw}). We enhance sections based on the HTML structure's density if the extracted content is insufficient. HTML documents without images are dropped in this stage. Some explicit advertisements or sidebars are excluded through HTML structure analysis and URL pattern matching for images. Then, we convert the HTML structure into the streaming data format, which is a unified data format applicable to different data sources.
It preserve tags for individual elements, including <text>, <image>, <code>, <header>, <detail>, <quote>, <video>, <audio>, <table>, and <list>.
During this step, we remove 47\% of documents.

\noindent\textbf{Preliminary Text Filtering.}
Given the streaming data from the main body extraction, we perform preliminary text filtering by employing strategies from Gopher~\cite{rae2021scaling} and C4~\cite{2020t5} to eliminate extremely low-quality content, such as documents with excessive numbers, documents with texts that are too long or too short, and documents containing ``lorem ipsum.'' Additionally, we introduce some heuristic rules to further filter the text, such as removing documents with too many continuous line breaks or documents where a single word's frequency is excessively high. During this step, we remove 80\% documents from the remaining HTML documents.

\noindent\textbf{Document Deduplication with Text.}
We remove duplicate documents by comparing their text content using minihash~\cite{broder1997resemblance} values with a threshold of 0.8 and retaining the latest version. This step significantly reduces redundancy, discarding approximately 90\% of duplicates.

\noindent\textbf{Image Downloading \& Filtering.}
In this step, we discard invalid images that were not successfully downloaded. Adhering to MMC4~\cite{zhu2024mmc4} guidelines, we filter out images with a height or width of fewer than 150 pixels and an aspect ratio greater than 2 or less than 0.5. Following LAION-5B~\cite{schuhmann2022laion5b}, we exclude images with an aesthetic score below 3.7 or a Not Safe for Work (NSFW) score above 0.8. Additionally, we identify and remove images that appear more than 10 times across HTML documents by computing perceptual hash (phash) and difference hash (dhash) values.

\noindent\textbf{Detailed Text Filtering.}
We finetune models based on BERT \cite{devlin2018bert} for scoring advertisement content, political content, toxic content, NSFW material, and document fluency. Using these models, we discard documents containing excessive ads, inappropriate content, or poor language quality. In addition, to further improve data quality, we use a human-feedback filtering strategy (see Section \ref{sec:human-feedback-filtering}) to develop a multimodal filter suitable for English and non-English content.

In addition, we enhance the diversity of our dataset by creating storyboard datasets from various video sources. This includes extracting keyframes and transcribing audio content from YT-Temporal-1B~\cite{zellers2022merlot}, HD-VILA-100M~\cite{xue2022hdvila}, HowTo100M~\cite{miech2019howto100m}, and InternVid~\cite{wang2023internvid}. More details can be found in the supplementary material.

\subsection{Tweakings}
\label{sec:tw}

To enhance the effectiveness and efficiency of our pipeline, we carefully refine the data pipeline from key aspects as follows:

\noindent\textbf{Pre-Deduplication.}
The resources required for image downloading, filtering, and detailed text filtering are substantial, involving significant bandwidth, GPU resources, and human feedback. Given that the deduplication step filters out a large number of documents and images, we choose to perform deduplication in advance. This approach effectively reduces the number of images to be downloaded and the volume of documents requiring detailed text filtering. As a result, it saves approximately 86 PB seconds of bandwidth in downloading images, 4500 A100 GPU days in image filtering, and 130 GPU days along with 45 person-days in detailed text filtering.

\noindent\textbf{Improved Main Body Extraction.}
Our extraction algorithm has been significantly improved compared to the vanilla Trafilatura \cite{barbaresi-2021-Trafilatura}. 
In terms of accuracy, we have addressed the issue where Trafilatura would overlook the main content of an HTML document when extracting images, and enhanced its capability to handle Chinese, Japanese, and Arabic documents. Additionally, we have incorporated techniques to trim web noise regions based on HTML structure (such as clusters of lists and navigation bars) and style (targeting elements like advertisements, comments, JavaScript, and CSS). 
In terms of efficiency, we optimized the process based on HTML nodes and streamlined the processing pipeline by eliminating the fallback process in challenging cases. With these two improvements, we can not only extract more informative content from the main body but also double the speed of the extraction process.

\noindent\textbf{Improved Image Downloading.}
We integrate efficient download task scheduling and resource allocation while employing Bloom filtering technology \cite{bloom1970space} to deduplicate URLs of images that have been downloaded or are pending processing. 
This method effectively prevents redundant download requests, optimizing storage resources and bandwidth usage. Consequently, it provides robust technical support for the efficient collection and analysis of large-scale image data. 
Specifically, our approach reduces URL download requests from 30 billion to 9.65 billion and accelerates the downloading process by a factor of 1.5.

\textbf{Pipeline Parallelism.} 
Our pipeline runs in a modular parallel manner, offering several benefits. (1) {The system will have greater fault tolerance since we can modify or improve each section of the pipeline independently.}
(2) Different parts of the pipeline require different types of resources, {such as main body extraction runs on CPUs, image filtering runs on GPUs, and image downloading requires bandwidth,} so a modular design is more reasonable. (3) by allocating resources based on throughput rather than evenly distributing them, we can significantly speed up the process. Compared to equal resource allocation, our parallel assembly line achieves a 1.39 times speed increase.

With all these improvements, the dataset processing pipeline can now scale up to thousands of CPUs, thousands of GPUs, and 3Gbps bandwidth, tripling its processing speed in that configuration.

\subsection{Human-Feedback Filtering}
\label{sec:human-feedback-filtering}

Based on the pipeline introduced in Section~\ref{sec:overall-pipeline}, a significant portion of low-quality data has been removed.
However, the remaining documents are still noisy.
In this section, we introduce the human-feedback filtering method used to optimize the text filters, further improving the document quality.
The optimized filter comprises nearly 30 filtering rules for English and 40 for Chinese. These filtering rules can be found in the Appendix.

To build these filtering rules, we first sample a subset of documents according to various criteria, including completeness, comprehensibility, fluency, relevance, and safety.
After that, we manually design additional filtering rules to remove the low-quality documents from these sampled documents.
These rules are then evaluated on a human-annotated evaluation set, and those achieving excellent performance are added to our filtering pipeline. The evaluation metric includes the miss rate and the false positive rate.
By repeating the above process, we can iteratively optimize the quality and comprehensiveness of text filters based on human feedback.

\subsection{Streaming Data Format}
\label{sec:streaming-data-format}

We use a comprehensive and unified streaming data format to preserve rich and diverse information about the original data.
Given an HTML document, we first split it into several chunks according to its layout, each formulated as image-text interleaved sequences $x=\left(x_1,x_2,...,x_n\right)$, where $x_i$ can be a text sentence or an image.
Then we concatenate these chunks in a top-to-bottom, left-to-right order to obtain a streaming interleaved sequence.

Based on this data format, the formulation of HTML documents, image-text pairs, and video sequences can be easily unified, which means that we can process these heterogeneous data from different sources in a unified manner.
In addition to the content of the given data, other meta-annotations, including image aesthetic scores, image/text NSFW scores, political scores, toxic scores, unsafe scores, and text fluency, are also included in the streaming data. We hope that these meta-annotations can help researchers to better understand and utilize the dataset for various applications.

\section{Exploring \dsname}

\begin{figure}[t]
    \centering
    \begin{minipage}{0.475\textwidth}
        \centering
        \includegraphics[width=0.7\linewidth]{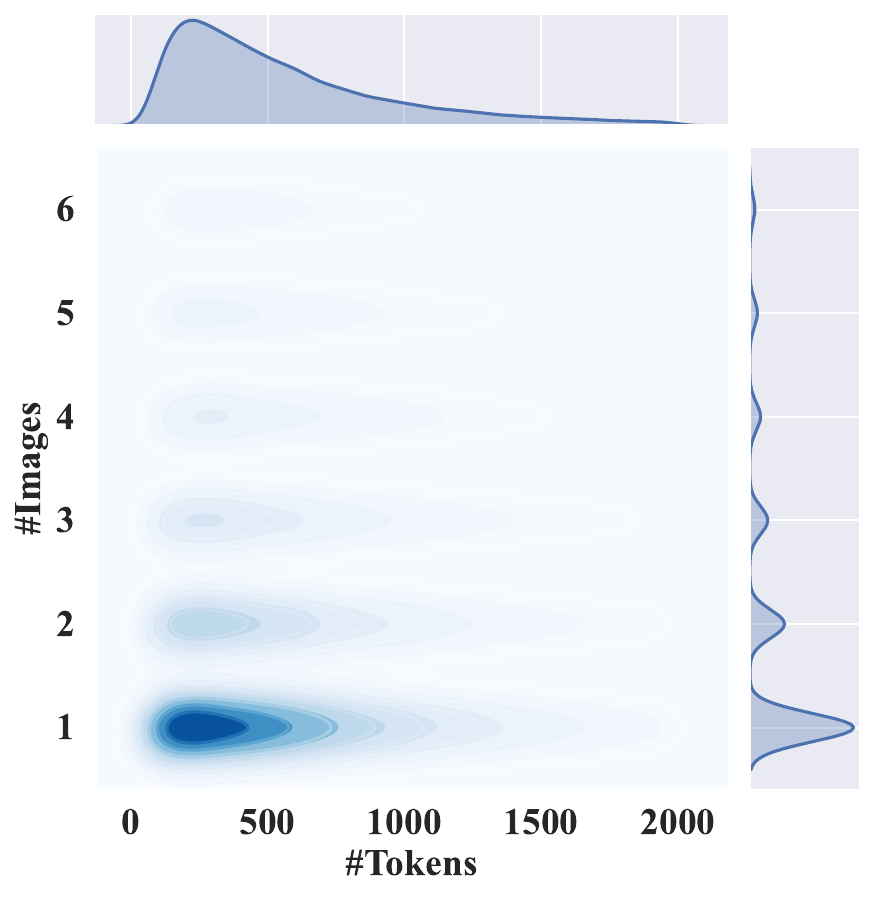}
        \caption{
            \textbf{Joint distribution of the image and token numbers per document.}
            We use kernel density estimate to get the distribution.
        }
        \label{fig:img-text distribution}
    \end{minipage}\hfill
    \begin{minipage}{0.475\textwidth}
        \centering
        \includegraphics[width=0.7\textwidth]{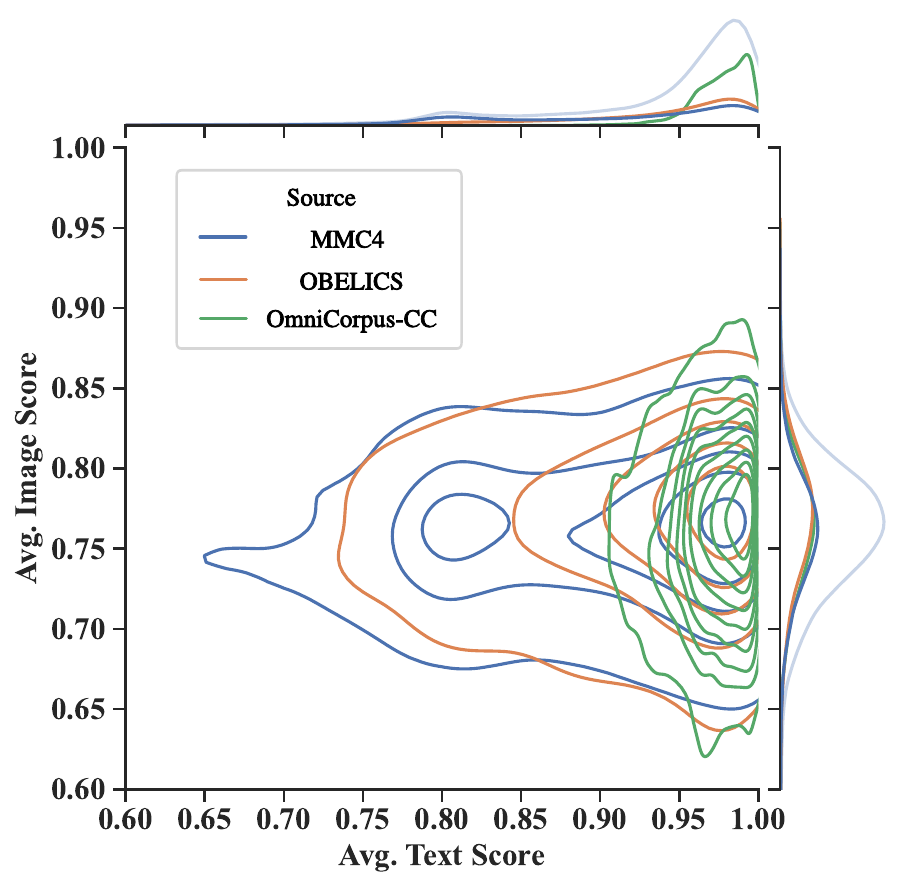}
        \caption{
            \textbf{Joint distribution of text and image score PDFs.}
            We visualized and compared the joint distribution of the PDFs of the Text Scores and Image Scores across each dataset.
        }
        \label{fig:text_score_cdf}
    \end{minipage}
\end{figure}

\noindent\textbf{General Statistics.}
As shown in Table~\ref{tab:dataset_statistics}, our {\dsname} is currently the largest and the first open-source multilingual interleaved dataset.
It surpasses the combined totals of all previous interleaved datasets~\cite{huang2023kosmos-1, mckinzie2024mm1, zhu2024mmc4, laurenccon2024obelics}.
Figure~\ref{fig:img-text distribution} illustrates the joint distribution of text tokens and images in the interleaved sequences from {\dsname}.
See Appendix for more details.

\begin{figure}
    \centering
    \includegraphics[width=0.9\linewidth]{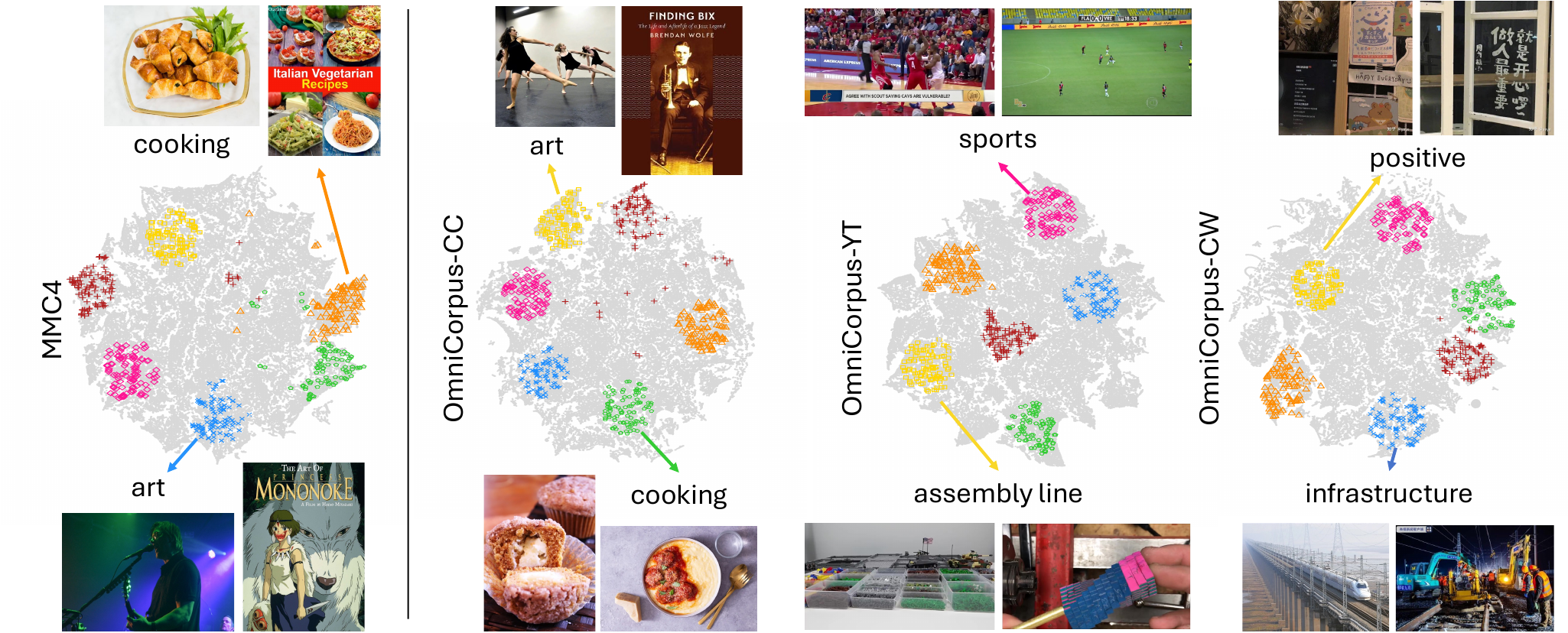}
    \caption{
        \textbf{Visualization of topic clusters and example images.} 
        The four diagrams from left to right correspond to MMC4~\cite{zhu2024mmc4}, \subsetnamecc, \subsetnamevideo, and \subsetnamecn. The clusters are T-SNE~\cite{van2008visualizing} projection of LDA~\cite{blei2003lda} topic modeling results.
        We select 2 topics for each dataset and show two image examples for each topic.
    }
    \label{fig:diversity-visualization}
\end{figure}

\noindent\textbf{Diversity Analysis.} %
To measure and analyze the diversity of document content, we follow previous studies~\cite{zhu2024mmc4,laurenccon2024obelics} and employ Latent Dirichlet Allocation (LDA)~\cite{blei2003lda} to assess the topic diversity of the dataset. 
Figure~\ref{fig:diversity-visualization} illustrates the significant differences in topics across documents from different sources, highlighting the importance of various sources in enhancing data diversity. 
The detailed topic modeling results are presented in the Appendix.

\noindent\textbf{Qualitative Assessment of Dataset Samples.}
We randomly sample 200 documents from {\subsetnamecc} to evaluate their quality. There are 405 images in these documents. Among them, 88.4\% are relevant to the documents, 8.0\% contain watermarks, 4.0\% contain logos, and 0.2\% are advertisements. 
Additionally, 86.4\% of the documents feature photographic images, while 13.6\% included graphic images such as cartoons. Furthermore, 32.1\% of the images contain at least one written word, and 22.7\% of the images contain structured text. No NSFW images were found.

\noindent\textbf{Quality Validation.}
As illustrated in Figure \ref{fig:text_score_cdf}, we present the joint distribution of text scores and image scores across each set of 1 million sampled documents. The image score is calculated as the average of the aesthetic score and the NSFW score. The text score is determined by averaging the advertisement content score, the NSFW content score, and the document fluency score.
In terms of image scores, all datasets perform similarly. The {\subsetnamecc} exhibits superior text quality. Specifically, our {\subsetnamecc} has a lower proportion of low-quality text compared to other datasets, with the difference diminishing as test quality increases. This indicates a higher proportion of high-quality tests in {\subsetnamecc}.

\section{Experiments}

\subsection{Experimental Settings}
\label{sec:experiments-settings}

\noindent\textbf{Baselines.}
We construct our baseline models following LLaVA-1.5~\cite{liu2023llava1_5}, which comprises a vision encoder, a multimodal projector, and an LLM.
The input sequence to the LLM is a token sequence consisting of interleaved visual and textual tokens. The language modeling loss is used to train the model, which is only calculated on text tokens.
Unless otherwise specified, we employ CLIP-ViT-L-336px~\cite{radford2021clip} as the vision encoder and Vicuna-1.5-7B~\cite{zheng2023vicuna} as the LLM.

\noindent\textbf{Evaluation.}
We evaluate our models on VQA benchmarks~\cite{goyal2017vqav2,singh2019textvqa,gurari2018vizwiz,marino2019okvqa} and image captioning benchmarks~\cite{chen2015coco-caption,flickr30k}.
The accuracy score is used for VQA, while CIDEr~\cite{vedantam2015cider} is used for image captioning. 
Following OpenFlamingo~\cite{awadalla2023openflamingo}, we extend the benchmarks to few-shot settings to assess in-context learning. 
Specifically, in-context examples are sampled using RICES~\cite{yang2022rices}.

\subsection{Main Findings}

\begin{figure}[t]
  \begin{minipage}[b]{0.41\linewidth}
    \centering
    \includegraphics[width=0.9\textwidth]{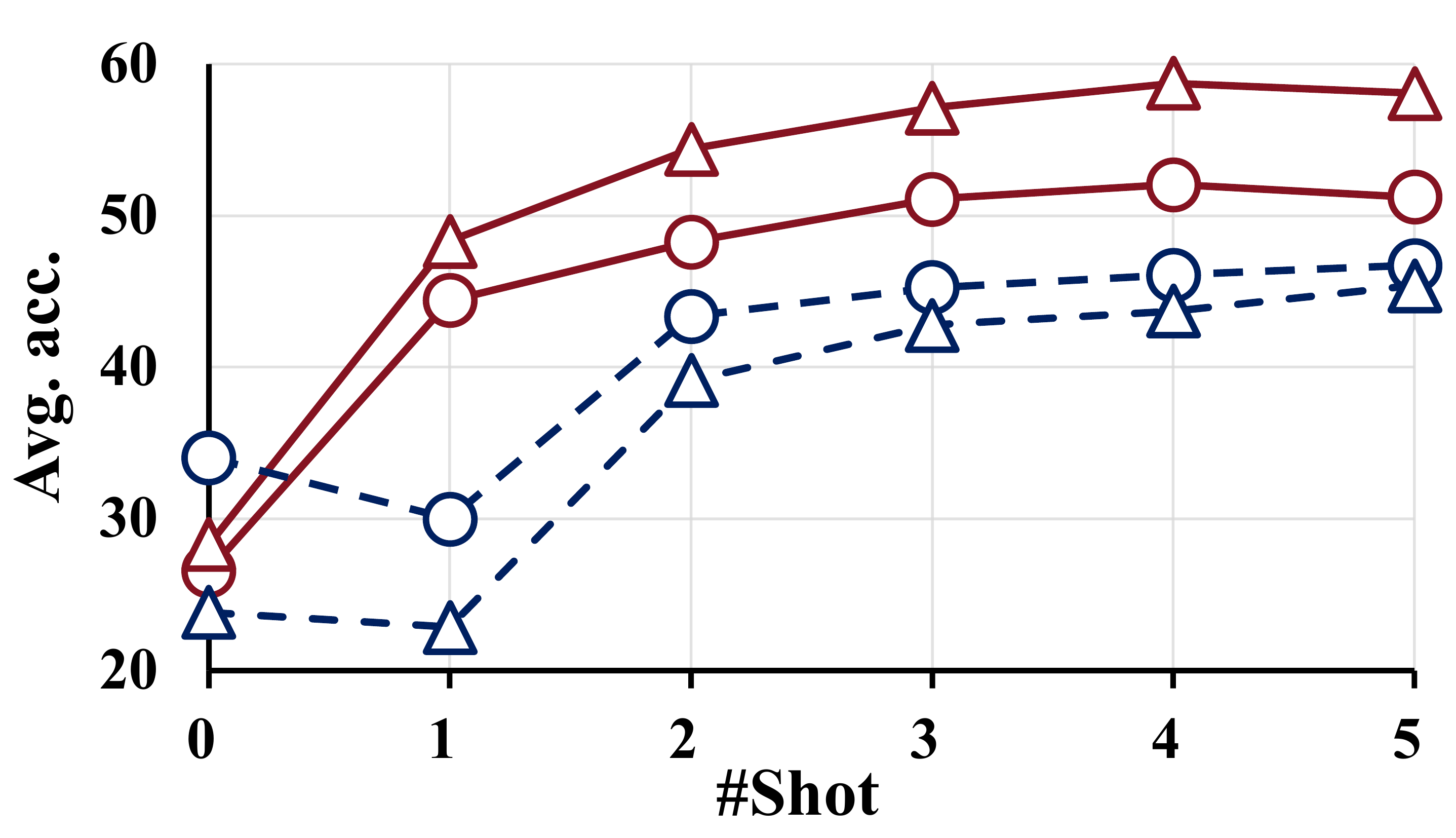}
    \caption{
        \textbf{Ablation on image position strategies.}
        Red solid: fully autoregressive. Blue dashed: cross-attention. Triangular: natural. Circular: retrieval-based.
    }
    \label{fig:res_img_pos}
  \end{minipage}
  \hfill
  \begin{minipage}[b]{0.55\linewidth}
    \centering
\captionof{table}{
    \textbf{Ablation on pre-training and SFT data types.} 
    We report the zero/few-shot average accuracies of the four MLLM benchmarks and the text-only MMLU benchmark.
    The first row hosts the initialized model which has not been trained with vision-language data.
}
\label{tab:res_sft}
\vspace{2px}
\resizebox{\textwidth}{!}{%
\begin{tabular}{ll|cccc|cc}
\toprule
\renewcommand{\arraystretch}{0.95}
\multirow{2}{*}{\begin{tabular}[c]{@{}l@{}}Pre-training\\ Data\end{tabular}} & \multirow{2}{*}{\begin{tabular}[c]{@{}l@{}}SFT \\ Data\end{tabular}} & \multicolumn{4}{c|}{Avg. MLLM acc.} & \multicolumn{2}{c}{MMLU acc.} \\
                                                                             &                                                                      & 0       & 1      & 2      & 4      & 0             & 5             \\ \midrule
-                                                                            & -                                                                    & -       & -      & -      & -      & 48.7          & 49.9          \\
Interleaved                                                                  & -                                                                    & 28.3    & 48.3   & 54.4   & 58.7   & 47.1          & 48.6          \\
-                                                                            & Common                                                               & 76.3    & 71.7   & 72.6   & 73.1   & 50.3          & 50.5          \\
Interleaved                                                                  & Common                                                               & 76.5    & 73.0   & 73.3   & 73.9   & 50.4          & 51.2          \\
\rowcolor[HTML]{EBEBEB} 
Interleaved                                                                  & Interleaved                                                          & 74.5    & 77.7   & 78.1   & 77.9   & 50.8          & 51.3          \\ \bottomrule
\end{tabular}%
}

  \end{minipage}
\end{figure}

\noindent\textbf{Different image position strategies excel in different architectures.}
Existing multimodal document datasets organize interleaved image and text sequences in two main ways.
The MMC4 dataset~\cite{zhu2024mmc4} employed a retrieval strategy, inserting images into text sequence based on CLIP similarities, while the OBELICS dataset~\cite{laurenccon2024obelics} maintained the natural layout of the source webpage.
We conducted ablation studies on {\subsetnamecc} to evaluate both strategies using a fully autoregressive architecture like LLaVA-1.5~\cite{liu2023llava1_5} and a cross-attention architecture like Flamingo~\cite{alayrac2022flamingo}.
As shown in Figure~\ref{fig:res_img_pos}, the natural strategy performs better with the fully autoregressive architecture, whereas the retrieval-based strategy excels with the cross-attention architecture.
This suggests that the cross-attention architecture benefits from optimal correlation between images and their surrounding paragraphs, while the fully autoregressive architecture prefers a natural arrangement that aligns with typical reading habits.

\begin{table}[t]
\centering
\caption{
    \textbf{Pre-training ablation on curated subsets.} 
    We report the zero/few-shot results on four MLLM benchmarks, including two VQA and two image captioning tasks. 
    The first column shows the number of documents per subset, with 1M documents randomly sampled for training.
}
\vspace{2px}
\scriptsize
\setlength\tabcolsep{3pt}
\label{tab:res_curated_subsets}
\begin{tabular}{r|cccc|cccc|cccc|cccc|cccc}
\toprule
Eval Set  & \multicolumn{4}{c|}{OKVQA}                                    & \multicolumn{4}{c|}{TextVQA}                                  & \multicolumn{4}{c|}{COCO}                                     & \multicolumn{4}{c|}{Flickr30k}                                & \multicolumn{4}{c}{Avg.}                                      \\
\#Shot & 0             & 1             & 2             & 4             & 0             & 1             & 2             & 4             & 0             & 1             & 2             & 4             & 0             & 1             & 2             & 4             & 0             & 1             & 2             & 4             \\ \midrule
\phantom{.}988M        & 15.2          & 34.1          & 31.8          & 32.8          & 21.7          & 30.5          & 34.6          & 37.7          & 41.9          & 73.6          & 85.0          & 94.9          & 34.2          & 41.4          & 47.5          & 52.6          & 28.2          & 44.9          & 49.7          & 54.5          \\
\phantom{.}600M        & \textbf{17.1} & 34.9          & 32.3          & 30.1          & \textbf{23.0} & 31.7          & 35.8          & 37.9          & 41.4          & 75.3          & 85.7          & 96.9          & 34.2          & 43.6          & 48.8          & \textbf{55.8} & \textbf{28.9} & 46.4          & 50.6          & 55.1          \\
\rowcolor[HTML]{EBEBEB} 
\phantom{.}200M        & 12.7          & \textbf{36.0} & 38.8          & 41.1          & 17.7          & 32.6          & \textbf{38.0} & \textbf{42.0} & \textbf{46.9} & \textbf{80.8} & \textbf{92.2} & \textbf{97.2} & \textbf{36.1} & 43.9          & 48.6          & 54.3          & 28.3          & \textbf{48.3} & \textbf{54.4} & \textbf{58.7} \\
\phantom{0.}40M        & 13.4          & 35.5          & 38.6          & \textbf{41.4} & 17.1          & 32.1          & 35.9          & 39.4          & 38.3          & 79.8          & 91.6          & 96.0          & 29.5          & \textbf{44.0} & 47.7          & 53.6          & 24.6          & 47.8          & 53.5          & 57.6          \\
\phantom{00.}8M        & 12.2          & 35.6          & 38.2          & 40.8          & 15.9          & 32.9          & 36.3          & 38.2          & 41.5          & 78.2          & 89.4          & 93.5          & 32.4          & 42.9          & \textbf{49.0} & 51.6          & 25.5          & 47.4          & 53.2          & 56.0          \\
\phantom{0}2.5M        & 13.5          & 35.7          & \textbf{39.1} & 41.3          & 18.2          & \textbf{33.2} & 37.7          & 41.1          & 46.4          & 78.9          & 91.9          & 95.9          & 35.4          & 43.7          & 48.8          & 54.5          & 28.4          & 47.9          & 54.4          & 58.2          \\ \bottomrule
\end{tabular}%
\vspace{-3mm}
\end{table}

\begin{table}[t]
\centering
\caption{
    \textbf{Comparison with open-source interleaved image-text datasets.} 
    We report the zero/few-shot results on four MLLM benchmarks. 
    The best two results are highlighted with bold font.
}
\scriptsize
\setlength\tabcolsep{2.4pt}
\label{tab:res_dataset_comparison}
\begin{tabular}{l|cccc|cccc|cccc|cccc|cccc}
\toprule
Eval Set              & \multicolumn{4}{c|}{OKVQA}                                                            & \multicolumn{4}{c|}{TextVQA}                                                          & \multicolumn{4}{c|}{COCO}                                                             & \multicolumn{4}{c|}{Flickr30k}                                                        & \multicolumn{4}{c}{Avg.}                                                        \\
\#Shot                & 0                   & 1                   & 2                   & 4                   & 0                   & 1                   & 2                   & 4                   & 0                   & 1                   & 2                   & 4                   & 0                   & 1                   & 2                   & 4                   & 0             & 1                   & 2                   & 4                   \\ \midrule
MMC4~\cite{zhu2024mmc4}          & \textbf{15.1}       & 29.0                & 24.0                & 23.2                & \textbf{21.2}       & 27.6                & 30.3                & 33.8                & 45.7                & 70.9                & 82.1                & 88.4                & \textbf{36.3}       & 32.5                & 39.0                & 43.8                & \textbf{29.6} & 40.0                & 43.9                & 47.3                \\
MMC4-Core~\cite{zhu2024mmc4}   & 13.5                & 29.5                & 27.1                & 26.8                & 20.5                & 27.1                & 32.1                & 35.6                & 41.0                & 72.1                & 84.6                & 90.3                & 34.3                & 37.5                & 41.1                & 45.6                & 27.3          & 41.5                & 46.2                & 49.6                \\
OBELICS~\cite{laurenccon2024obelics}        & 13.9                & 35.0                & 36.8                & \textbf{40.2}       & 17.9                & 30.3                & 35.7                & 40.7                & {\textbf{50.7}} & \textbf{74.7}       & \textbf{91.3}       & \textbf{97.1}       & {\textbf{42.7}} & \textbf{41.4}       & \textbf{47.5}       & {\textbf{54.7}} & \textbf{31.3} & \textbf{45.3}       & \textbf{52.9}       & \textbf{58.2}       \\ \midrule
\rowcolor[HTML]{EBEBEB} 
{\subsetnamevideo}    & {\textbf{16.5}} & {\textbf{36.1}} & \textbf{38.4}       & 40.1                & {\textbf{22.9}} & {\textbf{34.5}} & {\textbf{38.1}} & \textbf{41.0}       & 40.6                & 71.2                & 78.0                & 83.8                & 32.9                & 30.0                & 32.2                & 36.0                & 28.2          & 43.0                & 46.6                & 50.2                \\
\rowcolor[HTML]{EBEBEB} 
{\subsetnamecc}       & 12.7                & \textbf{36.0}       & {\textbf{38.8}} & {\textbf{41.1}} & 17.7                & \textbf{32.6}       & \textbf{38.0}       & {\textbf{42.0}} & \textbf{46.9}       & {\textbf{80.8}} & {\textbf{92.2}} & {\textbf{97.2}} & 36.1                & {\textbf{43.9}} & {\textbf{48.6}} & \textbf{54.3}       & 28.3          & {\textbf{48.3}} & {\textbf{54.4}} & {\textbf{58.7}} \\ \bottomrule
\end{tabular}%
\vspace{-3mm}
\end{table}

\begin{table}[t]
\centering
\caption{
    \textbf{Comparison with state-of-the-art MLLMs pre-trained with interleaved image-text data.} 
    ``*'' indicates that the zero-shot evaluation follows Flamingo~\cite{alayrac2022flamingo}, which actually includes two text-only examples.
    The prompt for TextVQA~\cite{singh2019textvqa} does not contain OCR tokens.
    To align with the evaluation setting of comparison models, we sample the in-context examples randomly. 
}
\label{tab:res_final}
\scriptsize
\setlength\tabcolsep{5.7pt}
\begin{tabular}{lcccccccc}
\toprule
Model                             & Pre-training Data                                                                                               & \#Shot        & COCO                                  & Flickr30k                             & OKVQA                                 & TextVQA                               & VQAv2                        & VizWiz                       \\ \midrule
                                  &                                                                                                                 & \phantom{*}0* & 79.5                                  & \textbf{59.5}                         & 37.8                                  & 24.2                                  & 52.7                         & 27.5                         \\
                                  &                                                                                                                 & 4             & \cellcolor[HTML]{F5F0ED}89.0          & \cellcolor[HTML]{F5F0ED}\textbf{65.8} & \cellcolor[HTML]{F5F0ED}40.1          & \cellcolor[HTML]{F5F0ED}28.2          & \cellcolor[HTML]{F5F0ED}54.8 & \cellcolor[HTML]{F5F0ED}34.1 \\
\multirow{-3}{*}{OpenFlamingo-9B~\cite{awadalla2023openflamingo}} & \multirow{-3}{*}{\begin{tabular}[c]{@{}c@{}}MMC4 \\ LAION\end{tabular}}                                          & 8             & \cellcolor[HTML]{DFEBF7}96.3          & \cellcolor[HTML]{DFEBF7}62.9          & \cellcolor[HTML]{DFEBF7}41.1          & \cellcolor[HTML]{DFEBF7}29.1          & \cellcolor[HTML]{DFEBF7}54.8 & \cellcolor[HTML]{DFEBF7}38.5 \\ \midrule
                                  &                                                                                                                 & \phantom{*}0* & 46.0                                  & 27.3                                  & 38.4                                  & 25.9                                  & 50.9                         & 35.5                         \\
                                  &                                                                                                                 & 4             & \cellcolor[HTML]{F5F0ED}\textbf{93.0} & \cellcolor[HTML]{F5F0ED}59.7          & \cellcolor[HTML]{F5F0ED}45.4          & \cellcolor[HTML]{F5F0ED}27.6          & \cellcolor[HTML]{F5F0ED}55.4 & \cellcolor[HTML]{F5F0ED}36.9 \\
\multirow{-3}{*}{IDEFICS-9B~\cite{laurenccon2024obelics}}      & \multirow{-3}{*}{\begin{tabular}[c]{@{}c@{}}OBELICS\\ Wikipedia\\ LAION, PMD\end{tabular}}                       & 8             & \cellcolor[HTML]{DFEBF7}97.0          & \cellcolor[HTML]{DFEBF7}61.9          & \cellcolor[HTML]{DFEBF7}\textbf{47.7} & \cellcolor[HTML]{DFEBF7}27.5          & \cellcolor[HTML]{DFEBF7}56.4 & \cellcolor[HTML]{DFEBF7}40.4 \\ \midrule
                                  &                                                                                                                 & \phantom{*}0* & $-$                                     & $-$                                     & 42.8                                  & $-$                                     & 52.9                         & 34.4                         \\
                                  &                                                                                                                 & 4             & \cellcolor[HTML]{F5F0ED}$-$             & \cellcolor[HTML]{F5F0ED}$-$             & \cellcolor[HTML]{F5F0ED}$-$             & \cellcolor[HTML]{F5F0ED}$-$             & \cellcolor[HTML]{F5F0ED}58.4 & \cellcolor[HTML]{F5F0ED}41.3 \\
\multirow{-3}{*}{Emu-14B~\cite{sun2023emu1}}         & \multirow{-3}{*}{\begin{tabular}[c]{@{}c@{}}LAION, LAION-COCO\\ MMC4, WebVid-10M\\ YT-Storyboard-1B\end{tabular}} & 8             & \cellcolor[HTML]{DFEBF7}$-$             & \cellcolor[HTML]{DFEBF7}$-$             & \cellcolor[HTML]{DFEBF7}$-$             & \cellcolor[HTML]{DFEBF7}$-$             & \cellcolor[HTML]{DFEBF7}59.0 & \cellcolor[HTML]{DFEBF7}43.9 \\ \midrule
                                  &                                                                                                                 & \phantom{*}0* & \textbf{81.2}                         & 59.2                                  & \textbf{45.0}                         & \textbf{43.0}                         & \textbf{63.2}                           & \textbf{49.8}                           \\
                                  &                                                                                                                 & 4             & \cellcolor[HTML]{F5F0ED}91.9          & \cellcolor[HTML]{F5F0ED}63.2          & \cellcolor[HTML]{F5F0ED}\textbf{45.5} & \cellcolor[HTML]{F5F0ED}\textbf{45.4} & \cellcolor[HTML]{F5F0ED}\textbf{64.5}   & \cellcolor[HTML]{F5F0ED}\textbf{51.3}   \\
\multirow{-3}{*}{\textbf{Ours (7B)}} & \multirow{-3}{*}{\begin{tabular}[c]{@{}c@{}}LAION\\ {\subsetnamecc}\end{tabular}}                            & 8             & \cellcolor[HTML]{DFEBF7}\textbf{97.6} & \cellcolor[HTML]{DFEBF7}\textbf{63.5} & \cellcolor[HTML]{DFEBF7}46.6          & \cellcolor[HTML]{DFEBF7}\textbf{45.6} & \cellcolor[HTML]{DFEBF7}\textbf{64.7}   & \cellcolor[HTML]{DFEBF7}\textbf{52.2}   \\ \bottomrule
\end{tabular}%
\vspace{-3mm}
\end{table}

\noindent\textbf{Data filtering benefits MLLMs to some extent.}
We further construct several curated subsets of approximately 600M, 200M, 40M, 8M, and 2.5M documents from {\subsetnamecc}, according to the meta-annotations introduced in Section~\ref{sec:streaming-data-format}.
To validate the benefits of data filtering, we trained baseline models using 1M documents randomly sampled from subsets, separately. 
As shown in Table~\ref{tab:res_curated_subsets}, the model trained on the 200M subset outperforms those trained on larger subsets and performs similarly to the model trained on smaller subsets. This suggests that data filtering can improve data quality, but over-filtering may harm performance due to data homogenization.

\noindent\textbf{Image-text interleaved fine-tuning maintains in-context learning ability.}
We pre-train the baseline architecture with 1M documents randomly sampled from {\subsetnamecc} and fine-tune it using the LLaVA-665K dataset \cite{liu2023llava1_5}. 
We compare zero-shot and few-shot performance on four MLLM benchmarks, as well as a text-only benchmark (\ie, MMLU~\cite{hendrycks2020mmlu}), as shown in Table~\ref{tab:res_sft}. 
The image-text interleaved pre-trained model shows a stepwise improvement with more in-context examples. After fine-tuning with high-quality conversation samples, there are overall enhancements for the average performance on four MLLM benchmarks, but the positive correlation with the example number is no longer maintained. Additionally, we replace the caption and VQA samples in the SFT data with few-shot samples whose format is aligned with the evaluation, yielding significantly improved few-shot performance. Despite the slight decline in zero-shot performance, the best few-shot average score shows considerable improvement compared to the baseline. Therefore, including image-text interleaved samples in SFT data is still essential. Furthermore, due to the absence of text-only instruction following samples in this setting, the model's language capability decreased. However, the high-quality data used in SFT significantly improved the language ability, effectively mitigating the disadvantages introduced during the pre-training phase.

\noindent\textbf{{\subsetnamevideo} boosts VQA performance while degrading captioning ability.}
The previous studies have merely incorporated storyboard samples into a pre-training data mixture without thoroughly investigating the specific impact. Our goal is to pre-train an MLLM exclusively using documents collected from video and evaluate it on image-text benchmarks. 
We randomly selected 1M samples from {\subsetnamevideo}. For each sample of video frames with text, we uniformly extracted six frames as images for the document and removed the remaining frames, constructing an image-text interleaved document. As shown in Table~\ref{tab:res_dataset_comparison}, the model trained on sampled {\subsetnamevideo} achieves the best VQA capabilities, but its captioning scores are the lowest. The results demonstrate the feasibility of extracting image-text interleaved documents from video resources.

\noindent\textbf{{\subsetnamecn} improves the Chinese ability.}
We pre-train on 1M Chinese documents randomly sampled from {\subsetnamecn} and fine-tune with LLaVA-665K data~\cite{liu2023llava1_5}. We find that the scores improve from 59.8 to 62.5 (+2.7) for MMBench-CN~\cite{liu2023mmbench} and from 23.6 to 24.9 (+1.3) for CMMMU~\cite{zhang2024cmmmu},
demonstrating the effectiveness of our {\subsetnamecn} data.

\subsection{Comparison Experiments}

To compare the data quality to the related dataset, we train the same baseline architecture with 1M documents randomly selected from MMC4, MMC4-Core~\cite{zhu2024mmc4}, OBELICS~\cite{laurenccon2024obelics}, and {\subsetnamecc}, respectively. As is shown in Table~\ref{tab:res_dataset_comparison}, the {\subsetnamecc} exhibits optimal few-shot performance and near-optimal zero-shot performance. 

To demonstrate the potential of the {\dsname} for large-scale MLLMs pre-training, we design a recipe for training a competitive 7B baseline foundation model with our dataset. We replace the LLM with InternLM2-7B~\cite{cai2024internlm2}. Additionally, we collect a large-scale data mixture, including image-text interleaved data ({\subsetnamecc}), paired image-text data (LAION~\cite{schuhmann2022laion5b}), and text-only data. We compare our model with OpenFlamingo~\cite{awadalla2023openflamingo} mainly pre-trained with MMC4~\cite{zhu2024mmc4} and IDEFICS mainly pre-trained with OBELICS~\cite{laurenccon2024obelics}. We follow them to add two evaluation sets, VQAv2~\cite{goyal2017vqav2} and VizWiz~\cite{gurari2018vizwiz}, for evaluating the pre-trained models. The evaluation setting is aligned with the OpenFlamingo~\cite{awadalla2023openflamingo}. The comparison performance is presented in Table~\ref{tab:res_final}. We can see that our 7B model is superior to the larger 9B OpenFlamingo and IDEFICS in most cases. Especially for VQAv2 and TextVQA, our model achieves a cliff lead.

\section{Conclusion \& Limitation}
\label{sec:conclusion}

In this work, we introduce the {\dsname} dataset, the largest multimodal dataset to date. This dataset contains 8.6 billion images, 1,696 billion text tokens, and 2.2 billion documents, which are collected from three data sources: Common Crawl, Chinese websites, and video platforms.
We elaborate on the data engine used to construct this dataset and carefully analyze its diversity and quality.
Experimental results demonstrate the effectiveness of our {\dsname}. We also provide some new insights according to these experiments.

Regarding limitations, the current filtering process offers limited improvements to the model's performance. Demonstrating which specific factors meet the conditions that benefit the model is complicated and is not thoroughly explored in this study and will be left for future work.

\noindent\textbf{Broader Impact.}
We hope this work can provide a solid data foundation for the future advancement of MLLMs.
We do not foresee obvious undesirable ethical/social impacts at this moment.

{
\small
\bibliographystyle{ieeenat_fullname}
\bibliography{egbib}

\begin{thebibliography}{135}
\providecommand{\natexlab}[1]{#1}
\providecommand{\url}[1]{\texttt{#1}}
\expandafter\ifx\csname urlstyle\endcsname\relax
  \providecommand{\doi}[1]{doi: #1}\else
  \providecommand{\doi}{doi: \begingroup \urlstyle{rm}\Url}\fi

\bibitem[Achiam et~al.(2023)Achiam, Adler, Agarwal, Ahmad, Akkaya, Aleman, Almeida, Altenschmidt, Altman, Anadkat, et~al.]{openai2023gpt4}
Josh Achiam, Steven Adler, Sandhini Agarwal, Lama Ahmad, Ilge Akkaya, Florencia~Leoni Aleman, Diogo Almeida, Janko Altenschmidt, Sam Altman, Shyamal Anadkat, et~al.
\newblock Gpt-4 technical report.
\newblock \emph{arXiv preprint arXiv:2303.08774}, 2023.

\bibitem[Alayrac et~al.(2022)Alayrac, Donahue, Luc, Miech, Barr, Hasson, Lenc, Mensch, Millican, Reynolds, et~al.]{alayrac2022flamingo}
Jean-Baptiste Alayrac, Jeff Donahue, Pauline Luc, Antoine Miech, Iain Barr, Yana Hasson, Karel Lenc, Arthur Mensch, Katie Millican, Malcolm Reynolds, et~al.
\newblock Flamingo: a visual language model for few-shot learning.
\newblock \emph{NeurIPS}, 2022.

\bibitem[Awadalla et~al.(2023)Awadalla, Gao, Gardner, Hessel, Hanafy, Zhu, Marathe, Bitton, Gadre, Sagawa, et~al.]{awadalla2023openflamingo}
Anas Awadalla, Irena Gao, Josh Gardner, Jack Hessel, Yusuf Hanafy, Wanrong Zhu, Kalyani Marathe, Yonatan Bitton, Samir Gadre, Shiori Sagawa, et~al.
\newblock Openflamingo: An open-source framework for training large autoregressive vision-language models.
\newblock \emph{arXiv preprint arXiv:2308.01390}, 2023.

\bibitem[Bai et~al.(2023{\natexlab{a}})Bai, Bai, Chu, Cui, Dang, Deng, Fan, Ge, Han, Huang, et~al.]{bai2023qwen}
Jinze Bai, Shuai Bai, Yunfei Chu, Zeyu Cui, Kai Dang, Xiaodong Deng, Yang Fan, Wenbin Ge, Yu Han, Fei Huang, et~al.
\newblock Qwen technical report.
\newblock \emph{arXiv preprint arXiv:2309.16609}, 2023{\natexlab{a}}.

\bibitem[Bai et~al.(2023{\natexlab{b}})Bai, Bai, Yang, Wang, Tan, Wang, Lin, Zhou, and Zhou]{bai2023qwenvl}
Jinze Bai, Shuai Bai, Shusheng Yang, Shijie Wang, Sinan Tan, Peng Wang, Junyang Lin, Chang Zhou, and Jingren Zhou.
\newblock Qwen-vl: A frontier large vision-language model with versatile abilities.
\newblock \emph{arXiv preprint arXiv:2308.12966}, 2023{\natexlab{b}}.

\bibitem[Bao et~al.(2022)Bao, Wang, Dong, and Wei]{bao2022vlbert}
Hangbo Bao, Wenhui Wang, Li Dong, and Furu Wei.
\newblock Vl-beit: Generative vision-language pretraining.
\newblock \emph{arXiv preprint arXiv:2206.01127}, 2022.

\bibitem[Barbaresi(2021)]{barbaresi-2021-Trafilatura}
Adrien Barbaresi.
\newblock Trafilatura: A web scraping library and command-line tool for text discovery and extraction.
\newblock In \emph{Proceedings of the 59th Annual Meeting of the Association for Computational Linguistics and the 11th International Joint Conference on Natural Language Processing: System Demonstrations}, pages 122--131, 2021.

\bibitem[Bi et~al.(2024)Bi, Chen, Chen, Chen, Dai, Deng, Ding, Dong, Du, Fu, et~al.]{bi2024deepseekllm}
Xiao Bi, Deli Chen, Guanting Chen, Shanhuang Chen, Damai Dai, Chengqi Deng, Honghui Ding, Kai Dong, Qiushi Du, Zhe Fu, et~al.
\newblock Deepseek llm: Scaling open-source language models with longtermism.
\newblock \emph{arXiv preprint arXiv:2401.02954}, 2024.

\bibitem[Blei et~al.(2003)Blei, Ng, and Jordan]{blei2003lda}
David~M Blei, Andrew~Y Ng, and Michael~I Jordan.
\newblock Latent dirichlet allocation.
\newblock \emph{JMLR}, 3\penalty0 (Jan):\penalty0 993--1022, 2003.

\bibitem[Bloom(1970)]{bloom1970space}
Burton~H Bloom.
\newblock Space/time trade-offs in hash coding with allowable errors.
\newblock \emph{Communications of the ACM}, 13\penalty0 (7):\penalty0 422--426, 1970.

\bibitem[Broder(1997)]{broder1997resemblance}
Andrei~Z Broder.
\newblock On the resemblance and containment of documents.
\newblock In \emph{Proceedings. Compression and Complexity of SEQUENCES 1997 (Cat. No. 97TB100171)}, pages 21--29. IEEE, 1997.

\bibitem[Brown et~al.(2020)Brown, Mann, Ryder, Subbiah, Kaplan, Dhariwal, Neelakantan, Shyam, Sastry, Askell, et~al.]{brown2020gpt3}
Tom Brown, Benjamin Mann, Nick Ryder, Melanie Subbiah, Jared~D Kaplan, Prafulla Dhariwal, Arvind Neelakantan, Pranav Shyam, Girish Sastry, Amanda Askell, et~al.
\newblock Language models are few-shot learners.
\newblock \emph{Adv. Neural Inform. Process. Syst.}, 2020.

\bibitem[Byeon et~al.(2022)Byeon, Park, Kim, Lee, Baek, and Kim]{kakaobrain2022coyo700m}
Minwoo Byeon, Beomhee Park, Haecheon Kim, Sungjun Lee, Woonhyuk Baek, and Saehoon Kim.
\newblock Coyo-700m: Image-text pair dataset.
\newblock \url{https://github.com/kakaobrain/coyo-dataset}, 2022.

\bibitem[Cai et~al.(2024)Cai, Cao, Chen, Chen, Chen, Chen, Chen, Chen, Chen, Chu, et~al.]{cai2024internlm2}
Zheng Cai, Maosong Cao, Haojiong Chen, Kai Chen, Keyu Chen, Xin Chen, Xun Chen, Zehui Chen, Zhi Chen, Pei Chu, et~al.
\newblock Internlm2 technical report.
\newblock \emph{arXiv preprint arXiv:2403.17297}, 2024.

\bibitem[Cao and Xiao(2022)]{cao2022geoqa_plus}
Jie Cao and Jing Xiao.
\newblock An augmented benchmark dataset for geometric question answering through dual parallel text encoding.
\newblock In \emph{COLING}, pages 1511--1520, 2022.

\bibitem[Changpinyo et~al.(2021)Changpinyo, Sharma, Ding, and Soricut]{changpinyo2021cc12m}
Soravit Changpinyo, Piyush Sharma, Nan Ding, and Radu Soricut.
\newblock Conceptual 12m: Pushing web-scale image-text pre-training to recognize long-tail visual concepts.
\newblock In \emph{CVPR}, 2021.

\bibitem[Chen et~al.(2024{\natexlab{a}})Chen, Chen, Zhang, Chen, Wu, Zhang, Chen, Li, Wan, and Wang]{chen2024allava}
Guiming~Hardy Chen, Shunian Chen, Ruifei Zhang, Junying Chen, Xiangbo Wu, Zhiyi Zhang, Zhihong Chen, Jianquan Li, Xiang Wan, and Benyou Wang.
\newblock Allava: Harnessing gpt4v-synthesized data for a lite vision-language model.
\newblock \emph{arXiv preprint arXiv:2402.11684}, 2024{\natexlab{a}}.

\bibitem[Chen et~al.(2023{\natexlab{a}})Chen, Zhang, Zeng, Zhang, Zhu, and Zhao]{chen2023shikra}
Keqin Chen, Zhao Zhang, Weili Zeng, Richong Zhang, Feng Zhu, and Rui Zhao.
\newblock Shikra: Unleashing multimodal llm's referential dialogue magic.
\newblock \emph{arXiv preprint arXiv:2306.15195}, 2023{\natexlab{a}}.

\bibitem[Chen et~al.(2023{\natexlab{b}})Chen, Li, Dong, Zhang, He, Wang, Zhao, and Lin]{chen2023sharegpt4v}
Lin Chen, Jisong Li, Xiaoyi Dong, Pan Zhang, Conghui He, Jiaqi Wang, Feng Zhao, and Dahua Lin.
\newblock Sharegpt4v: Improving large multi-modal models with better captions.
\newblock \emph{arXiv preprint arXiv:2311.12793}, 2023{\natexlab{b}}.

\bibitem[Chen et~al.(2015)Chen, Fang, Lin, Vedantam, Gupta, Doll{\'a}r, and Zitnick]{chen2015coco-caption}
Xinlei Chen, Hao Fang, Tsung-Yi Lin, Ramakrishna Vedantam, Saurabh Gupta, Piotr Doll{\'a}r, and C~Lawrence Zitnick.
\newblock Microsoft coco captions: Data collection and evaluation server.
\newblock \emph{arXiv preprint arXiv:1504.00325}, 2015.

\bibitem[Chen et~al.(2023{\natexlab{c}})Chen, Wu, Wang, Su, Chen, Xing, Muyan, Zhang, Zhu, Lu, et~al.]{chen2023internvl}
Zhe Chen, Jiannan Wu, Wenhai Wang, Weijie Su, Guo Chen, Sen Xing, Zhong Muyan, Qinglong Zhang, Xizhou Zhu, Lewei Lu, et~al.
\newblock Internvl: Scaling up vision foundation models and aligning for generic visual-linguistic tasks.
\newblock \emph{arXiv preprint arXiv:2312.14238}, 2023{\natexlab{c}}.

\bibitem[Chen et~al.(2024{\natexlab{b}})Chen, Wang, Tian, Ye, Gao, Cui, Tong, Hu, Luo, Ma, et~al.]{chen2024internvl1_5}
Zhe Chen, Weiyun Wang, Hao Tian, Shenglong Ye, Zhangwei Gao, Erfei Cui, Wenwen Tong, Kongzhi Hu, Jiapeng Luo, Zheng Ma, et~al.
\newblock How far are we to gpt-4v? closing the gap to commercial multimodal models with open-source suites.
\newblock \emph{arXiv preprint arXiv:2404.16821}, 2024{\natexlab{b}}.

\bibitem[Chiang et~al.(2023)Chiang, Li, Lin, Sheng, Wu, Zhang, Zheng, Zhuang, Zhuang, Gonzalez, Stoica, and Xing]{vicuna}
Wei-Lin Chiang, Zhuohan Li, Zi Lin, Ying Sheng, Zhanghao Wu, Hao Zhang, Lianmin Zheng, Siyuan Zhuang, Yonghao Zhuang, Joseph~E. Gonzalez, Ion Stoica, and Eric~P. Xing.
\newblock Vicuna: An open-source chatbot impressing gpt-4 with 90\%* chatgpt quality, 2023.

\bibitem[Chng et~al.(2019)Chng, Liu, Sun, Ng, Luo, Ni, Fang, Zhang, Han, Ding, et~al.]{chng2019art}
Chee~Kheng Chng, Yuliang Liu, Yipeng Sun, Chun~Chet Ng, Canjie Luo, Zihan Ni, ChuanMing Fang, Shuaitao Zhang, Junyu Han, Errui Ding, et~al.
\newblock Icdar2019 robust reading challenge on arbitrary-shaped text-rrc-art.
\newblock In \emph{ICDAR}, pages 1571--1576, 2019.

\bibitem[Clark and Gardner(2018)]{clark2017docqa}
Christopher Clark and Matt Gardner.
\newblock Simple and effective multi-paragraph reading comprehension.
\newblock In \emph{ACL}, pages 845--855, 2018.

\bibitem[{Common Crawl}(2007)]{commoncrawl}
{Common Crawl}.
\newblock Common crawl - open repository of web crawl data.
\newblock \url{https://commoncrawl.org/}, 2007.

\bibitem[Dai et~al.(2023)Dai, Li, Li, Huat, Zhao, Wang, Li, Fung, and Hoi]{instructblip}
Wenliang Dai, Junnan Li, Dongxu Li, AnthonyMeng Huat, Junqi Zhao, Weisheng Wang, Boyang Li, Pascale Fung, and Steven Hoi.
\newblock Instructblip: Towards general-purpose vision-language models with instruction tuning.
\newblock \emph{arXiv preprint arXiv:2305.06500}, 2023.

\bibitem[Devlin et~al.(2018)Devlin, Chang, Lee, and Toutanova]{devlin2018bert}
Jacob Devlin, Ming-Wei Chang, Kenton Lee, and Kristina Toutanova.
\newblock Bert: Pre-training of deep bidirectional transformers for language understanding.
\newblock \emph{NAACL-HLT}, 2018.

\bibitem[Dong et~al.(2023)Dong, Han, Peng, Qi, Ge, Yang, Zhao, Sun, Zhou, Wei, et~al.]{dong2023dreamllm}
Runpei Dong, Chunrui Han, Yuang Peng, Zekun Qi, Zheng Ge, Jinrong Yang, Liang Zhao, Jianjian Sun, Hongyu Zhou, Haoran Wei, et~al.
\newblock Dreamllm: Synergistic multimodal comprehension and creation.
\newblock \emph{arXiv preprint arXiv:2309.11499}, 2023.

\bibitem[Fang et~al.(2022)Fang, Wang, Xie, Sun, Wu, Wang, Huang, Wang, and Cao]{fang2022eva}
Yuxin Fang, Wen Wang, Binhui Xie, Quan Sun, Ledell Wu, Xinggang Wang, Tiejun Huang, Xinlong Wang, and Yue Cao.
\newblock Eva: Exploring the limits of masked visual representation learning at scale.
\newblock \emph{arXiv preprint arXiv:2211.07636}, 2022.

\bibitem[Fu et~al.(2023)Fu, Chen, Shen, Qin, Zhang, Lin, Qiu, Lin, Yang, Zheng, et~al.]{fu2023mme}
Chaoyou Fu, Peixian Chen, Yunhang Shen, Yulei Qin, Mengdan Zhang, Xu Lin, Zhenyu Qiu, Wei Lin, Jinrui Yang, Xiawu Zheng, et~al.
\newblock Mme: A comprehensive evaluation benchmark for multimodal large language models.
\newblock \emph{arXiv preprint arXiv:2306.13394}, 2023.

\bibitem[Gadre et~al.(2023)Gadre, Ilharco, Fang, Hayase, Smyrnis, Nguyen, Marten, Wortsman, Ghosh, Zhang, et~al.]{gadre2023datacomp}
Samir~Yitzhak Gadre, Gabriel Ilharco, Alex Fang, Jonathan Hayase, Georgios Smyrnis, Thao Nguyen, Ryan Marten, Mitchell Wortsman, Dhruba Ghosh, Jieyu Zhang, et~al.
\newblock Datacomp: In search of the next generation of multimodal datasets.
\newblock \emph{arXiv preprint arXiv:2304.14108}, 2023.

\bibitem[Ge et~al.(2024)Ge, Cheng, Wang, Yuan, Gao, Song, Song, Huang, and Zheng]{ge2024convllava}
Chunjiang Ge, Sijie Cheng, Ziming Wang, Jiale Yuan, Yuan Gao, Jun Song, Shiji Song, Gao Huang, and Bo Zheng.
\newblock Convllava: Hierarchical backbones as visual encoder for large multimodal models.
\newblock \emph{arXiv preprint arXiv:2405.15738}, 2024.

\bibitem[Goyal et~al.(2017)Goyal, Khot, Summers-Stay, Batra, and Parikh]{goyal2017vqav2}
Yash Goyal, Tejas Khot, Douglas Summers-Stay, Dhruv Batra, and Devi Parikh.
\newblock Making the v in vqa matter: Elevating the role of image understanding in visual question answering.
\newblock In \emph{CVPR}, 2017.

\bibitem[Gurari et~al.(2018)Gurari, Li, Stangl, Guo, Lin, Grauman, Luo, and Bigham]{gurari2018vizwiz}
Danna Gurari, Qing Li, Abigale~J Stangl, Anhong Guo, Chi Lin, Kristen Grauman, Jiebo Luo, and Jeffrey~P Bigham.
\newblock Vizwiz grand challenge: Answering visual questions from blind people.
\newblock In \emph{CVPR}, pages 3608--3617, 2018.

\bibitem[Hendrycks et~al.(2020)Hendrycks, Burns, Basart, Zou, Mazeika, Song, and Steinhardt]{hendrycks2020mmlu}
Dan Hendrycks, Collin Burns, Steven Basart, Andy Zou, Mantas Mazeika, Dawn Song, and Jacob Steinhardt.
\newblock Measuring massive multitask language understanding.
\newblock \emph{arXiv preprint arXiv:2009.03300}, 2020.

\bibitem[Huang et~al.(2023)Huang, Dong, Wang, Hao, Singhal, Ma, Lv, Cui, Mohammed, Liu, et~al.]{huang2023kosmos-1}
Shaohan Huang, Li Dong, Wenhui Wang, Yaru Hao, Saksham Singhal, Shuming Ma, Tengchao Lv, Lei Cui, Owais~Khan Mohammed, Qiang Liu, et~al.
\newblock Language is not all you need: Aligning perception with language models.
\newblock \emph{arXiv preprint arXiv:2302.14045}, 2023.

\bibitem[Hudson and Manning(2019)]{hudson2019gqa}
Drew~A Hudson and Christopher~D Manning.
\newblock Gqa: A new dataset for real-world visual reasoning and compositional question answering.
\newblock In \emph{CVPR}, 2019.

\bibitem[{IDEFICS}(2023)]{idefics2023}
{IDEFICS}.
\newblock Introducing idefics: An open reproduction of state-of-the-art visual language model.
\newblock \url{https://huggingface.co/blog/idefics}, 2023.

\bibitem[Ilharco et~al.(2021)Ilharco, Wortsman, Wightman, Gordon, Carlini, Taori, Dave, Shankar, Namkoong, Miller, Hajishirzi, Farhadi, and Schmidt]{openclip}
Gabriel Ilharco, Mitchell Wortsman, Ross Wightman, Cade Gordon, Nicholas Carlini, Rohan Taori, Achal Dave, Vaishaal Shankar, Hongseok Namkoong, John Miller, Hannaneh Hajishirzi, Ali Farhadi, and Ludwig Schmidt.
\newblock \emph{OpenCLIP}, 2021.

\bibitem[Jia et~al.(2021)Jia, Yang, Xia, Chen, Parekh, Pham, Le, Sung, Li, and Duerig]{jia2021align}
Chao Jia, Yinfei Yang, Ye Xia, Yi-Ting Chen, Zarana Parekh, Hieu Pham, Quoc Le, Yun-Hsuan Sung, Zhen Li, and Tom Duerig.
\newblock Scaling up visual and vision-language representation learning with noisy text supervision.
\newblock In \emph{ICML}, 2021.

\bibitem[Jimmycarter(2023)]{textocr_gpt4v_dataset}
Jimmycarter.
\newblock Textocr gpt-4v dataset.
\newblock \url{https://huggingface.co/datasets/jimmycarter/textocr-gpt4v}, 2023.

\bibitem[Jin et~al.(2023)Jin, Xu, Chen, Liao, Tan, Chen, Lei, Liu, Song, Lei, et~al.]{jin2023lavit}
Yang Jin, Kun Xu, Liwei Chen, Chao Liao, Jianchao Tan, Bin Chen, Chenyi Lei, An Liu, Chengru Song, Xiaoqiang Lei, et~al.
\newblock Unified language-vision pretraining with dynamic discrete visual tokenization.
\newblock \emph{arXiv preprint arXiv:2309.04669}, 2023.

\bibitem[Kafle et~al.(2018)Kafle, Price, Cohen, and Kanan]{kafle2018dvqa}
Kushal Kafle, Brian Price, Scott Cohen, and Christopher Kanan.
\newblock Dvqa: Understanding data visualizations via question answering.
\newblock In \emph{CVPR}, pages 5648--5656, 2018.

\bibitem[Kalkowski et~al.(2015)Kalkowski, Schulze, Dengel, and Borth]{yfcc15m}
Sebastian Kalkowski, Christian Schulze, Andreas Dengel, and Damian Borth.
\newblock Real-time analysis and visualization of the yfcc100m dataset.
\newblock In \emph{Proceedings of the 2015 workshop on community-organized multimodal mining: opportunities for novel solutions}, pages 25--30, 2015.

\bibitem[Kembhavi et~al.(2016)Kembhavi, Salvato, Kolve, Seo, Hajishirzi, and Farhadi]{kembhavi2016ai2d}
Aniruddha Kembhavi, Mike Salvato, Eric Kolve, Minjoon Seo, Hannaneh Hajishirzi, and Ali Farhadi.
\newblock A diagram is worth a dozen images.
\newblock In \emph{ECCV}, pages 235--251, 2016.

\bibitem[Kembhavi et~al.(2017)Kembhavi, Seo, Schwenk, Choi, Farhadi, and Hajishirzi]{kembhavi2017tqa}
Aniruddha Kembhavi, Minjoon Seo, Dustin Schwenk, Jonghyun Choi, Ali Farhadi, and Hannaneh Hajishirzi.
\newblock Are you smarter than a sixth grader? textbook question answering for multimodal machine comprehension.
\newblock In \emph{CVPR}, pages 4999--5007, 2017.

\bibitem[Kim et~al.(2022)Kim, Hong, Yim, Nam, Park, Yim, Hwang, Yun, Han, and Park]{kim2022synthdog}
Geewook Kim, Teakgyu Hong, Moonbin Yim, JeongYeon Nam, Jinyoung Park, Jinyeong Yim, Wonseok Hwang, Sangdoo Yun, Dongyoon Han, and Seunghyun Park.
\newblock Ocr-free document understanding transformer.
\newblock In \emph{ECCV}, 2022.

\bibitem[Krishna et~al.(2017)Krishna, Zhu, Groth, Johnson, Hata, Kravitz, Chen, Kalantidis, Li, Shamma, et~al.]{krishna2017visualgenome}
Ranjay Krishna, Yuke Zhu, Oliver Groth, Justin Johnson, Kenji Hata, Joshua Kravitz, Stephanie Chen, Yannis Kalantidis, Li-Jia Li, David~A Shamma, et~al.
\newblock Visual genome: Connecting language and vision using crowdsourced dense image annotations.
\newblock \emph{IJCV}, 123:\penalty0 32--73, 2017.

\bibitem[LAION(2023)]{laion_gpt4v_dataset}
LAION.
\newblock Gpt-4v dataset.
\newblock \url{https://huggingface.co/datasets/laion/gpt4v-dataset}, 2023.

\bibitem[Lauren{\c{c}}on et~al.(2024{\natexlab{a}})Lauren{\c{c}}on, Saulnier, Tronchon, Bekman, Singh, Lozhkov, Wang, Karamcheti, Rush, Kiela, et~al.]{laurenccon2024obelics}
Hugo Lauren{\c{c}}on, Lucile Saulnier, L{\'e}o Tronchon, Stas Bekman, Amanpreet Singh, Anton Lozhkov, Thomas Wang, Siddharth Karamcheti, Alexander Rush, Douwe Kiela, et~al.
\newblock Obelics: An open web-scale filtered dataset of interleaved image-text documents.
\newblock \emph{NeurIPS}, 36, 2024{\natexlab{a}}.

\bibitem[Lauren{\c{c}}on et~al.(2024{\natexlab{b}})Lauren{\c{c}}on, Tronchon, Cord, and Sanh]{laurenccon2024idefics2}
Hugo Lauren{\c{c}}on, L{\'e}o Tronchon, Matthieu Cord, and Victor Sanh.
\newblock What matters when building vision-language models?
\newblock \emph{arXiv preprint arXiv:2405.02246}, 2024{\natexlab{b}}.

\bibitem[Lerner et~al.(2022)Lerner, Ferret, Guinaudeau, Le~Borgne, Besan{\c{c}}on, Moreno, and Lov{\'o}n~Melgarejo]{lerner2022viquae}
Paul Lerner, Olivier Ferret, Camille Guinaudeau, Herv{\'e} Le~Borgne, Romaric Besan{\c{c}}on, Jos{\'e}~G Moreno, and Jes{\'u}s Lov{\'o}n~Melgarejo.
\newblock Viquae, a dataset for knowledge-based visual question answering about named entities.
\newblock In \emph{SIGIR}, pages 3108--3120, 2022.

\bibitem[Li et~al.(2023{\natexlab{a}})Li, Wang, Wang, Ge, Ge, and Shan]{li2023seed}
Bohao Li, Rui Wang, Guangzhi Wang, Yuying Ge, Yixiao Ge, and Ying Shan.
\newblock Seed-bench: Benchmarking multimodal llms with generative comprehension.
\newblock \emph{arXiv preprint arXiv:2307.16125}, 2023{\natexlab{a}}.

\bibitem[Li et~al.(2023{\natexlab{b}})Li, Li, Savarese, and Hoi]{li2023blip2}
Junnan Li, Dongxu Li, Silvio Savarese, and Steven Hoi.
\newblock Blip-2: Bootstrapping language-image pre-training with frozen image encoders and large language models.
\newblock \emph{arXiv preprint arXiv:2301.12597}, 2023{\natexlab{b}}.

\bibitem[Li et~al.(2023{\natexlab{c}})Li, He, Wang, Li, Wang, Luo, Wang, Wang, and Qiao]{li2023videochat}
KunChang Li, Yinan He, Yi Wang, Yizhuo Li, Wenhai Wang, Ping Luo, Yali Wang, Limin Wang, and Yu Qiao.
\newblock Videochat: Chat-centric video understanding.
\newblock \emph{arXiv preprint arXiv:2305.06355}, 2023{\natexlab{c}}.

\bibitem[Li et~al.(2023{\natexlab{d}})Li, Du, Zhou, Wang, Zhao, and Wen]{li2023pope}
Yifan Li, Yifan Du, Kun Zhou, Jinpeng Wang, Wayne~Xin Zhao, and Ji-Rong Wen.
\newblock Evaluating object hallucination in large vision-language models.
\newblock \emph{EMNLP}, 2023{\natexlab{d}}.

\bibitem[Li et~al.(2023{\natexlab{e}})Li, Wang, Stengel-Eskin, Kortylewski, Ma, Van~Durme, and Yuille]{li2023superclevr}
Zhuowan Li, Xingrui Wang, Elias Stengel-Eskin, Adam Kortylewski, Wufei Ma, Benjamin Van~Durme, and Alan~L Yuille.
\newblock Super-clevr: A virtual benchmark to diagnose domain robustness in visual reasoning.
\newblock In \emph{CVPR}, pages 14963--14973, 2023{\natexlab{e}}.

\bibitem[Lin et~al.(2023)Lin, Yin, Ping, Lu, Molchanov, Tao, Mao, Kautz, Shoeybi, and Han]{lin2023vila}
Ji Lin, Hongxu Yin, Wei Ping, Yao Lu, Pavlo Molchanov, Andrew Tao, Huizi Mao, Jan Kautz, Mohammad Shoeybi, and Song Han.
\newblock Vila: On pre-training for visual language models.
\newblock \emph{arXiv preprint arXiv:2312.07533}, 2023.

\bibitem[Lindstr{\"o}m and Abraham(2022)]{lindstrom2022clevrmath}
Adam~Dahlgren Lindstr{\"o}m and Savitha~Sam Abraham.
\newblock Clevr-math: A dataset for compositional language, visual and mathematical reasoning.
\newblock \emph{arXiv preprint arXiv:2208.05358}, 2022.

\bibitem[Liu et~al.(2023{\natexlab{a}})Liu, Emerson, and Collier]{liu2023vsr}
Fangyu Liu, Guy Emerson, and Nigel Collier.
\newblock Visual spatial reasoning.
\newblock \emph{TACL}, 11:\penalty0 635--651, 2023{\natexlab{a}}.

\bibitem[Liu et~al.(2023{\natexlab{b}})Liu, Lin, Li, Wang, Yacoob, and Wang]{liu2023lrv-instruction}
Fuxiao Liu, Kevin Lin, Linjie Li, Jianfeng Wang, Yaser Yacoob, and Lijuan Wang.
\newblock Aligning large multi-modal model with robust instruction tuning.
\newblock \emph{arXiv preprint arXiv:2306.14565}, 2023{\natexlab{b}}.

\bibitem[Liu et~al.(2023{\natexlab{c}})Liu, Wang, Yao, Chen, Song, Cho, Yacoob, and Yu]{liu2023mmcinst}
Fuxiao Liu, Xiaoyang Wang, Wenlin Yao, Jianshu Chen, Kaiqiang Song, Sangwoo Cho, Yaser Yacoob, and Dong Yu.
\newblock Mmc: Advancing multimodal chart understanding with large-scale instruction tuning.
\newblock \emph{arXiv preprint arXiv:2311.10774}, 2023{\natexlab{c}}.

\bibitem[Liu et~al.(2023{\natexlab{d}})Liu, Li, Li, and Lee]{liu2023llava1_5}
Haotian Liu, Chunyuan Li, Yuheng Li, and Yong~Jae Lee.
\newblock Improved baselines with visual instruction tuning.
\newblock \emph{arXiv preprint arXiv:2310.03744}, 2023{\natexlab{d}}.

\bibitem[Liu et~al.(2023{\natexlab{e}})Liu, Li, Wu, and Lee]{liu2023llava}
Haotian Liu, Chunyuan Li, Qingyang Wu, and Yong~Jae Lee.
\newblock Visual instruction tuning.
\newblock \emph{NeruPIS}, 2023{\natexlab{e}}.

\bibitem[Liu et~al.(2023{\natexlab{f}})Liu, Duan, Zhang, Li, Zhang, Zhao, Yuan, Wang, He, Liu, et~al.]{liu2023mmbench}
Yuan Liu, Haodong Duan, Yuanhan Zhang, Bo Li, Songyang Zhang, Wangbo Zhao, Yike Yuan, Jiaqi Wang, Conghui He, Ziwei Liu, et~al.
\newblock Mmbench: Is your multi-modal model an all-around player?
\newblock \emph{arXiv preprint arXiv:2307.06281}, 2023{\natexlab{f}}.

\bibitem[Liu et~al.(2023{\natexlab{g}})Liu, He, Wang, Wang, Wang, Chen, Zhang, Yang, Li, Yu, et~al.]{liu2023interngpt}
Zhaoyang Liu, Yinan He, Wenhai Wang, Weiyun Wang, Yi Wang, Shoufa Chen, Qinglong Zhang, Yang Yang, Qingyun Li, Jiashuo Yu, et~al.
\newblock Interngpt: Solving vision-centric tasks by interacting with chatbots beyond language.
\newblock \emph{arXiv preprint arXiv:2305.05662}, 2023{\natexlab{g}}.

\bibitem[Lu et~al.(2021)Lu, Gong, Jiang, Qiu, Huang, Liang, and Zhu]{lu2021geometry3k}
Pan Lu, Ran Gong, Shibiao Jiang, Liang Qiu, Siyuan Huang, Xiaodan Liang, and Song-Chun Zhu.
\newblock Inter-gps: Interpretable geometry problem solving with formal language and symbolic reasoning.
\newblock \emph{arXiv preprint arXiv:2105.04165}, 2021.

\bibitem[Lu et~al.(2022{\natexlab{a}})Lu, Mishra, Xia, Qiu, Chang, Zhu, Tafjord, Clark, and Kalyan]{lu2022scienceqa}
Pan Lu, Swaroop Mishra, Tanglin Xia, Liang Qiu, Kai-Wei Chang, Song-Chun Zhu, Oyvind Tafjord, Peter Clark, and Ashwin Kalyan.
\newblock Learn to explain: Multimodal reasoning via thought chains for science question answering.
\newblock \emph{NeurIPS}, 35:\penalty0 2507--2521, 2022{\natexlab{a}}.

\bibitem[Lu et~al.(2022{\natexlab{b}})Lu, Qiu, Chang, Wu, Zhu, Rajpurohit, Clark, and Kalyan]{lu2022tablemwp}
Pan Lu, Liang Qiu, Kai-Wei Chang, Ying~Nian Wu, Song-Chun Zhu, Tanmay Rajpurohit, Peter Clark, and Ashwin Kalyan.
\newblock Dynamic prompt learning via policy gradient for semi-structured mathematical reasoning.
\newblock \emph{arXiv preprint arXiv:2209.14610}, 2022{\natexlab{b}}.

\bibitem[Mao et~al.(2016)Mao, Huang, Toshev, Camburu, Yuille, and Murphy]{mao2016refcocog}
Junhua Mao, Jonathan Huang, Alexander Toshev, Oana Camburu, Alan~L Yuille, and Kevin Murphy.
\newblock Generation and comprehension of unambiguous object descriptions.
\newblock In \emph{CVPR}, 2016.

\bibitem[Marino et~al.(2019)Marino, Rastegari, Farhadi, and Mottaghi]{marino2019okvqa}
Kenneth Marino, Mohammad Rastegari, Ali Farhadi, and Roozbeh Mottaghi.
\newblock Ok-vqa: A visual question answering benchmark requiring external knowledge.
\newblock In \emph{CVPR}, pages 3195--3204, 2019.

\bibitem[Masry et~al.(2022)Masry, Do, Tan, Joty, and Hoque]{masry2022chartqa}
Ahmed Masry, Xuan~Long Do, Jia~Qing Tan, Shafiq Joty, and Enamul Hoque.
\newblock Chartqa: A benchmark for question answering about charts with visual and logical reasoning.
\newblock In \emph{ACL}, pages 2263--2279, 2022.

\bibitem[Mathew et~al.(2022)Mathew, Bagal, Tito, Karatzas, Valveny, and Jawahar]{mathew2022infographicvqa}
Minesh Mathew, Viraj Bagal, Rub{\`e}n Tito, Dimosthenis Karatzas, Ernest Valveny, and CV Jawahar.
\newblock Infographicvqa.
\newblock In \emph{WACV}, pages 1697--1706, 2022.

\bibitem[McKinzie et~al.(2024)McKinzie, Gan, Fauconnier, Dodge, Zhang, Dufter, Shah, Du, Peng, Weers, et~al.]{mckinzie2024mm1}
Brandon McKinzie, Zhe Gan, Jean-Philippe Fauconnier, Sam Dodge, Bowen Zhang, Philipp Dufter, Dhruti Shah, Xianzhi Du, Futang Peng, Floris Weers, et~al.
\newblock Mm1: Methods, analysis \& insights from multimodal llm pre-training.
\newblock \emph{arXiv preprint arXiv:2403.09611}, 2024.

\bibitem[Methani et~al.(2020)Methani, Ganguly, Khapra, and Kumar]{methani2020plotqa}
Nitesh Methani, Pritha Ganguly, Mitesh~M Khapra, and Pratyush Kumar.
\newblock Plotqa: Reasoning over scientific plots.
\newblock In \emph{WACV}, pages 1527--1536, 2020.

\bibitem[Miech et~al.(2019)Miech, Zhukov, Alayrac, Tapaswi, Laptev, and Sivic]{miech2019howto100m}
Antoine Miech, Dimitri Zhukov, Jean-Baptiste Alayrac, Makarand Tapaswi, Ivan Laptev, and Josef Sivic.
\newblock Howto100m: Learning a text-video embedding by watching hundred million narrated video clips.
\newblock In \emph{ICCV}, pages 2630--2640, 2019.

\bibitem[Mishra et~al.(2019)Mishra, Shekhar, Singh, and Chakraborty]{mishra2019ocrvqa}
Anand Mishra, Shashank Shekhar, Ajeet~Kumar Singh, and Anirban Chakraborty.
\newblock Ocr-vqa: Visual question answering by reading text in images.
\newblock In \emph{ICDAR}, pages 947--952, 2019.

\bibitem[OpenAI(2023)]{gpt4v}
OpenAI.
\newblock Gpt-4v(ision) system card.
\newblock \url{https://cdn.openai.com/papers/GPTV_System_Card.pdf}, 2023.

\bibitem[Peng et~al.(2023)Peng, Wang, Dong, Hao, Huang, Ma, and Wei]{peng2023kosmos2}
Zhiliang Peng, Wenhui Wang, Li Dong, Yaru Hao, Shaohan Huang, Shuming Ma, and Furu Wei.
\newblock Kosmos-2: Grounding multimodal large language models to the world.
\newblock \emph{arXiv preprint arXiv:2306.14824}, 2023.

\bibitem[Qiu et~al.(2024)Qiu, Lv, Jin, Wang, Ning, Yu, Zhang, Chu, Qu, Peng, et~al.]{qiu2024wanjuan}
Jiantao Qiu, Haijun Lv, Zhenjiang Jin, Rui Wang, Wenchang Ning, Jia Yu, ChaoBin Zhang, Pei Chu, Yuan Qu, Runyu Peng, et~al.
\newblock Wanjuan-cc: A safe and high-quality open-sourced english webtext dataset.
\newblock \emph{arXiv preprint arXiv:2402.19282}, 2024.

\bibitem[Radford et~al.(2021)Radford, Kim, Hallacy, Ramesh, Goh, Agarwal, Sastry, Askell, Mishkin, Clark, et~al.]{radford2021clip}
Alec Radford, Jong~Wook Kim, Chris Hallacy, Aditya Ramesh, Gabriel Goh, Sandhini Agarwal, Girish Sastry, Amanda Askell, Pamela Mishkin, Jack Clark, et~al.
\newblock Learning transferable visual models from natural language supervision.
\newblock In \emph{ICML}, 2021.

\bibitem[Rae et~al.(2021)Rae, Borgeaud, Cai, Millican, Hoffmann, Song, Aslanides, Henderson, Ring, Young, et~al.]{rae2021scaling}
Jack~W Rae, Sebastian Borgeaud, Trevor Cai, Katie Millican, Jordan Hoffmann, Francis Song, John Aslanides, Sarah Henderson, Roman Ring, Susannah Young, et~al.
\newblock Scaling language models: Methods, analysis \& insights from training gopher.
\newblock \emph{arXiv preprint arXiv:2112.11446}, 2021.

\bibitem[Raffel et~al.(2020)Raffel, Shazeer, Roberts, Lee, Narang, Matena, Zhou, Li, and Liu]{2020t5}
Colin Raffel, Noam Shazeer, Adam Roberts, Katherine Lee, Sharan Narang, Michael Matena, Yanqi Zhou, Wei Li, and Peter~J Liu.
\newblock Exploring the limits of transfer learning with a unified text-to-text transformer.
\newblock \emph{JMLR}, 21\penalty0 (140):\penalty0 1--67, 2020.

\bibitem[Reid et~al.(2024)Reid, Savinov, Teplyashin, Lepikhin, Lillicrap, Alayrac, Soricut, Lazaridou, Firat, Schrittwieser, et~al.]{reid2024gemini1_5}
Machel Reid, Nikolay Savinov, Denis Teplyashin, Dmitry Lepikhin, Timothy Lillicrap, Jean-baptiste Alayrac, Radu Soricut, Angeliki Lazaridou, Orhan Firat, Julian Schrittwieser, et~al.
\newblock Gemini 1.5: Unlocking multimodal understanding across millions of tokens of context.
\newblock \emph{arXiv preprint arXiv:2403.05530}, 2024.

\bibitem[Schuhman et~al.(2022)Schuhman, Köpf, Vencu, Coombes, and Beaumont]{laioncoco}
Christoph Schuhman, Andreas Köpf, Richard Vencu, Theo Coombes, and Romain Beaumont.
\newblock \emph{LAION COCO: 600M Synthetic Captions From LAION2B-EN}, 2022.

\bibitem[Schuhmann et~al.(2021)Schuhmann, Vencu, Beaumont, Kaczmarczyk, Mullis, Katta, Coombes, Jitsev, and Komatsuzaki]{schuhmann2021laion400m}
Christoph Schuhmann, Richard Vencu, Romain Beaumont, Robert Kaczmarczyk, Clayton Mullis, Aarush Katta, Theo Coombes, Jenia Jitsev, and Aran Komatsuzaki.
\newblock Laion-400m: Open dataset of clip-filtered 400 million image-text pairs.
\newblock \emph{arXiv preprint arXiv:2111.02114}, 2021.

\bibitem[Schuhmann et~al.(2022)Schuhmann, Beaumont, Vencu, Gordon, Wightman, Cherti, Coombes, Katta, Mullis, Wortsman, et~al.]{schuhmann2022laion5b}
Christoph Schuhmann, Romain Beaumont, Richard Vencu, Cade Gordon, Ross Wightman, Mehdi Cherti, Theo Coombes, Aarush Katta, Clayton Mullis, Mitchell Wortsman, et~al.
\newblock Laion-5b: An open large-scale dataset for training next generation image-text models.
\newblock \emph{Adv. Neural Inform. Process. Syst.}, 2022.

\bibitem[Schwenk et~al.(2022)Schwenk, Khandelwal, Clark, Marino, and Mottaghi]{schwenk2022aokvqa}
Dustin Schwenk, Apoorv Khandelwal, Christopher Clark, Kenneth Marino, and Roozbeh Mottaghi.
\newblock A-okvqa: A benchmark for visual question answering using world knowledge.
\newblock In \emph{ECCV}, pages 146--162, 2022.

\bibitem[Shah et~al.(2019)Shah, Mishra, Yadati, and Talukdar]{shah2019kvqa}
Sanket Shah, Anand Mishra, Naganand Yadati, and Partha~Pratim Talukdar.
\newblock Kvqa: Knowledge-aware visual question answering.
\newblock In \emph{AAAI}, pages 8876--8884, 2019.

\bibitem[Sharma et~al.(2018)Sharma, Ding, Goodman, and Soricut]{sharma2018cc3m}
Piyush Sharma, Nan Ding, Sebastian Goodman, and Radu Soricut.
\newblock Conceptual captions: A cleaned, hypernymed, image alt-text dataset for automatic image captioning.
\newblock In \emph{ACL}, 2018.

\bibitem[Shi et~al.(2017)Shi, Yao, Liao, Yang, Xu, Cui, Belongie, Lu, and Bai]{shi2017rctw17}
Baoguang Shi, Cong Yao, Minghui Liao, Mingkun Yang, Pei Xu, Linyan Cui, Serge Belongie, Shijian Lu, and Xiang Bai.
\newblock Icdar2017 competition on reading chinese text in the wild (rctw-17).
\newblock In \emph{ICDAR}, pages 1429--1434, 2017.

\bibitem[Sidorov et~al.(2020)Sidorov, Hu, Rohrbach, and Singh]{sidorov2020textcaps}
Oleksii Sidorov, Ronghang Hu, Marcus Rohrbach, and Amanpreet Singh.
\newblock Textcaps: a dataset for image captioning with reading comprehension.
\newblock In \emph{ECCV}, pages 742--758, 2020.

\bibitem[Singh et~al.(2019)Singh, Natarjan, Shah, Jiang, Chen, Parikh, and Rohrbach]{singh2019textvqa}
Amanpreet Singh, Vivek Natarjan, Meet Shah, Yu Jiang, Xinlei Chen, Devi Parikh, and Marcus Rohrbach.
\newblock Towards vqa models that can read.
\newblock In \emph{CVPR}, pages 8317--8326, 2019.

\bibitem[Sun et~al.(2023{\natexlab{a}})Sun, Cui, Zhang, Zhang, Yu, Luo, Wang, Rao, Liu, Huang, et~al.]{sun2023emu2}
Quan Sun, Yufeng Cui, Xiaosong Zhang, Fan Zhang, Qiying Yu, Zhengxiong Luo, Yueze Wang, Yongming Rao, Jingjing Liu, Tiejun Huang, et~al.
\newblock Generative multimodal models are in-context learners.
\newblock \emph{arXiv preprint arXiv:2312.13286}, 2023{\natexlab{a}}.

\bibitem[Sun et~al.(2023{\natexlab{b}})Sun, Fang, Wu, Wang, and Cao]{sun2023evaclip}
Quan Sun, Yuxin Fang, Ledell Wu, Xinlong Wang, and Yue Cao.
\newblock Eva-clip: Improved training techniques for clip at scale.
\newblock \emph{arXiv preprint arXiv:2303.15389}, 2023{\natexlab{b}}.

\bibitem[Sun et~al.(2023{\natexlab{c}})Sun, Yu, Cui, Zhang, Zhang, Wang, Gao, Liu, Huang, and Wang]{sun2023emu1}
Quan Sun, Qiying Yu, Yufeng Cui, Fan Zhang, Xiaosong Zhang, Yueze Wang, Hongcheng Gao, Jingjing Liu, Tiejun Huang, and Xinlong Wang.
\newblock Emu: Generative pretraining in multimodality.
\newblock In \emph{ICLR}, 2023{\natexlab{c}}.

\bibitem[Sun et~al.(2019)Sun, Ni, Chng, Liu, Luo, Ng, Han, Ding, Liu, Karatzas, et~al.]{sun2019lsvt}
Yipeng Sun, Zihan Ni, Chee-Kheng Chng, Yuliang Liu, Canjie Luo, Chun~Chet Ng, Junyu Han, Errui Ding, Jingtuo Liu, Dimosthenis Karatzas, et~al.
\newblock Icdar 2019 competition on large-scale street view text with partial labeling-rrc-lsvt.
\newblock In \emph{ICDAR}, pages 1557--1562, 2019.

\bibitem[Taori et~al.(2023)Taori, Gulrajani, Zhang, Dubois, Li, Guestrin, Liang, and Hashimoto]{taori2023alpaca}
Rohan Taori, Ishaan Gulrajani, Tianyi Zhang, Yann Dubois, Xuechen Li, Carlos Guestrin, Percy Liang, and Tatsunori~B Hashimoto.
\newblock Alpaca: A strong, replicable instruction-following model.
\newblock \emph{Stanford Center for Research on Foundation Models. https://crfm. stanford. edu/2023/03/13/alpaca. html}, 2023.

\bibitem[Team et~al.(2023)Team, Anil, Borgeaud, Wu, Alayrac, Yu, Soricut, Schalkwyk, Dai, Hauth, et~al.]{team2023gemini}
Gemini Team, Rohan Anil, Sebastian Borgeaud, Yonghui Wu, Jean-Baptiste Alayrac, Jiahui Yu, Radu Soricut, Johan Schalkwyk, Andrew~M Dai, Anja Hauth, et~al.
\newblock Gemini: a family of highly capable multimodal models.
\newblock \emph{arXiv preprint arXiv:2312.11805}, 2023.

\bibitem[Team(2023)]{2023internlm}
InternLM Team.
\newblock \emph{InternLM: A Multilingual Language Model with Progressively Enhanced Capabilities}, 2023.

\bibitem[Teknium(2023)]{OpenHermes2_5}
Teknium.
\newblock Openhermes 2.5: An open dataset of synthetic data for generalist llm assistants.
\newblock \url{https://huggingface.co/datasets/teknium/OpenHermes-2.5}, 2023.

\bibitem[Thomee et~al.(2016)Thomee, Shamma, Friedland, Elizalde, Ni, Poland, Borth, and Li]{thomee2016yfcc100m}
Bart Thomee, David~A Shamma, Gerald Friedland, Benjamin Elizalde, Karl Ni, Douglas Poland, Damian Borth, and Li-Jia Li.
\newblock Yfcc100m: The new data in multimedia research.
\newblock \emph{Communications of the ACM}, 59\penalty0 (2):\penalty0 64--73, 2016.

\bibitem[Tian et~al.(2024)Tian, Zhu, Xiong, Wang, Chen, Wang, Chen, Lu, Lu, Zhou, et~al.]{tian2024mminterleaved}
Changyao Tian, Xizhou Zhu, Yuwen Xiong, Weiyun Wang, Zhe Chen, Wenhai Wang, Yuntao Chen, Lewei Lu, Tong Lu, Jie Zhou, et~al.
\newblock Mm-interleaved: Interleaved image-text generative modeling via multi-modal feature synchronizer.
\newblock \emph{arXiv preprint arXiv:2401.10208}, 2024.

\bibitem[Touvron et~al.(2023{\natexlab{a}})Touvron, Lavril, Izacard, Martinet, Lachaux, Lacroix, Rozi{\`e}re, Goyal, Hambro, Azhar, et~al.]{touvron2023llama}
Hugo Touvron, Thibaut Lavril, Gautier Izacard, Xavier Martinet, Marie-Anne Lachaux, Timoth{\'e}e Lacroix, Baptiste Rozi{\`e}re, Naman Goyal, Eric Hambro, Faisal Azhar, et~al.
\newblock Llama: Open and efficient foundation language models.
\newblock \emph{arXiv preprint arXiv:2302.13971}, 2023{\natexlab{a}}.

\bibitem[Touvron et~al.(2023{\natexlab{b}})Touvron, Martin, Stone, Albert, Almahairi, Babaei, Bashlykov, Batra, Bhargava, Bhosale, et~al.]{touvron2023llama2}
Hugo Touvron, Louis Martin, Kevin Stone, Peter Albert, Amjad Almahairi, Yasmine Babaei, Nikolay Bashlykov, Soumya Batra, Prajjwal Bhargava, Shruti Bhosale, et~al.
\newblock Llama 2: Open foundation and fine-tuned chat models.
\newblock \emph{arXiv preprint arXiv:2307.09288}, 2023{\natexlab{b}}.

\bibitem[Van~der Maaten and Hinton(2008)]{van2008visualizing}
Laurens Van~der Maaten and Geoffrey Hinton.
\newblock Visualizing data using t-sne.
\newblock \emph{JMLR}, 9\penalty0 (11), 2008.

\bibitem[Vedantam et~al.(2015)Vedantam, Lawrence~Zitnick, and Parikh]{vedantam2015cider}
Ramakrishna Vedantam, C Lawrence~Zitnick, and Devi Parikh.
\newblock Cider: Consensus-based image description evaluation.
\newblock In \emph{CVPR}, 2015.

\bibitem[Veit et~al.(2016)Veit, Matera, Neumann, Matas, and Belongie]{veit2016cocotext}
Andreas Veit, Tomas Matera, Lukas Neumann, Jiri Matas, and Serge Belongie.
\newblock Coco-text: Dataset and benchmark for text detection and recognition in natural images.
\newblock \emph{arXiv preprint arXiv:1601.07140}, 2016.

\bibitem[Wang et~al.(2023{\natexlab{a}})Wang, Meng, Weng, He, Wu, and Jiang]{wang2023lvisinstruct4v}
Junke Wang, Lingchen Meng, Zejia Weng, Bo He, Zuxuan Wu, and Yu-Gang Jiang.
\newblock To see is to believe: Prompting gpt-4v for better visual instruction tuning.
\newblock \emph{arXiv preprint arXiv:2311.07574}, 2023{\natexlab{a}}.

\bibitem[Wang et~al.(2022)Wang, Bao, Dong, Bjorck, Peng, Liu, Aggarwal, Mohammed, Singhal, Som, et~al.]{wang2022beit3}
Wenhui Wang, Hangbo Bao, Li Dong, Johan Bjorck, Zhiliang Peng, Qiang Liu, Kriti Aggarwal, Owais~Khan Mohammed, Saksham Singhal, Subhojit Som, et~al.
\newblock Image as a foreign language: Beit pretraining for all vision and vision-language tasks.
\newblock \emph{arXiv preprint arXiv:2208.10442}, 2022.

\bibitem[Wang et~al.(2023{\natexlab{b}})Wang, Chen, Chen, Wu, Zhu, Zeng, Luo, Lu, Zhou, Qiao, et~al.]{wang2023visionllm}
Wenhai Wang, Zhe Chen, Xiaokang Chen, Jiannan Wu, Xizhou Zhu, Gang Zeng, Ping Luo, Tong Lu, Jie Zhou, Yu Qiao, et~al.
\newblock Visionllm: Large language model is also an open-ended decoder for vision-centric tasks.
\newblock \emph{NeurIPS}, 2023{\natexlab{b}}.

\bibitem[Wang et~al.(2024{\natexlab{a}})Wang, Ren, Luo, Li, Yan, Chen, Wang, Li, Lu, Zhu, et~al.]{wang2024allseeingv2}
Weiyun Wang, Yiming Ren, Haowen Luo, Tiantong Li, Chenxiang Yan, Zhe Chen, Wenhai Wang, Qingyun Li, Lewei Lu, Xizhou Zhu, et~al.
\newblock The all-seeing project v2: Towards general relation comprehension of the open world.
\newblock \emph{arXiv preprint arXiv:2402.19474}, 2024{\natexlab{a}}.

\bibitem[Wang et~al.(2024{\natexlab{b}})Wang, Shi, Li, Wang, Huang, Xing, Chen, Li, Zhu, Cao, et~al.]{wang2023allseeing}
Weiyun Wang, Min Shi, Qingyun Li, Wenhai Wang, Zhenhang Huang, Linjie Xing, Zhe Chen, Hao Li, Xizhou Zhu, Zhiguo Cao, et~al.
\newblock The all-seeing project: Towards panoptic visual recognition and understanding of the open world.
\newblock In \emph{ICLR}, 2024{\natexlab{b}}.

\bibitem[Wang et~al.(2020)Wang, Liu, Shen, Ng, Luo, Jin, Chan, Hengel, and Wang]{wang2020estvqa}
Xinyu Wang, Yuliang Liu, Chunhua Shen, Chun~Chet Ng, Canjie Luo, Lianwen Jin, Chee~Seng Chan, Anton van~den Hengel, and Liangwei Wang.
\newblock On the general value of evidence, and bilingual scene-text visual question answering.
\newblock In \emph{CVPR}, pages 10126--10135, 2020.

\bibitem[Wang et~al.(2023{\natexlab{c}})Wang, He, Li, Li, Yu, Ma, Li, Chen, Chen, Wang, et~al.]{wang2023internvid}
Yi Wang, Yinan He, Yizhuo Li, Kunchang Li, Jiashuo Yu, Xin Ma, Xinhao Li, Guo Chen, Xinyuan Chen, Yaohui Wang, et~al.
\newblock Internvid: A large-scale video-text dataset for multimodal understanding and generation.
\newblock \emph{arXiv preprint arXiv:2307.06942}, 2023{\natexlab{c}}.

\bibitem[Xue et~al.(2022)Xue, Hang, Zeng, Sun, Liu, Yang, Fu, and Guo]{xue2022hdvila}
Hongwei Xue, Tiankai Hang, Yanhong Zeng, Yuchong Sun, Bei Liu, Huan Yang, Jianlong Fu, and Baining Guo.
\newblock Advancing high-resolution video-language representation with large-scale video transcriptions.
\newblock In \emph{CVPR}, pages 5036--5045, 2022.

\bibitem[Yang et~al.(2022)Yang, Gan, Wang, Hu, Lu, Liu, and Wang]{yang2022rices}
Zhengyuan Yang, Zhe Gan, Jianfeng Wang, Xiaowei Hu, Yumao Lu, Zicheng Liu, and Lijuan Wang.
\newblock An empirical study of gpt-3 for few-shot knowledge-based vqa.
\newblock In \emph{AAAI}, pages 3081--3089, 2022.

\bibitem[Young et~al.(2014)Young, Lai, Hodosh, and Hockenmaier]{flickr30k}
Peter Young, Alice Lai, Micah Hodosh, and Julia Hockenmaier.
\newblock From image descriptions to visual denotations: New similarity metrics for semantic inference over event descriptions.
\newblock \emph{TACL}, 2:\penalty0 67--78, 2014.

\bibitem[Yu et~al.(2016)Yu, Poirson, Yang, Berg, and Berg]{yu2016refcoco}
Licheng Yu, Patrick Poirson, Shan Yang, Alexander~C Berg, and Tamara~L Berg.
\newblock Modeling context in referring expressions.
\newblock In \emph{Eur. Conf. Comput. Vis.}, 2016.

\bibitem[Yu et~al.(2023{\natexlab{a}})Yu, Jiang, Shi, Yu, Liu, Zhang, Kwok, Li, Weller, and Liu]{yu2023mathqa}
Longhui Yu, Weisen Jiang, Han Shi, Jincheng Yu, Zhengying Liu, Yu Zhang, James~T Kwok, Zhenguo Li, Adrian Weller, and Weiyang Liu.
\newblock Metamath: Bootstrap your own mathematical questions for large language models.
\newblock \emph{arXiv preprint arXiv:2309.12284}, 2023{\natexlab{a}}.

\bibitem[Yu et~al.(2023{\natexlab{b}})Yu, Yang, Li, Wang, Lin, Liu, Wang, and Wang]{yu2023mmvet}
Weihao Yu, Zhengyuan Yang, Linjie Li, Jianfeng Wang, Kevin Lin, Zicheng Liu, Xinchao Wang, and Lijuan Wang.
\newblock Mm-vet: Evaluating large multimodal models for integrated capabilities.
\newblock \emph{arXiv preprint arXiv:2308.02490}, 2023{\natexlab{b}}.

\bibitem[Yuan et~al.(2019)Yuan, Zhu, Xu, Li, Mu, and Hu]{yuan2019ctw}
Tai-Ling Yuan, Zhe Zhu, Kun Xu, Cheng-Jun Li, Tai-Jiang Mu, and Shi-Min Hu.
\newblock A large chinese text dataset in the wild.
\newblock \emph{Journal of Computer Science and Technology}, 34:\penalty0 509--521, 2019.

\bibitem[Zellers et~al.(2022)Zellers, Lu, Lu, Yu, Zhao, Salehi, Kusupati, Hessel, Farhadi, and Choi]{zellers2022merlot}
Rowan Zellers, Jiasen Lu, Ximing Lu, Youngjae Yu, Yanpeng Zhao, Mohammadreza Salehi, Aditya Kusupati, Jack Hessel, Ali Farhadi, and Yejin Choi.
\newblock Merlot reserve: Neural script knowledge through vision and language and sound.
\newblock In \emph{CVPR}, pages 16375--16387, 2022.

\bibitem[Zeng et~al.(2022)Zeng, Liu, Du, Wang, Lai, Ding, Yang, Xu, Zheng, Xia, et~al.]{zeng2022glm}
Aohan Zeng, Xiao Liu, Zhengxiao Du, Zihan Wang, Hanyu Lai, Ming Ding, Zhuoyi Yang, Yifan Xu, Wendi Zheng, Xiao Xia, et~al.
\newblock Glm-130b: An open bilingual pre-trained model.
\newblock \emph{ICLR}, 2022.

\bibitem[Zhai et~al.(2023)Zhai, Mustafa, Kolesnikov, and Beyer]{zhai2023siglip}
Xiaohua Zhai, Basil Mustafa, Alexander Kolesnikov, and Lucas Beyer.
\newblock Sigmoid loss for language image pre-training.
\newblock In \emph{ICCV}, pages 11975--11986, 2023.

\bibitem[Zhang et~al.(2024)Zhang, Du, Chen, Liang, Luo, Zheng, Zhu, Cheng, Xu, Guo, et~al.]{zhang2024cmmmu}
Ge Zhang, Xinrun Du, Bei Chen, Yiming Liang, Tongxu Luo, Tianyu Zheng, Kang Zhu, Yuyang Cheng, Chunpu Xu, Shuyue Guo, et~al.
\newblock Cmmmu: A chinese massive multi-discipline multimodal understanding benchmark.
\newblock \emph{arXiv preprint arXiv:2401.11944}, 2024.

\bibitem[Zhang et~al.(2019)Zhang, Zhou, Jiang, Song, Li, Zhou, Wang, Wang, Liao, Yang, et~al.]{zhang2019rects}
Rui Zhang, Yongsheng Zhou, Qianyi Jiang, Qi Song, Nan Li, Kai Zhou, Lei Wang, Dong Wang, Minghui Liao, Mingkun Yang, et~al.
\newblock Icdar 2019 robust reading challenge on reading chinese text on signboard.
\newblock In \emph{ICDAR}, pages 1577--1581, 2019.

\bibitem[Zhang et~al.(2023)Zhang, Zhang, Gu, Zhou, Lipka, Yang, and Sun]{zhang2023llavar}
Yanzhe Zhang, Ruiyi Zhang, Jiuxiang Gu, Yufan Zhou, Nedim Lipka, Diyi Yang, and Tong Sun.
\newblock Llavar: Enhanced visual instruction tuning for text-rich image understanding.
\newblock \emph{arXiv preprint arXiv:2306.17107}, 2023.

\bibitem[Zhao et~al.(2023)Zhao, Wu, and Huang]{zhao2023svit}
Bo Zhao, Boya Wu, and Tiejun Huang.
\newblock Svit: Scaling up visual instruction tuning.
\newblock \emph{arXiv preprint arXiv:2307.04087}, 2023.

\bibitem[Zheng et~al.(2024)Zheng, Chiang, Sheng, Zhuang, Wu, Zhuang, Lin, Li, Li, Xing, et~al.]{zheng2023vicuna}
Lianmin Zheng, Wei-Lin Chiang, Ying Sheng, Siyuan Zhuang, Zhanghao Wu, Yonghao Zhuang, Zi Lin, Zhuohan Li, Dacheng Li, Eric Xing, et~al.
\newblock Judging llm-as-a-judge with mt-bench and chatbot arena.
\newblock \emph{NeurIPS}, 36, 2024.

\bibitem[Zhu et~al.(2023{\natexlab{a}})Zhu, Chen, Shen, Li, and Elhoseiny]{zhu2023minigpt-4}
Deyao Zhu, Jun Chen, Xiaoqian Shen, Xiang Li, and Mohamed Elhoseiny.
\newblock Minigpt-4: Enhancing vision-language understanding with advanced large language models.
\newblock \emph{arXiv preprint arXiv:2304.10592}, 2023{\natexlab{a}}.

\bibitem[Zhu et~al.(2023{\natexlab{b}})Zhu, Ding, Ge, Ge, Zhao, Zhao, Wang, and Shan]{zhu2023vl_gpt}
Jinguo Zhu, Xiaohan Ding, Yixiao Ge, Yuying Ge, Sijie Zhao, Hengshuang Zhao, Xiaohua Wang, and Ying Shan.
\newblock Vl-gpt: A generative pre-trained transformer for vision and language understanding and generation.
\newblock \emph{arXiv preprint arXiv:2312.09251}, 2023{\natexlab{b}}.

\bibitem[Zhu et~al.(2024)Zhu, Hessel, Awadalla, Gadre, Dodge, Fang, Yu, Schmidt, Wang, and Choi]{zhu2024mmc4}
Wanrong Zhu, Jack Hessel, Anas Awadalla, Samir~Yitzhak Gadre, Jesse Dodge, Alex Fang, Youngjae Yu, Ludwig Schmidt, William~Yang Wang, and Yejin Choi.
\newblock Multimodal c4: An open, billion-scale corpus of images interleaved with text.
\newblock \emph{NeurIPS}, 36, 2024.

\bibitem[Zhu et~al.(2022)Zhu, Zhu, Li, Wu, Li, Wang, and Dai]{zhu2022uni_p}
Xizhou Zhu, Jinguo Zhu, Hao Li, Xiaoshi Wu, Hongsheng Li, Xiaohua Wang, and Jifeng Dai.
\newblock Uni-perceiver: Pre-training unified architecture for generic perception for zero-shot and few-shot tasks.
\newblock In \emph{CVPR}, 2022.

\end{thebibliography}
}

\section*{Checklist}

\begin{enumerate}

\item For all authors...
\begin{enumerate}
  \item Do the main claims made in the abstract and introduction accurately reflect the paper's contributions and scope?
    \answerYes{}
  \item Did you describe the limitations of your work?
    \answerYes{See Section~\ref{sec:conclusion}.}
  \item Did you discuss any potential negative societal impacts of your work?
    \answerYes{See Section~\ref{sec:conclusion}.}
  \item Have you read the ethics review guidelines and ensured that your paper conforms to them?
    \answerYes{}
\end{enumerate}

\item If you are including theoretical results...
\begin{enumerate}
  \item Did you state the full set of assumptions of all theoretical results?
    \answerNA{}
	\item Did you include complete proofs of all theoretical results?
    \answerNA{}
\end{enumerate}

\item If you ran experiments (e.g. for benchmarks)...
\begin{enumerate}
  \item Did you include the code, data, and instructions needed to reproduce the main experimental results (either in the supplemental material or as a URL)?
    \answerYes{This paper provides detailed descriptions of the experimental setup, training steps, and the datasets used. We will also release the code later.}
  \item Did you specify all the training details (e.g., data splits, hyperparameters, how they were chosen)?
    \answerYes{See Appendix \textcolor{red}{B}.}
	\item Did you report error bars (e.g., with respect to the random seed after running experiments multiple times)?
    \answerNo{Most experiments have stable results with little variance.}
	\item Did you include the total amount of compute and the type of resources used (e.g., type of GPUs, internal cluster, or cloud provider)?
    \answerYes{See Appendix \textcolor{red}{B}.}
\end{enumerate}

\item If you are using existing assets (e.g., code, data, models) or curating/releasing new assets...
\begin{enumerate}
  \item If your work uses existing assets, did you cite the creators?
    \answerYes{We mentioned these libraries we used.}
  \item Did you mention the license of the assets?
    \answerYes{We only used open-source libraries.}
  \item Did you include any new assets either in the supplemental material or as a URL?
    \answerNA{}
  \item Did you discuss whether and how consent was obtained from people whose data you're using/curating?
    \answerYes{See Appendix \textcolor{red}{A}.}
  \item Did you discuss whether the data you are using/curating contains personally identifiable information or offensive content?
    \answerYes{See Appendix \textcolor{red}{A}.}
\end{enumerate}

\item If you used crowdsourcing or conducted research with human subjects...
\begin{enumerate}
  \item Did you include the full text of instructions given to participants and screenshots, if applicable?
    \answerYes{}
    The participants are authors of this paper, who know the details of this project.
  \item Did you describe any potential participant risks, with links to Institutional Review Board (IRB) approvals, if applicable?
    \answerNA{}
  \item Did you include the estimated hourly wage paid to participants and the total amount spent on participant compensation?
    \answerNA{}
\end{enumerate}

\end{enumerate}

\clearpage
\appendix

\section{Dataset Information}

\subsection{Datasheet for \dsname dataset}

\label{sec:appendix-datasheet}

\subsubsection{Motivation}

\begin{enumerate}[label=Q\arabic*]

\item \textbf{For what purpose was the dataset created?} Was there a specific task in mind? Was there a specific gap that needed to be filled? Please provide a description.

\begin{itemize}
\item OmniCorpus was created to address the limitations of existing image-text interleaved datasets, specifically their scale and diversity. The dataset contains 10 billion-level image-text pairs, with the goal of enhancing multimodal large language models (MLLMs). Unlike previous datasets that often focus on English and text-centric sources, OmniCorpus includes a broad range of data from both English and non-English websites as well as video-centric platforms, providing a more diverse and comprehensive resource for training MLLMs. The dataset's flexibility in data formats (pure text corpus, image-text pairs, and interleaved data) aims to support various research applications in multimodal learning.
\end{itemize}

\item \textbf{Who created the dataset (e.g., which team, research group) and on behalf of which entity (e.g., company, institution, organization)?}

\begin{itemize}
\item Due to the restrictions of double-blind conditions, answers regarding this question will be updated in the camera-ready version of the paper.
\end{itemize}

\item \textbf{Who funded the creation of the dataset?} If there is an associated grant, please provide the name of the granter and the grant name and number.

\begin{itemize}
\item Due to the restrictions of double-blind conditions, answers regarding this question will be updated in the camera-ready version of the paper.
\end{itemize}

\item \textbf{Any other comments?}

\begin{itemize}
\item No.
\end{itemize}

\end{enumerate}

\subsubsection{Composition}

\begin{enumerate}[resume*]

\item \textbf{What do the instances that comprise the dataset represent (e.g., documents, photos, people, countries)?} \textit{Are there multiple types of instances (e.g., movies, users, and ratings; people and interactions between them; nodes and edges)? Please provide a description.}

\begin{itemize}
\item Each instance in {\dsname} represents an image-text interleaved document. These instances include a variety of image types and corresponding textual descriptions. The dataset is diverse, encompassing images and text from English and non-English websites, as well as video platforms. The data is structured in a streaming format that allows for various configurations, such as pure text corpora, image-text pairs, and interleaved sequences.
\end{itemize}

\item \textbf{How many instances are there in total (of each type, if appropriate)?}

\begin{itemize}
\item {\dsname} consists of 8.6 billion images, 1,696 billion text tokens, and 2.2 billion documents. The dataset is significantly larger and more diverse compared to previous multimodal datasets.
\end{itemize}

\item \textbf{Does the dataset contain all possible instances or is it a sample (not necessarily random) of instances from a larger set?} \textit{If the dataset is a sample, then what is the larger set? Is the sample representative of the larger set (e.g., geographic coverage)? If so, please describe how this representativeness was validated/verified. If it is not representative of the larger set, please describe why not (e.g., to cover a more diverse range of instances, because instances were withheld or unavailable).}

\begin{itemize}
\item {\dsname} is a sample from Common Crawl \cite{commoncrawl}, Chinese websites, YT-Temporal-1B~\cite{zellers2022merlot}, HD-VILA-100M~\cite{xue2022hdvila}, HowTo100M~\cite{miech2019howto100m}, and InternVid~\cite{wang2023internvid}. The data was filtered and processed to maintain high quality and relevance, though it may not capture every possible instance from the larger set of internet data.
\end{itemize}

\item \textbf{What data does each instance consist of?} \textit{“Raw” data (e.g., unprocessed text or images) or features? In either case, please provide a description.}

\begin{itemize}
\item Each instance consists of an interleaved sequence of images and text. The data includes raw image URLs, associated text, and metadata such as image dimensions, language, and various quality scores. The text can be captions, descriptions, or other types of annotations related to the images.
\end{itemize}

\item \textbf{Is there a label or target associated with each instance?} \textit{If so, please provide a description.}

\begin{itemize}
\item No, OmniCorpus does not provide specific labels or targets for each instance. The dataset is designed to be flexible and can be used for various tasks such as image recognition, captioning, and visual question answering, depending on the researcher's needs.
\end{itemize}

\item \textbf{Is any information missing from individual instances?} \textit{If so, please provide a description, explaining why this information is missing (e.g., because it was unavailable). This does not include intentionally removed information, but might include, e.g., redacted text.}

\begin{itemize}
\item No.
\end{itemize}

\item \textbf{Are relationships between individual instances made explicit (e.g., users' movie ratings, social network links)?} \textit{If so, please describe how these relationships are made explicit.}

\begin{itemize}
\item No.
\end{itemize}

\item \textbf{Are there recommended data splits (e.g., training, development/validation, testing)?} \textit{If so, please provide a description of these splits, explaining the rationale behind them.}

\begin{itemize}
\item No.
\end{itemize}

\item \textbf{Are there any errors, sources of noise, or redundancies in the dataset?} \textit{If so, please provide a description.}

\begin{itemize}
\item {\dsname} is generated through a data engine and may contain some noise or errors. However, efforts were made to filter and clean the data, including human feedback and filtering processes.
\end{itemize}

\item \textbf{Is the dataset self-contained, or does it link to or otherwise rely on external resources (e.g., websites, tweets, other datasets)?} \textit{If it links to or relies on external resources, a) are there guarantees that they will exist, and remain constant, over time; b) are there official archival versions of the complete dataset (i.e., including the external resources as they existed at the time the dataset was created); c) are there any restrictions (e.g., licenses, fees) associated with any of the external resources that might apply to a future user? Please provide descriptions of all external resources and any restrictions associated with them, as well as links or other access points, as appropriate.}

\begin{itemize}
\item The dataset relies on URLs to images hosted on the web. While the data was collected to be as stable as possible, there are no guarantees that all external resources will remain available indefinitely. The dataset includes URLs and annotations, but not the media content itself. Users must respect the original sources' licenses and restrictions when accessing the data.
\end{itemize}

\item \textbf{Does the dataset contain data that might be considered confidential (e.g., data that is protected by legal privilege or by doctor-patient confidentiality, data that includes the content of individuals’ non-public communications)?} \textit{If so, please provide a description.}

\begin{itemize}
\item No.
\end{itemize}

\item \textbf{Does the dataset contain data that, if viewed directly, might be offensive, insulting, threatening, or might otherwise cause anxiety?} \textit{If so, please describe why.}
\label{Q16}

\begin{itemize}
\item The dataset includes images and text from various internet sources, and despite filtering efforts, it may still contain content that some users might find offensive or distressing. However, a subset with higher scrutiny and manual verification is available to minimize exposure to such content.
\end{itemize}

\item \textbf{Does the dataset relate to people?} \textit{If not, you may skip the remaining questions in this section.}

\begin{itemize}
\item People may appear in images or be mentioned in text, but they are not the primary focus of the dataset.
\end{itemize}

\item \textbf{Does the dataset identify any subpopulations (e.g., by age, gender)?}

\begin{itemize}
\item The dataset does not explicitly identify subpopulations. Any such information would be incidental and not a primary attribute of the dataset.
\end{itemize}

\item \textbf{Is it possible to identify individuals (i.e., one or more natural persons), either directly or indirectly (i.e., in combination with other data) from the dataset?} \textit{If so, please describe how.}

\begin{itemize}
\item Yes, the dataset comes from the internet, containing a huge range of images with people.
\end{itemize}

\item \textbf{Does the dataset contain data that might be considered sensitive in any way (e.g., data that reveals racial or ethnic origins, sexual orientations, religious beliefs, political opinions or union memberships, or locations; financial or health data; biometric or genetic data; forms of government identification, such as social security numbers; criminal history)?} \textit{If so, please provide a description.}

\begin{itemize}
\item Yes, the dataset includes images and text from various internet sources, and despite filtering efforts, it may still contain content that some users might find sensitive. However, a subset with higher scrutiny and manual verification is available to minimize exposure to such content.
\end{itemize}

\item \textbf{Any other comments?}

\begin{itemize}
\item  No.
\end{itemize}

\end{enumerate}

\subsubsection{Collection Process}

\begin{enumerate}[resume*]

\item \textbf{How was the data associated with each instance acquired?} \textit{Was the data directly observable (e.g., raw text, movie ratings), reported by subjects (e.g., survey responses), or indirectly inferred/derived from other data (e.g., part-of-speech tags, model-based guesses for age or language)? If data was reported by subjects or indirectly inferred/derived from other data, was the data validated/verified? If so, please describe how.}

\begin{itemize}
\item The {\dsname} is directly observable, from Common Crawl~\cite{commoncrawl}, chinese websites, YT-Temporal-1B~\cite{zellers2022merlot}, HD-VILA-100M~\cite{xue2022hdvila}, HowTo100M~\cite{miech2019howto100m}, and InternVid~\cite{wang2023internvid}.
\end{itemize}

\item \textbf{What mechanisms or procedures were used to collect the data (e.g., hardware apparatus or sensor, manual human curation, software program, software API)?} \textit{How were these mechanisms or procedures validated?}

\begin{itemize}
\item We ran the data engine in Python, over 128 8-A100(80G) GPU machine, 30000 CPU machine and 3Gbps bandwidth. 
\end{itemize}

\item \textbf{If the dataset is a sample from a larger set, what was the sampling strategy (e.g., deterministic, probabilistic with specific sampling probabilities)?}

\begin{itemize}
\item {\dsname} is created based on the Common Crawl~\cite{commoncrawl} and YT-Temporal-1B~\cite{zellers2022merlot}, HD-VILA-100M~\cite{xue2022hdvila}, HowTo100M~\cite{miech2019howto100m}, and InternVid~\cite{wang2023internvid}.
\end{itemize}

\item \textbf{Who was involved in the data collection process (e.g., students, crowdworkers, contractors) and how were they compensated (e.g., how much were crowdworkers paid)?}

\begin{itemize}
\item Due to the restrictions of double-blind conditions, answers regarding this question will be updated in the camera-ready version of the paper.
\end{itemize}

\item \textbf{Over what timeframe was the data collected? Does this timeframe match the creation timeframe of the data associated with the instances (e.g., recent crawl of old news articles)?} \textit{If not, please describe the timeframe in which the data associated with the instances was created.}

\begin{itemize}
\item The data for the {\dsname} dataset was collected over a timeframe that encompasses multiple years, as it includes a vast and diverse range of sources such as Common Crawl, Chinese websites, and YouTube. This comprehensive collection effort aims to cover a wide spectrum of content types and contexts. The timeframe of the data collection does not necessarily match the creation timeframe of the data associated with the instances. For instance, the dataset includes recent crawls of older news articles and video frames extracted from previously published videos. This approach ensures the inclusion of both contemporary and historical content, thus providing a rich and varied dataset for training multimodal models.
\end{itemize}

\item \textbf{Were any ethical review processes conducted (e.g., by an institutional review board)?} \textit{If so, please provide a description of these review processes, including the outcomes, as well as a link or other access point to any supporting documentation.}

\begin{itemize}
\item We did not conduct a formal ethical review process via institutional review boards. However, we employed several filtering mechanisms
to try and remove instances that could be problematic.
\end{itemize}

\item \textbf{Does the dataset relate to people?} \textit{If not, you may skip the remaining questions in this section.}

\begin{itemize}
\item People might be present in the images and descriptions, although they are not the sole emphasis of the dataset. 
\end{itemize}

\item \textbf{Did you collect the data from the individuals in question directly, or obtain it via third parties or other sources (e.g., websites)?}

\begin{itemize}
\item The data for {\dsname} was obtained from third-party sources, including Common Crawl, Chinese websites, and YouTube, rather than collected directly from individuals.
\end{itemize}

\item \textbf{Were the individuals in question notified about the data collection?} \textit{If so, please describe (or show with screenshots or other information) how notice was provided, and provide a link or other access point to, or otherwise reproduce, the exact language of the notification itself.}
\label{Q30}

\begin{itemize}
\item Individuals were not notified about the data collection. Our dataset is built upon Common Crawl~\cite{commoncrawl}, chinese websites, YT-Temporal-1B~\cite{zellers2022merlot}, HD-VILA-100M~\cite{xue2022hdvila}, HowTo100M~\cite{miech2019howto100m}, and InternVid~\cite{wang2023internvid}, which only contains information that is publicly available on the Internet. The publishers of these information are usually aware that it will be made public to the world, but they may not be aware that it will be collected in this way.
\end{itemize}

\item \textbf{Did the individuals in question consent to the collection and use of their data?} \textit{If so, please describe (or show with screenshots or other information) how consent was requested and provided, and provide a link or other access point to, or otherwise reproduce, the exact language to which the individuals consented.}

\begin{itemize}
\item No. See \ref{Q30} for more details.
\end{itemize}

\item \textbf{If consent was obtained, were the consenting individuals provided with a mechanism to revoke their consent in the future or for certain uses?} \textit{If so, please provide a description, as well as a link or other access point to the mechanism (if appropriate).}

\begin{itemize}
\item Users can contact the research team of the {\dsname} for image(s) removal. Besides, users can contact us to remove any annotation in our proposed {\dsname}.
\end{itemize}

\item \textbf{Has an analysis of the potential impact of the dataset and its use on data subjects (e.g., a data protection impact analysis) been conducted?} \textit{If so, please provide a description of this analysis, including the outcomes, as well as a link or other access point to any supporting documentation.}

\begin{itemize}
\item No.
\end{itemize}

\item \textbf{Any other comments?}

\begin{itemize}
\item No.
\end{itemize}

\end{enumerate}

\subsubsection{Preprocessing, Cleaning, and/or Labeling}

\begin{enumerate}[resume*]

\item \textbf{Was any preprocessing/cleaning/labeling of the data done (e.g., discretization or bucketing, tokenization, part-of-speech tagging, SIFT feature extraction, removal of instances, processing of missing values)?} \textit{If so, please provide a description. If not, you may skip the remainder of the questions in this section.}

\begin{itemize}
\item Yes. The preprocessing involves several steps: main body extraction using an improved version of Trafilatura \cite{barbaresi-2021-Trafilatura}, preliminary text filtering employing strategies from Gopher~\cite{rae2021scaling} and C4~\cite{2020t5}, document deduplication using minihash values, image downloading and filtering according to MMC4~\cite{zhu2024mmc4} guidelines and LAION-5B~\cite{schuhmann2022laion5b}, detailed text filtering based on BERT~\cite{devlin2018bert} models, and human-feedback filtering to enhance data quality.
\end{itemize}

\item \textbf{Was the “raw” data saved in addition to the preprocessed/cleaned/labeled data (e.g., to support unanticipated future uses)?} \textit{If so, please provide a link or other access point to the “raw” data.}

\begin{itemize}
\item No.
\end{itemize}

\item \textbf{Is the software used to preprocess/clean/label the instances available?} \textit{If so, please provide a link or other access point.}

\begin{itemize}
\item Yes, the data collection code will be open-sourced and accessible from the dataset website.
\end{itemize}

\item \textbf{Any other comments?}

\begin{itemize}
\item No.
\end{itemize}

\end{enumerate}

\subsubsection{Uses}

\begin{enumerate}[resume*]

\item \textbf{Has the dataset been used for any tasks already?} \textit{If so, please provide a description.}

\begin{itemize}
    \item Yes, the OmniCorpus dataset has been used for training multimodal large language models (MLLMs), specifically demonstrating its effectiveness in tasks such as image captioning and visual question answering (VQA) .
\end{itemize}

\item \textbf{Is there a repository that links to any or all papers or systems that use the dataset?} \textit{If so, please provide a link or other access point.}

\begin{itemize}
\item  No.
\end{itemize}

\item \textbf{What (other) tasks could the dataset be used for?}

\begin{itemize}
\item The dataset could be used for a variety of vision-and-language (V\&L) tasks, such as image captioning, visual question answering, and other multimodal tasks that require the integration of visual and textual data.
\end{itemize}

\item \textbf{Is there anything about the composition of the dataset or the way it was collected and preprocessed/cleaned/labeled that might impact future uses?} \textit{For example, is there anything that a future user might need to know to avoid uses that could result in unfair treatment of individuals or groups (e.g., stereotyping, quality of service issues) or other undesirable harms (e.g., financial harms, legal risks) If so, please provide a description. Is there anything a future user could do to mitigate these undesirable harms?}

\begin{itemize}
\item Yes, the dataset includes data from diverse sources including non-English websites and video platforms, which enhances its diversity. However, the dataset also includes data from the internet which may contain biases or low-quality content. Measures have been taken to filter out low-quality and irrelevant content through human-feedback text filters.
\end{itemize}

\item \textbf{Are there tasks for which the dataset should not be used?} \textit{If so, please provide a description.}

\begin{itemize}
\item The dataset should only be used for non-commercial academic research due to potential biases and the need for careful curation to avoid harmful outcomes .
\end{itemize}

\item \textbf{Any other comments?}

\begin{itemize}
\item No.
\end{itemize}

\end{enumerate}

\subsubsection{Distribution}

\begin{enumerate}[resume*]

\item \textbf{Will the dataset be distributed to third parties outside of the entity (e.g., company, institution, organization) on behalf of which the dataset was created?} \textit{If so, please provide a description.}

\begin{itemize}
\item Yes, the dataset will be open-source.
\end{itemize}

\item \textbf{How will the dataset be distributed (e.g., tarball on website, API, GitHub)?} \textit{Does the dataset have a digital object identifier (DOI)?}

\begin{itemize}
\item The data link will be available through GitHub.
\end{itemize}

\item \textbf{When will the dataset be distributed?}

\begin{itemize}
\item 01/09/2024 and onward.
\end{itemize}

\item \textbf{Will the dataset be distributed under a copyright or other intellectual property (IP) license, and/or under applicable terms of use (ToU)?} \textit{If so, please describe this license and/or ToU, and provide a link or other access point to, or otherwise reproduce, any relevant licensing terms or ToU, as well as any fees associated with these restrictions.}

\begin{itemize}
\item Apache 2.0 license
\end{itemize}

\item \textbf{Have any third parties imposed IP-based or other restrictions on the data associated with the instances?} \textit{If so, please describe these restrictions, and provide a link or other access point to, or otherwise reproduce, any relevant licensing terms, as well as any fees associated with these restrictions.}

\begin{itemize}
\item {\dsname} owns the metadata and release as Apache 2.0 license.
\item We do not own the copyright of the images.
\end{itemize}

\item \textbf{Do any export controls or other regulatory restrictions apply to the dataset or to individual instances?} \textit{If so, please describe these restrictions, and provide a link or other access point to, or otherwise reproduce, any supporting documentation.}

\begin{itemize}
\item No.
\end{itemize}

\item \textbf{Any other comments?}

\begin{itemize}
\item No.
\end{itemize}

\end{enumerate}

\subsubsection{Maintenance}

\begin{enumerate}[resume*]

\item \textbf{Who will be supporting/hosting/maintaining the dataset?}

\begin{itemize}
\item Due to the restrictions of double-blind conditions, answers regarding this question will be updated in the camera-ready version of the paper.
\end{itemize}

\item \textbf{How can the owner/curator/manager of the dataset be contacted (e.g., email address)?}

\begin{itemize}
\item Due to the restrictions of double-blind conditions, answers regarding this question will be updated in the camera-ready version of the paper.
\end{itemize}

\item \textbf{Is there an erratum?} \textit{If so, please provide a link or other access point.}

\begin{itemize}
\item No.
\end{itemize}

\item \textbf{Will the dataset be updated (e.g., to correct labeling errors, add new instances, delete instances)?} \textit{If so, please describe how often, by whom, and how updates will be communicated to users (e.g., mailing list, GitHub)?}

\begin{itemize}
\item No. However, specific samples can be removed on request.
\end{itemize}

\item \textbf{If the dataset relates to people, are there applicable limits on the retention of the data associated with the instances (e.g., were individuals in question told that their data would be retained for a fixed period of time and then deleted)?} \textit{If so, please describe these limits and explain how they will be enforced.}

\begin{itemize}
\item People may contact us to add specific samples to a blacklist.
\end{itemize}

\item \textbf{Will older versions of the dataset continue to be supported/hosted/maintained?} \textit{If so, please describe how. If not, please describe how its obsolescence will be communicated to users.}

\begin{itemize}
\item We will only support and maintain the latest version at all times, and a new version release of {\dsname} will automatically deprecate its previous version.
\end{itemize}

\item \textbf{If others want to extend/augment/build on/contribute to the dataset, is there a mechanism for them to do so?} \textit{If so, please provide a description. Will these contributions be validated/verified? If so, please describe how. If not, why not? Is there a process for communicating/distributing these contributions to other users? If so, please provide a description.}

\begin{itemize}
\item We welcome any contributions to {\dsname}, and we will announce updates regarding dataset extensions on GitHub.
However, contributors must demonstrate the quality and harmlessness of the extended data annotations; otherwise, we will not accept these extensions.
\end{itemize}

\item \textbf{Any other comments?}

\begin{itemize}
\item No.
\end{itemize}

\end{enumerate}

\subsection{Ethical discussion}

During the development of the OmniCorpus dataset, several ethical considerations were taken into account to ensure responsible data collection, usage, and sharing.

The OmniCorpus dataset comprises a vast collection of multimodal data, including images and text from various sources. Given the scale of data collection, it was impractical to obtain explicit consent from all content creators. However, efforts were made to respect the choices of content creators by removing opted-out images. This approach, while not exhaustive, reflects a commitment to respecting individual privacy and consent as much as possible.

To mitigate the inclusion of undesirable content, a rigorous filtering process was implemented, which aimed to exclude pornographic content and other potentially harmful material. Despite these efforts, the nature of web-crawled data means some inappropriate content might still be present. Continuous monitoring and updating of the filtering mechanisms are necessary to improve the dataset's quality and safety.

Datasets of this scale often inherit biases present in the source data. These biases can manifest in various forms, including the under-representation of certain demographics or the reinforcement of harmful stereotypes. Acknowledging this, the OmniCorpus project incorporated measures to identify and mitigate biases. 

Transparency in data collection and processing is crucial for ethical research. The OmniCorpus dataset is accompanied by extensive documentation detailing the data sources, filtering processes, and known limitations. This transparency allows users to understand the dataset's composition and make informed decisions about its use. Additionally, an interactive visualization tool was developed to facilitate the exploration and inspection of the dataset, promoting transparency and accessibility.

The release of the OmniCorpus dataset is intended to advance research in multimodal machine learning by providing a robust and diverse data foundation. However, the potential negative societal impacts, such as misuse of data or reinforcement of biases, are acknowledged. By striking a balance between the benefits of large-scale data availability and the risks associated with it, the project aims to contribute positively to the field while remaining vigilant about ethical considerations.

The ethical challenges associated with large-scale datasets are ongoing. The OmniCorpus project is committed to continuously improving its ethical standards by updating filtering techniques, enhancing bias mitigation strategies, and maintaining transparency in all aspects of the dataset's lifecycle. Engaging with the research community and stakeholders to address ethical concerns collaboratively is also a priority.

By addressing these ethical considerations, the OmniCorpus project strives to set a standard for responsible data handling and usage in the realm of multimodal machine learning research.

\section{Supplementary Experiment Details}

\subsection{Evaluation Details}

We evaluate the pre-trained models on four VQA benchmarks (including OKVQA~\cite{marino2019okvqa}, TextVQA \cite{singh2019textvqa}, VQAv2~\cite{goyal2017vqav2}, and VizWiz~\cite{gurari2018vizwiz}) and two image captioning benchmarks (including COCO Caption~\cite{chen2015coco-caption} and Flickr30K Caption~\cite{flickr30k}). Since the baseline models in ablation experiments are based on LLaVA-1.5~\cite{liu2023llava1_5}, we support RICES-based few-shot prompting~\cite{yang2022rices} for the open-source evaluation tools of LLaVA-1.5, which do not post-process the response and use OCR tokens for TextVQA. When comparing with state-of-the-art MLLMs pre-trained with image-text interleaved data (in Table~\ref{tab:res_final}), we adapt our model to the open-source evaluation tools of OpenFlamingo~\cite{awadalla2023openflamingo}, which sample few-shot examples randomly. For both settings, we provide few-shot examples in the chatting history of multi-round conversations. The formats of few-shot prompting for VQA and image captioning are provided in Table~\ref{tab:eval_prompt}.

\begin{table*}[h]\centering
\small
\vspace{-2mm}
\begin{minipage}{0.99\columnwidth}
    \caption{
        \textbf{The formats of few-shot prompting for VQA and image captioning.}
        The demonstrated template is from Vicuna~\cite{vicuna}. 
        Only one-shot situations are illustrated here; in practice, the number of turns varies based on the number of shots. 
        $\Xmat_{\texttt{system-message}}$ indicates the system message. 
        The rest ${\bf V}$, ${\bf X}$, and ${\bf Y}$ represent the tokens for the image, prompt, and response for an example or a test sample, respectively. 
        \texttt{<STOP>} represents stop indicators. 
        The {\color{mygreen} green tokens} are the expected responses.
    }
    \label{tab:eval_prompt}
    \vspace{-1mm}
    \centering
    \begin{tcolorbox} 
        \raggedright
        \small
        \hspace{-6mm}
    
        \textbf{VQA Prompt:} \\ \vspace{1mm}
        $\Xmat_{\texttt{system-message}}$ \texttt{<STOP>} \\ \vspace{1mm}
        $\texttt{Human}: \Vmat_{\texttt{shot}}^1 \ \ \Xmat_{\texttt{shot}}^1 \ \ \texttt{Answer the question using a single word or phrase.}$  \texttt{<STOP>} \\ \vspace{1mm}
        $\texttt{Assistant}: \Ymat_{\texttt{shot}}^1$  \texttt{<STOP>} \\ \vspace{1mm}
        $\cdots$ \\
        $\texttt{Human}: \Vmat_{\texttt{test}} \ \ \Xmat_{\texttt{test}} \ \ \texttt{Answer the question using a single word or phrase.}$  \texttt{<STOP>} \\ \vspace{1mm}
        $\texttt{Assistant}$: 
        \PredSty{$\Ymat_{\texttt{response}}$} 
        \PredSty{\texttt{<STOP>}} \\  
    
        \vspace{1mm}
        \rule{\linewidth}{0.2mm} \\
        \vspace{1mm}
    
        \textbf{Image Captioning Prompt:} \\ \vspace{1mm}
        $\Xmat_{\texttt{system-message}}$ \texttt{<STOP>} \\ \vspace{1mm}
        $\texttt{Human}: \Vmat_{\texttt{shot}}^1 \ \ \texttt{Provide a one-sentence caption for the provided image.}$  \texttt{<STOP>} \\ \vspace{1mm}
        $\texttt{Assistant}: \Ymat_{\texttt{shot}}^1$  \texttt{<STOP>} \\ \vspace{1mm}
        $\cdots$ \\
        $\texttt{Human}: \Vmat_{\texttt{test}} \ \ \texttt{Provide a one-sentence caption for the provided image.}$  \texttt{<STOP>} \\ \vspace{1mm}
        $\texttt{Assistant}$: 
        \PredSty{$\Ymat_{\texttt{response}}$} 
        \PredSty{\texttt{<STOP>}} \\  
    
    \end{tcolorbox}
\end{minipage}
\end{table*}

\subsection{Training Details}

We build the baseline models based on the LLaVA-1.5~\cite{liu2023llava1_5}. The models in ablation studies employ CLIP-ViT-L-336px~\cite{radford2021clip} and Vicuna-1.5-7B~\cite{zheng2023vicuna} as the vision encoder and the LLM, respectively. 
For the final model in Table~\ref{tab:res_final}, we replace them with InternViT-300M-448px~\cite{chen2023internvl} and InternLM2-7B~\cite{2023internlm}. 
Additionally, we employ a two-layer MLP pre-aligned with captioning data as introduced in LLaVA-1.5. 
During the pre-training, we freeze the vision encoder and update the parameters of the MLP projector and the LLM. 
We train the models with 1 million image-text interleaved documents on 16 80GB A100 GPUs for about one day.

\subsection{SFT Experiment}

\begin{table}[t]
\renewcommand{\arraystretch}{0.85}
\setlength\tabcolsep{1.25mm}
\setlength{\belowcaptionskip}{1.2mm}
\scriptsize
\centering
\caption{
\textbf{Results on 12 general visual-language benchmarks}.
Benchmark names are abbreviated due to space limits. VQA-v2~\cite{goyal2017vqav2}; GQA~\cite{hudson2019gqa}; VizWiz~\cite{gurari2018vizwiz}; SQA$^\text{I}$: ScienceQA-IMG~\cite{lu2022scienceqa}; VQA$^\text{T}$: TextVQA~\cite{singh2019textvqa}; POPE~\cite{li2023pope}; MME~\cite{fu2023mme}; MMB: MMBench~\cite{liu2023mmbench}; MMB$^\text{CN}$: MMBench-Chinese~\cite{liu2023mmbench}; SEED: SEED-Bench~\cite{li2023seed}; LLaVA$^\text{W}$: LLaVA-Bench (In-the-Wild)~\cite{liu2023llava}; MM-Vet~\cite{yu2023mmvet}. $^*$The training images of the datasets are observed during training.
The best performances are marked \textbf{bold}.
}
\label{tab:res_final_sft}
\vspace{1.5mm}
\begin{tabular}{l|llclc|ccccccc}
\toprule
Model                                    & VQA$^\text{v2}$ & GQA           & VizWiz        & SQA$^\text{I}$ & VQA$^\text{T}$ & POPE          & MME             & MMB           & MMB$^\text{CN}$ & SEED & LLaVA$^\text{W}$ & MM-Vet \\
\midrule
BLIP-2~\cite{li2023blip2}                & 41.0            & 41.0          & 19.6          & 61.0           & 42.5           & 85.3          & 1293.8          & $-$           & $-$             & 46.4 & 38.1             & 22.4          \\
InstructBLIP-7B~\cite{instructblip}      & $-$             & 49.2          & 34.5          & 60.5           & 50.1           & $-$           & $-$             & 36.0          & 23.7            & 53.4 & 60.9             & 26.2          \\
InstructBLIP-13B~\cite{instructblip}     & $-$             & 49.5          & 33.4          & 63.1           & 50.7           & 78.9          & 1212.8          & $-$           & $-$             & -    & 58.2             & 25.6          \\
Shikra~\cite{chen2023shikra}             & 77.4*           & $-$           & $-$           & $-$            & $-$            & $-$           & $-$             & 58.8          & $-$             & -    & $-$              & $-$           \\
IDEFICS-9B~\cite{idefics2023}            & 50.9            & 38.4          & 35.5          & $-$            & 25.9           & $-$           & $-$             & 48.2          & 25.2            & -    & $-$              & $-$           \\
IDEFICS-80B~\cite{idefics2023}           & 60.0            & 45.2          & 36.0          & $-$            & 30.9           & $-$           & $-$             & 54.5          & 38.1            & -    & $-$              & $-$           \\
Qwen-VL~\cite{bai2023qwenvl}             & 78.8*           & 59.3*         & 35.2          & 67.1           & 63.8           & $-$           & $-$             & 38.2          & 7.4             & 56.3 & $-$              & $-$           \\
Qwen-VL-Chat~\cite{bai2023qwenvl}        & 78.2*           & 57.5*         & 38.9          & 68.2           & 61.5           & $-$           & 1487.5          & 60.6          & 56.7            & 58.2 & $-$              & $-$           \\
LLaVA-1.5-7B~\cite{liu2023llava1_5}      & 78.5*           & 62.0*         & 50.0          & 66.8           & 58.2           & 85.9          & 1510.7          & 64.3          & 58.3            & 58.6 & 63.4             & 30.5          \\
InternVL-Chat~\cite{chen2023internvl}    & 79.3*           & \textbf{62.9}* & 52.5          & 66.2           & 57.0           & 86.4          & 1525.1          & 64.6          & 57.6            & 60.6 & 65.9             & 30.9          \\
VILA-7B~\cite{lin2023vila}               & 79.9*           & 62.3*         & \textbf{57.8} & 68.2           & 64.4           & 85.5          & 1533.0          & 68.9          & 61.7            & 61.1 & 69.7             & 34.9          \\
LLaVA-NeXT-7B~\cite{liu2023llava1_5}     & \textbf{81.8}*           & 64.2*           & 57.6           & 70.1           & 64.9           & \textbf{86.5} & 1519.0          & 67.4          & $-$             & $-$  & \textbf{81.6}    & \textbf{43.9} \\ 
\rowcolor{mygray}
Ours-7B                                  & 81.2*            & 61.7*        & 57.0           & \textbf{91.8}*           & \textbf{65.2}           & 85.4          & \textbf{1602.3}  & \textbf{76.5}          & \textbf{75.4}           & \textbf{65.6} & 72.1             & 41.3           \\ 
\bottomrule
\end{tabular}
\end{table}

\begin{table}[t]\scriptsize
\renewcommand{\arraystretch}{1.1}
\centering
\setlength{\tabcolsep}{2.9mm} 
\caption{\textbf{Summary of datasets used in the SFT experiment.} 
To further validate the effectiveness of our image-text interleaved pre-training, we followed the approach of LLaVA-1.5~\cite{liu2023llava1_5}, MM1~\cite{mckinzie2024mm1}, and InternVL-1.5~\cite{chen2024internvl1_5} to collect approximately 3.3M SFT examples from a diverse set of datasets.}
\vspace{1em}
\begin{tabular}{l|p{11.5cm}}
    \toprule
     Task & Dataset \\
    \midrule
Captioning                    & TextCaps~\cite{sidorov2020textcaps}, ShareGPT4V~\cite{chen2023sharegpt4v}                       \\
\rowcolor{gray!15}
General VQA                           & VQAv2~\cite{goyal2017vqav2}, GQA~\cite{hudson2019gqa}, OKVQA~\cite{marino2019okvqa}, VSR~\cite{liu2023vsr}, KVQA~\cite{shah2019kvqa}, A-OKVQA~\cite{schwenk2022aokvqa}, ViQuAE~\cite{lerner2022viquae}            \\
Science      & AI2D~\cite{kembhavi2016ai2d}, ScienceQA~\cite{lu2022scienceqa}, TQA~\cite{kembhavi2017tqa}      \\
\rowcolor{gray!15}
Chart        & ChartQA~\cite{masry2022chartqa}, MMC-Inst~\cite{liu2023mmcinst}, DVQA~\cite{kafle2018dvqa}, PlotQA~\cite{methani2020plotqa}, LRV-Instruction~\cite{liu2023lrv-instruction}                       \\
Mathematics                   & GeoQA+~\cite{cao2022geoqa_plus}, TabMWP~\cite{lu2022tablemwp}, MathQA~\cite{yu2023mathqa}, CLEVR-Math/Super~\cite{lindstrom2022clevrmath, li2023superclevr}, Geometry3K~\cite{lu2021geometry3k} \\
\rowcolor{gray!15}
OCR                           & OCRVQA~\cite{mishra2019ocrvqa}, InfoVQA~\cite{mathew2022infographicvqa}, TextVQA~\cite{singh2019textvqa}, ArT~\cite{chng2019art}, COCO-Text~\cite{veit2016cocotext}, CTW~\cite{yuan2019ctw}, LSVT~\cite{sun2019lsvt}, RCTW-17~\cite{shi2017rctw17}, ReCTs~\cite{zhang2019rects},  SynthDoG~\cite{kim2022synthdog}, LLaVAR~\cite{zhang2023llavar}, DocVQA~\cite{clark2017docqa}                                     \\
Grounding                     & RefCOCO/+/g~\cite{yu2016refcoco,mao2016refcocog}, Visual Genome~\cite{krishna2017visualgenome}                            \\
\rowcolor{gray!15}
Conversation             & LLaVA-150K~\cite{liu2023llava}, LVIS-Instruct4V~\cite{wang2023lvisinstruct4v}, ALLaVA~\cite{chen2024allava}, Laion-GPT4V~\cite{laion_gpt4v_dataset}, TextOCR-GPT4V~\cite{textocr_gpt4v_dataset}, SVIT~\cite{zhao2023svit}                            \\
\multirow{-1}{*}{Text-only}   & OpenHermes2.5~\cite{OpenHermes2_5}, Alpaca-GPT4~\cite{taori2023alpaca}, ShareGPT~\cite{zheng2023vicuna}   \\
\bottomrule
                        
\end{tabular}

\label{tab:sft_data}
\end{table}

To further validate the effectiveness of our image-text interleaved pre-training, we followed the approach of LLaVA-1.5~\cite{liu2023llava1_5}, MM1~\cite{mckinzie2024mm1}, and InternVL-1.5~\cite{chen2024internvl1_5} to collect approximately 3.3M SFT examples from a diverse set of datasets, as shown in Table~\ref{tab:sft_data}. 
These datasets are formatted into the instruction-following format, the same as LLaVA-1.5. 
During SFT, we train the entire model, including the vision encoder, MLP projector, and LLM. 
We compare our model with state-of-the-art MLLMs, as presented in Table~\ref{tab:res_final_sft}.
The results demonstrate that our image-text interleaved pre-training significantly enhances the model's performance. 

\section{Details of the Data Engine}

\subsection{Advantages of our pipeline sequence}

In this section, we aim to demonstrate that our pipeline sequence is the fastest. We assume we have  10,000 CPU resources, 3 Gbps bandwidth, and 1,000 GPU resources, and we observe that there are, on average, 2.97 images in a document. It is evident that we must perform main body extraction first and preliminary text filtering before detailed text filtering. So we define step \ding{192}: Preliminary Text Filtering, step \ding{193}: Document Deduplication with Text, step \ding{194}: Image Downloading \& Filtering, step \ding{195}: Detailed Text Filtering. The detailed settings can be seen in Table \ref{tab:pipeline_setting}. Since the main resource cost in step \ding{194} is bandwidth, it can be performed in parallel with other steps. Considering 1 billion documents, Table \ref{tab:pipeline_time} shows the processing time for all scenarios, where the processes in parentheses indicate that they can be performed in parallel.

It can be observed from Table \ref{tab:pipeline_time} that the order \ding{192}\ding{193}\ding{195}\ding{194} is the most efficient. Since we aim to preserve more diverse documents, we choose to perform \ding{192}\ding{193}(\ding{194}\ding{195}), retaining all documents after \ding{192} and \ding{193} along with their filtering results \ding{194} and \ding{195}.

\begin{table}[ht]
\begin{minipage}[t]{0.48\textwidth}
    \centering
    \includegraphics[width=0.9\textwidth]{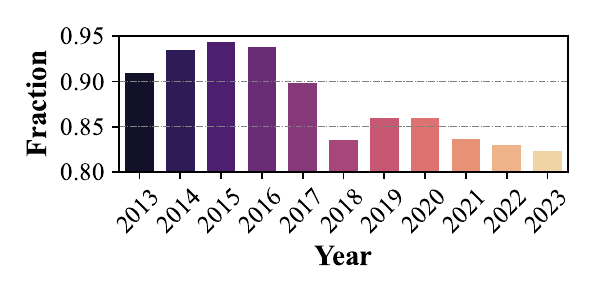}
    \captionof{figure}{\textbf{Trigger ratio of documents over years.} If a document is modified or filtered during our detailed text filtering, it will be included in the statistics.}
    \label{fig:filter_dump_year}

    \vspace{0pt}
    \centering
    \captionof{table}{
        \textbf{Detailed settings of each step.} The processing speed and filtering ratio are calculated as averages in the real data pipeline.
    }
    \label{tab:pipeline_setting}
    \begin{tabular}{c|r|r}
        \toprule
        \renewcommand{\arraystretch}{0.95}
         Step& \#Doc/Second & Filtering ratio\\
         \midrule
        \ding{192}&  1090k & 0.80\\
        \ding{193}&  388k & 0.90\\
        \ding{194} & 3k & 0.40 \\
        \ding{195}& 100k & 0.67\\ \bottomrule
    \end{tabular}
\end{minipage}
\hspace{10pt}
\begin{minipage}[t]{0.48\textwidth}
    \centering
    \vspace{-85pt}
    \captionof{table}{
        \textbf{Time to process 1B documents of different orders.} The processes in parentheses indicate that they can be performed in parallel. We find that \ding{192}\ding{193}\ding{195}\ding{194} is the optimal order, as changing any two steps would reduce the processing speed.
    }
    \label{tab:pipeline_time}
    \vspace{15pt}
    \begin{tabular}{l|r}
        \toprule
        \renewcommand{\arraystretch}{0.95}
         Order & Time (hours)\\
         \midrule
        \ding{192}\ding{193}\ding{195}\ding{194}&  \textbf{2.31}\\
        \ding{192}\ding{193}(\ding{194}\ding{195})&  5.95\\
        \ding{192}(\ding{194}\ding{193})\ding{195} & 56.14\\
        (\ding{194}\ding{192})\ding{193}\ding{195}& 278.37 \\
        \ding{193}\ding{192}\ding{195}\ding{194} & 2.65\\
        \ding{193}\ding{192}(\ding{194}\ding{195}) & 6.30\\
        \ding{193}(\ding{194}\ding{192})\ding{195} & 28.66\\
        (\ding{194}\ding{193}) \ding{192}\ding{195} & 278.26\\
        \ding{192}\ding{195}\ding{193}\ding{194} & 2.71\\
        \ding{192}\ding{195}(\ding{194}\ding{193}) & 19.33\\
        \ding{192}(\ding{194}\ding{195})\ding{193} & 55.90\\
        (\ding{194}\ding{192})\ding{195}\ding{193} & 279.59\\ \bottomrule
    \end{tabular}
\end{minipage}
\end{table}

\subsection{Details of the Human-Feedback Filtering}
The overall algorithm for our human-feedback filtering is shown in Algorithm \ref{alg: human_feedbak}. We iteratively update the filtering function set several times based on human feedback to generate high-quality documents, such as those without unfinished paragraphs or social media information. The detailed functions and their corresponding false positive rates can be seen in Table \ref{tab:filter_functions_en}. We sampled 1,000 documents to calculate the false positive rate. Many of these filtering functions have a false positive rate of zero, demonstrating the effectiveness of our designed filters. The trigger ratio of documents for each year can be seen in Figure \ref{fig:filter_dump_year}. We observe that our filtering functions work effectively across most documents, highlighting the necessity of our filters. Furthermore, we notice that the quality of documents in recent years is slightly better compared to older ones, resulting in a lower trigger ratio.

\begin{algorithm}[ht]
\caption{Human Feedback Algorithm}
\label{alg: human_feedbak}
\begin{algorithmic}[1]
\REQUIRE Documents $D^0 = \{d_1^0, d_2^0, ..., d_N^0\}$
\ENSURE Filtering functions $F = \{f_1, f_2, ..., f_M\}$
\STATE $F \leftarrow \{\}$
\FOR{$i = 1$ to $step$}
    \STATE Randomly sample $n$ documents $\hat{D}^{i-1}= \{d_1^{i-1}, d_2^{i-1}, ..., d_n^{i-1}\}$ from $D^{i-1}$
    \STATE Discovering $m$ problems by human feedback $P^{i} = \{p_1^{i}, p_2^{i}, ..., p_m^{i}\}$
    \STATE Generate $m$ filtering functions $F^{i} = \{f_1^{i}, f_2^{i}, ..., f_m^{i}\}$ according to $P^{i}$
    \STATE $F \leftarrow F + F^{i}$
    \STATE generate $D^i = \{d_1^i, d_2^i ..., d_N^i\}$, where
    \FOR{each $d^i \in D^i$}
        \FOR{each $f \in F^i$}
            \STATE $d^i \leftarrow f(d^{i-1})$
        \ENDFOR
    \ENDFOR
\ENDFOR
\end{algorithmic}
\end{algorithm}

\begin{longtable}{p{.70\textwidth} >{\raggedleft\arraybackslash}p{.20\textwidth}}
\caption{\textbf{Filtering rules.} The '-' indicates that the filtering function removed documents with hard indicators, rendering the false positive rate meaningless.}
\label{tab:filter_functions_en}
\\ \hline
\rule{0pt}{12pt}Filter Function & False Positive Rate\rule[-8pt]{0pt}{0pt}\\ \hline
\rule{0pt}{12pt}\textcolor{gray}{\emph{English Filtering Rules}} & \\
Remove abnormal newlines in text & 0.0\% \\
Split long underscores into paragraphs & 0.0\% \\
Remove elements related to videos & 0.0\% \\
Remove paragraphs with high number ratio & 0.0\% \\
Remove keywords related to social media & 0.0\% \\
Remove paragraphs with only one word & 0.0\% \\
Remove very short paragraphs with keywords & 5.8\% \\
Remove obviously aberrant list items & 0.0\% \\
Remove citation and related content & 6.0\% \\
Remove paragraphs ending with "readmore" & 0.0\% \\
Remove incomplete sentences at ends & 16.7\% \\
Remove video-related content & 0.0\% \\
Remove URLs from text & 0.0\% \\
Remove irrelevant image captions & 5.8\% \\
Remove specific ads from domain & 0.0\% \\
Mark articles with short paragraphs & 2.7\% \\
Mark articles with lists and tables & 0.0\% \\
Remove social media content & 2.1\% \\
Remove overly short paragraphs & 8.3\% \\
Remove paragraphs with many uppercase letters & 0.0\% \\
Remove paragraphs with pornographic content & 0.0\% \\
Remove footer content & - \\
Remove "like" and "follow" buttons & - \\
Remove short paragraphs & - \\
Remove paragraphs with word count issues & - \\
Remove documents with many non-letter words & - \\
Remove documents with few stop words & - \\
Remove documents with much pornographic content & - \\
Remove documents with bad paragraph length & - \\
\textcolor{gray}{\emph{Chinese Filtering Rules}} & \\
Remove duplicate lines and images & 4.0\% \\
Remove source info like author, photographer & 10.0\% \\
Remove sentences indicating newspaper flip & 0.0\% \\
Remove lines matching keywords & 0.0\% \\
Remove strange suffixes & 0.0\% \\
Mark articles with empty images & 0.0\% \\
Remove URLs from documents & 0.1\% \\
Remove documents with low text-image ratio & 0.0\% \\
Remove articles from cnnews-cepaper & 0.0\% \\
Remove keywords related to videos & 0.0\% \\
Fix empty titles from Baidu Baike & 0.0\% \\
Fix list format errors from Baidu Baike & 0.0\% \\
Remove recommendations and thanks to readers & 0.1\% \\
Remove disclaimers and copyright statements & 0.1\% \\
Remove content suspected of fraud & 0.1\% \rule[-8pt]{0pt}{0pt}\\ \hline
\end{longtable}

\section{Supplementary Data Analysis}

\subsection{Demonstrative Examples of {\dsname}}

We select two examples from {\subsetnamecc} as well as {\subsetnamecn} and one example from {\subsetnamevideo}, as presented in Table~\ref{tab:demo_cc}, Table~\ref{tab:demo_yt}, and Table~\ref{tab:demo_cw}, respectively.

\begin{longtable}{p{\textwidth}}
\caption{
    Two demonstrative documents selected from {\subsetnamecc}.
}
\label{tab:demo_cc}
\\ \hline
\rule{0pt}{12pt}\textbf{\textit{Example 1:}}\rule[-5pt]{0pt}{0pt}\\
Mother's Day is fast approaching. What better way to say 'i love you' to your Mum this year, by creating her this unique necklace, tailoring the fabrics, colours and beads all to your Mum's personal tastes.\\
Cut out your desired collar shape from a sturdy felt.\\
Choose a collection of clear acrylic stones in a selection of shapes. Cover them with a thin chiffon material, so you can still see the facets of the gems. Gather the fabric at the back of the gem and tack it together.\\
Sew the fabric covered stones onto your felt collar. Position them so that they sit slightly higher than the top edge of the collar to hide the felt.\\
Line up a string of multi-coloured beads made from precious stones along the bottom edge of the collar. Tack the string to the collar every 3 beads.\\
Fill in the gaps between the gems and beads with sew-on genuine crystal diamante stones in clasps.\rule[-10pt]{0pt}{0pt}\\
\includegraphics[width=0.3\textwidth]{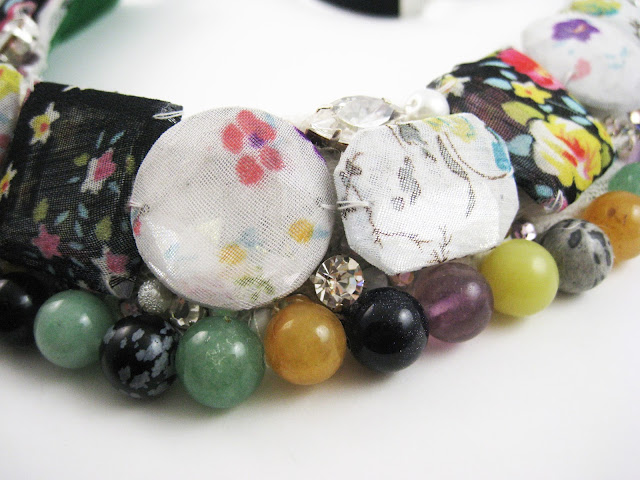}\\
\rule{0pt}{12pt}Measure a strip of black grosgrain ribbon to the length you wish your necklace to be. Cut it in half and stitch one end of each strip to the back of each tip to create the 'chain'.\\
Slot a ribbon end clasp onto the tip of each ribbon and close in place with a pair of jewellery pliers. Finish off with a screw clasp.\rule[-10pt]{0pt}{0pt}\\
\includegraphics[width=0.3\textwidth]{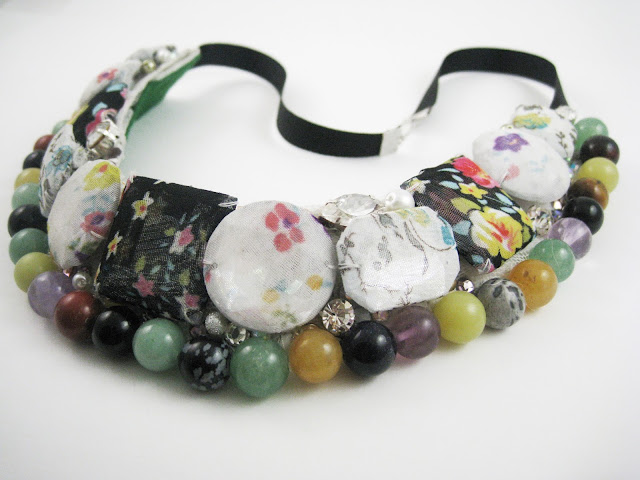}\rule[-10pt]{0pt}{0pt}\\
\hline
\rule{0pt}{12pt}\textbf{\textit{Example 2:}}\rule[-5pt]{0pt}{0pt}\\
When my craft room came into being, at the end of February (actually it's still not missing the pink glass splashback..) I wanted the first thing I did to be something a bit special...\rule[-10pt]{0pt}{0pt}\\
\includegraphics[width=0.3\textwidth]{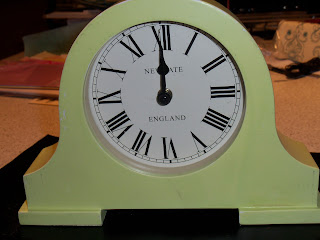}\rule[-10pt]{0pt}{0pt}\\
I found this clock on a clearance shelf, and whilst it was a bit in your face lime green, I liked the shape. I bought it, and put it to one side. Then I got inspiration...\rule[-10pt]{0pt}{0pt}\\
\includegraphics[width=0.3\textwidth]{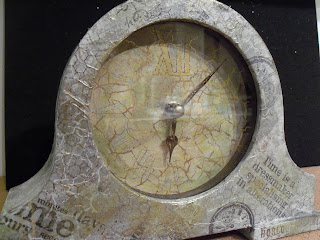}\rule[-10pt]{0pt}{0pt}\\
After a little bit of work, it now looks like this...\rule[-10pt]{0pt}{0pt}\\
\includegraphics[width=0.3\textwidth]{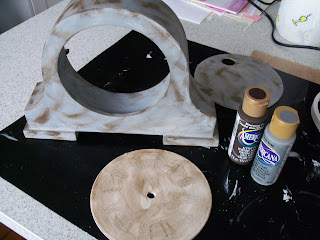}\rule[-10pt]{0pt}{0pt}\\
...and painted them up in decoart americana paint, roughly.\rule[-10pt]{0pt}{0pt}\\
\includegraphics[width=0.3\textwidth]{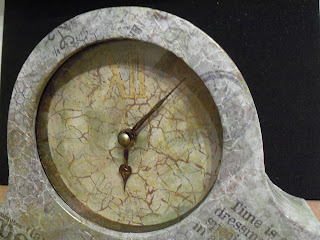}\rule[-10pt]{0pt}{0pt}\\
Putting it all together, the clock was sealed with claudine hellmuth multi medium, matte, which I also used as a 'glue' to cover the clock in the stamped tissue. I gave it another all over coat of the matte medium to seal it completely. There's also a smidge (or should I say smudges) of the grungold inka gold - it's so yummy! And now I have a really smart clock on my shelf!\rule[-10pt]{0pt}{0pt}\\
\hline
\end{longtable}

\begin{longtable}{p{\textwidth}}
\caption{
    One demonstrative document was selected from {\subsetnamevideo}.
}
\label{tab:demo_yt}
\\ \hline
\rule{0pt}{12pt}\textbf{\textit{Example:}}\rule[-5pt]{0pt}{0pt}\\
Merry Christmas guys or happy Christmas. If you live in the UK, the marbles and I are going to show you what we got for Christmas. \\
\includegraphics[width=0.9\textwidth]{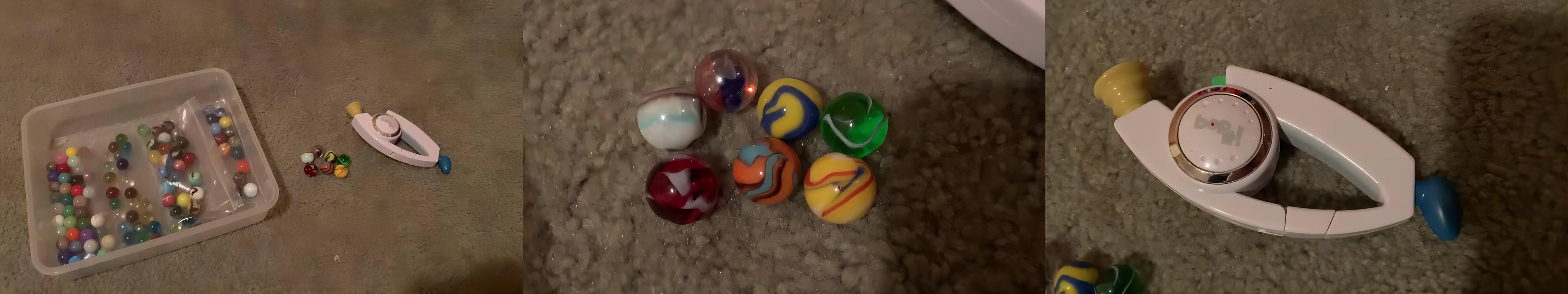}\\
We have seven new rainbow marbles and the 2009 Bobbitt carabiner or carabiner. However it's pronounced yes this is new as you can see, and it was really cheap it was like twelve dollars yes. Anton told me on the note I wrote to him telling him what I want for Christmas and this works perfect.\\
\includegraphics[width=0.3\textwidth]{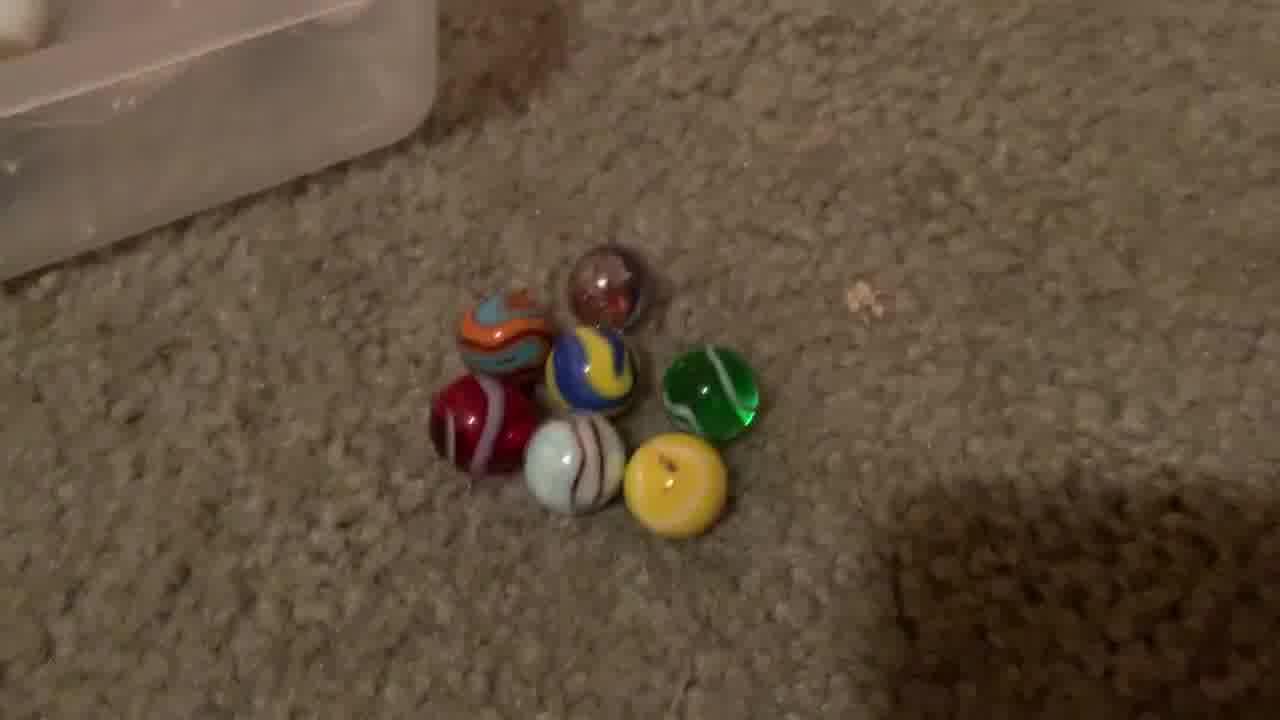}\\
Anyways, moving on to the marbles. We have seven new rainbows. \\
\includegraphics[width=0.3\textwidth]{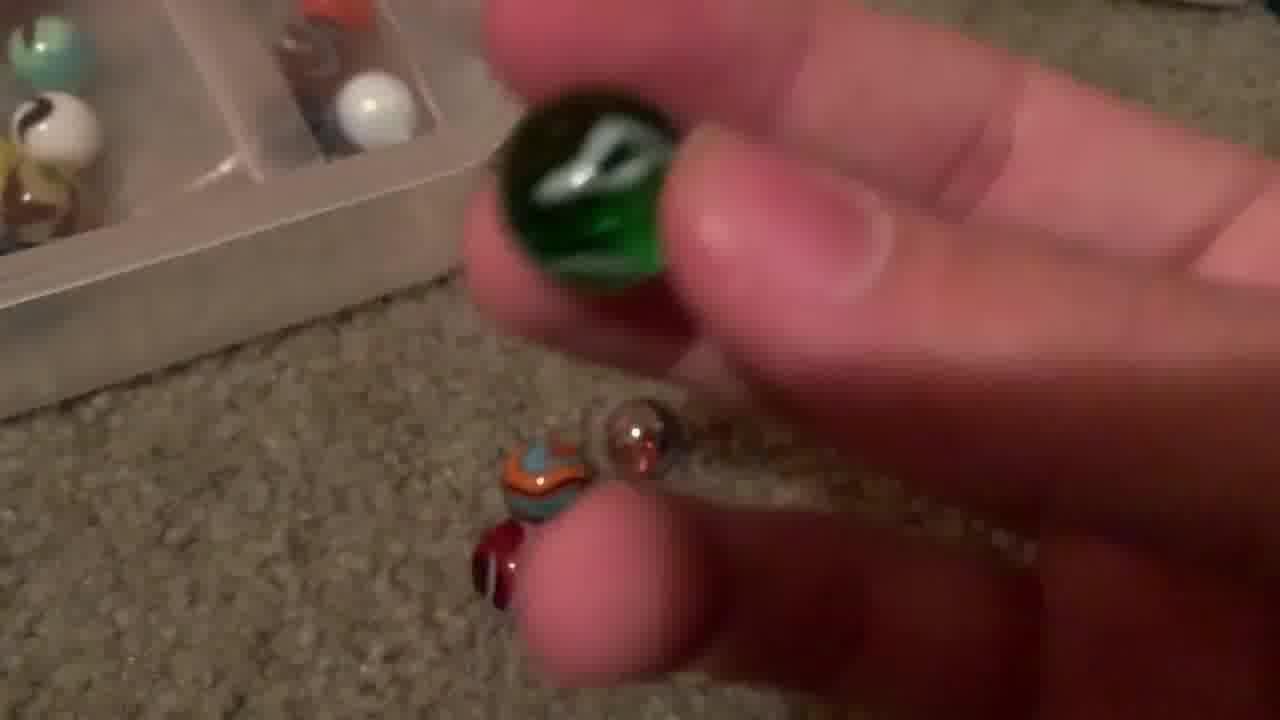}\\
We have enchanted forest which is a clear green marble with white swirls. \\
\includegraphics[width=0.3\textwidth]{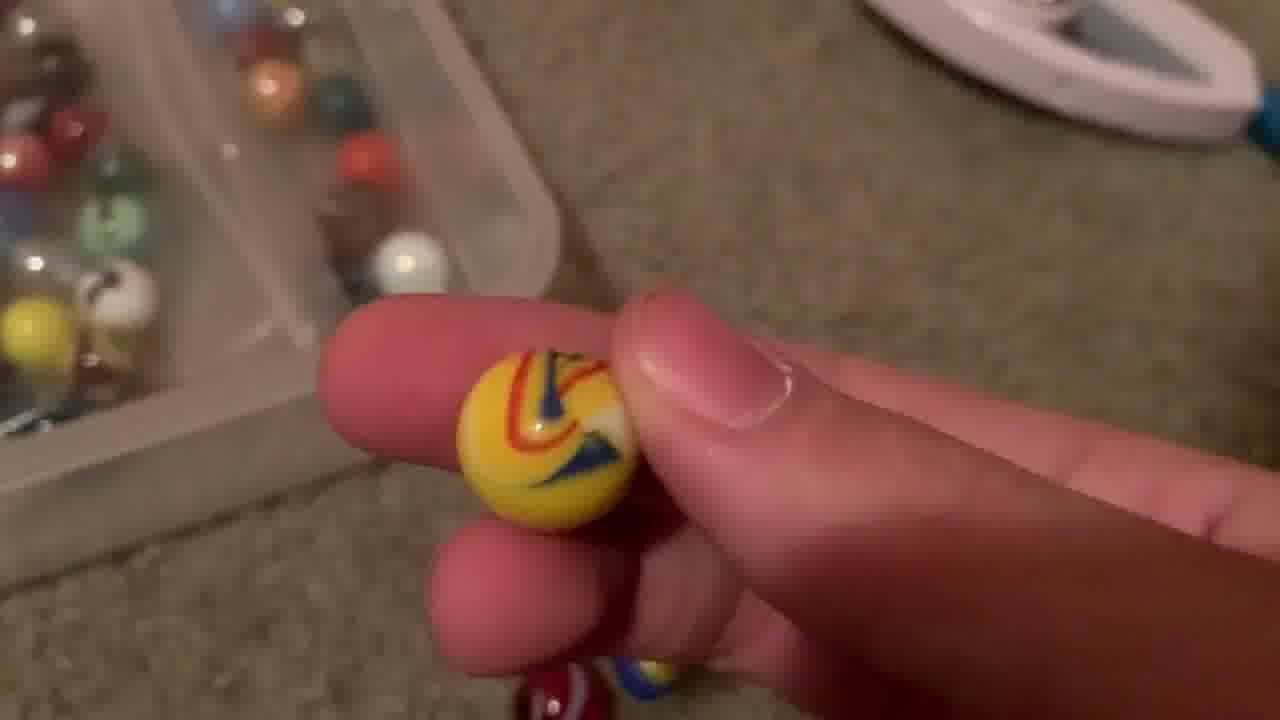}\\
We have parrot which is a yellow marble with red blue and white swirls. \\
\includegraphics[width=0.3\textwidth]{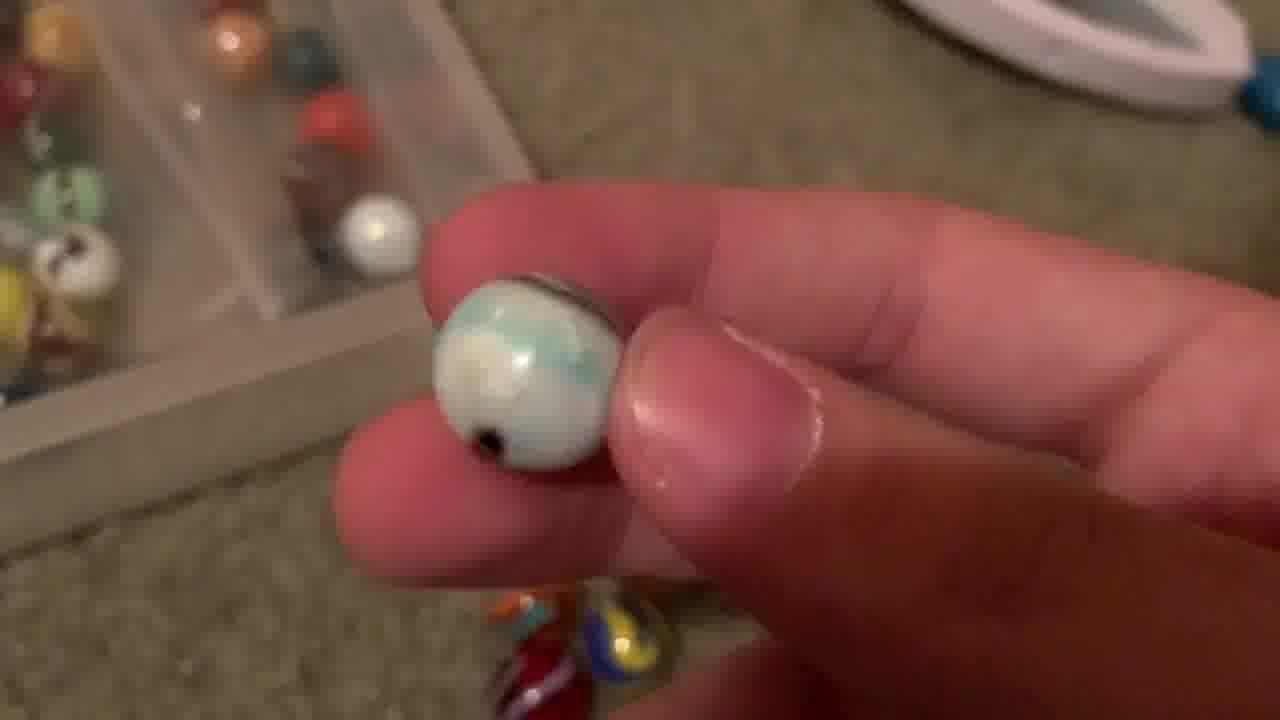}\\
We have white tiger which is an IRA dies Dwight marble with blue and black swirls. \\
\includegraphics[width=0.6\textwidth]{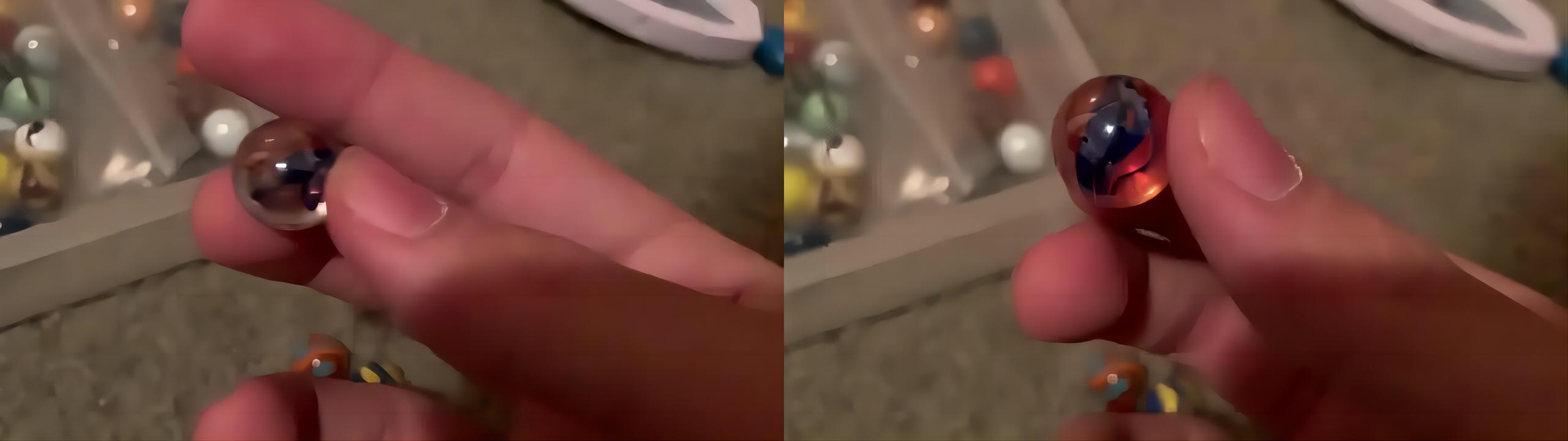}\\
Next up, we have sunrise which is an IRA diced clear marble with a red and blue cat eye. It actually does look like a sunrise. \\
\includegraphics[width=0.3\textwidth]{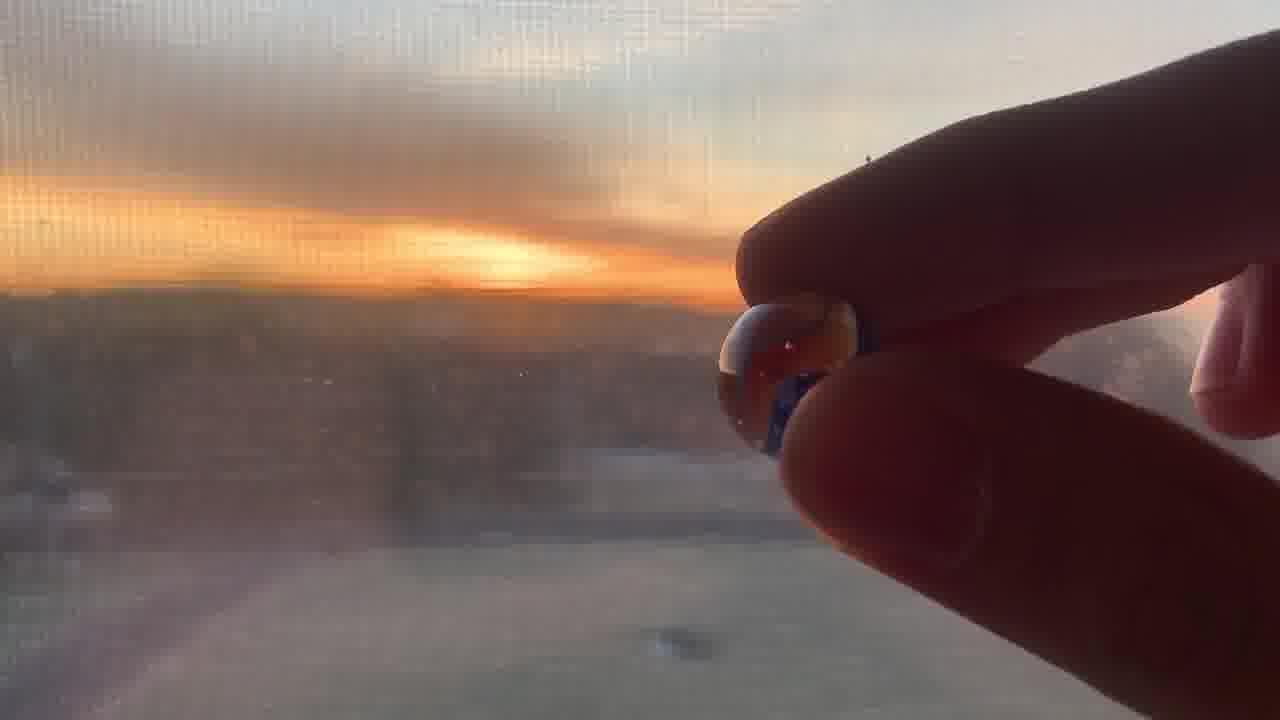}\\
And if you guys look there is a beautiful sunrise outside of my house and here's sunrise right here does that not look like a sunrise. And yes, I still have my air conditioner ready anyways. \\
\includegraphics[width=0.3\textwidth]{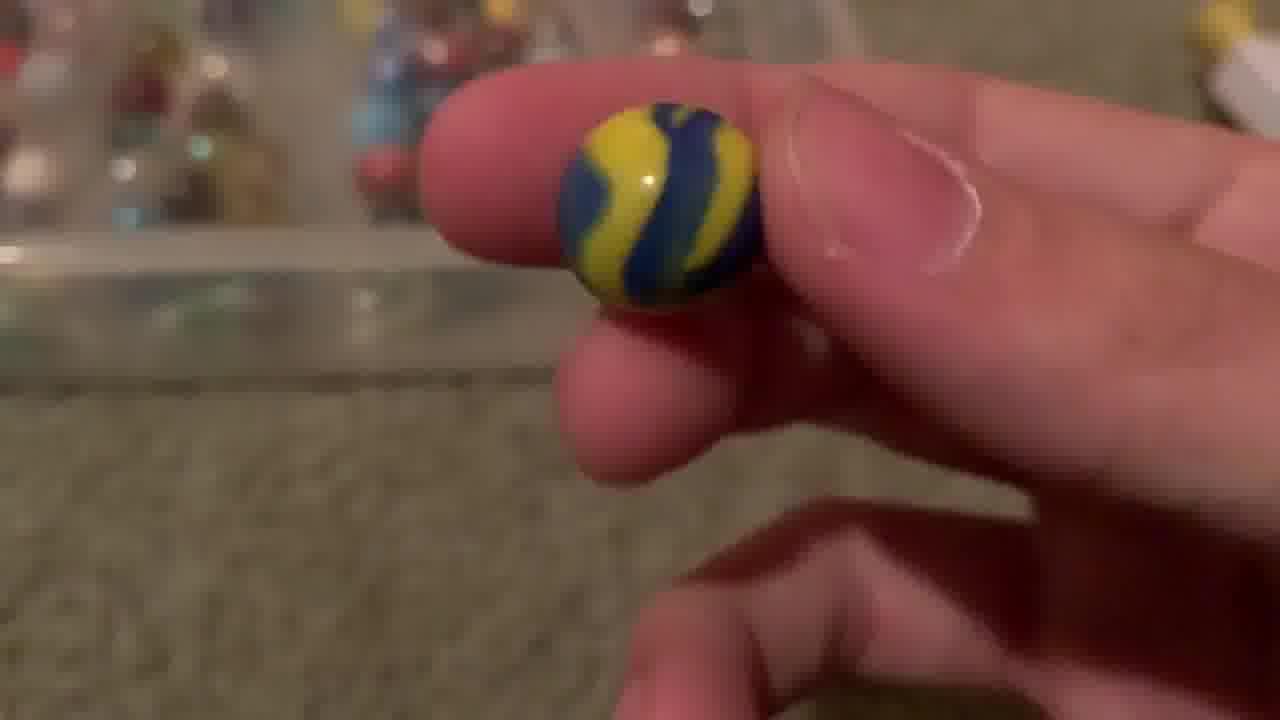}\\
Next up, we have blue tang which is a yellow marble with blue swirls. \\
\includegraphics[width=0.3\textwidth]{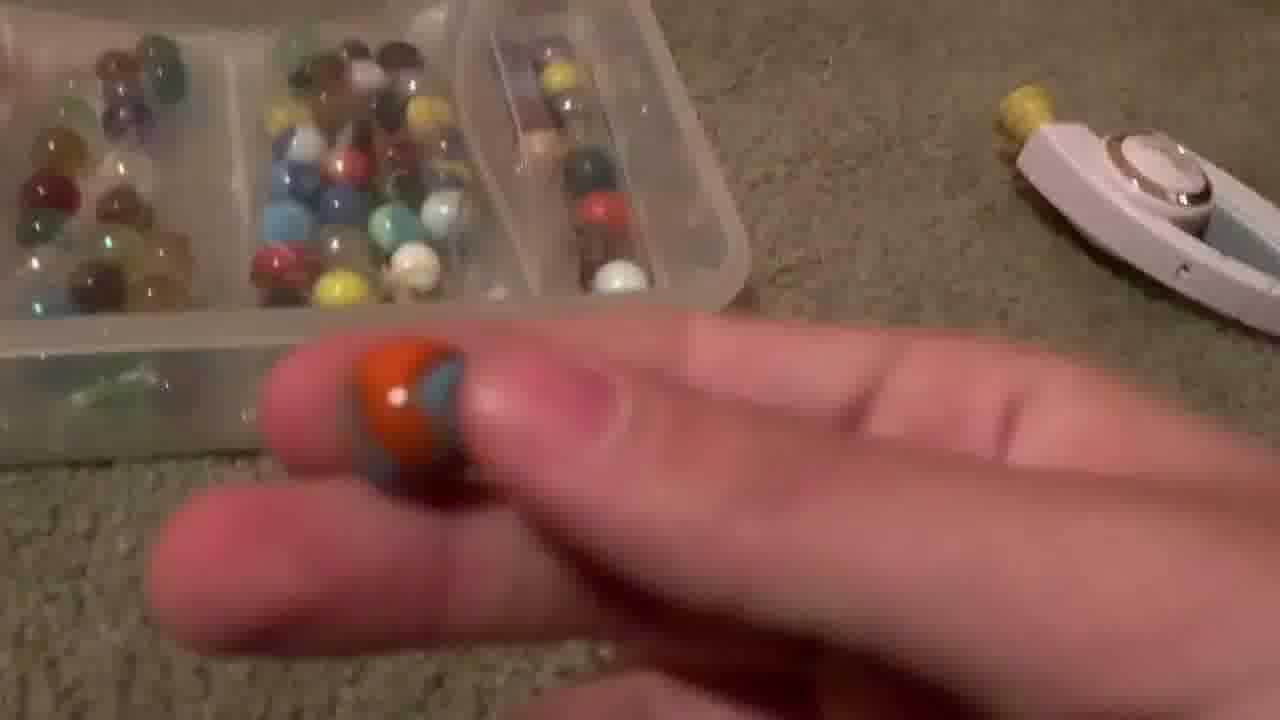}\\
Next up, we have seahorse which is an orange marble with blue and black swirls. \\
\includegraphics[width=0.3\textwidth]{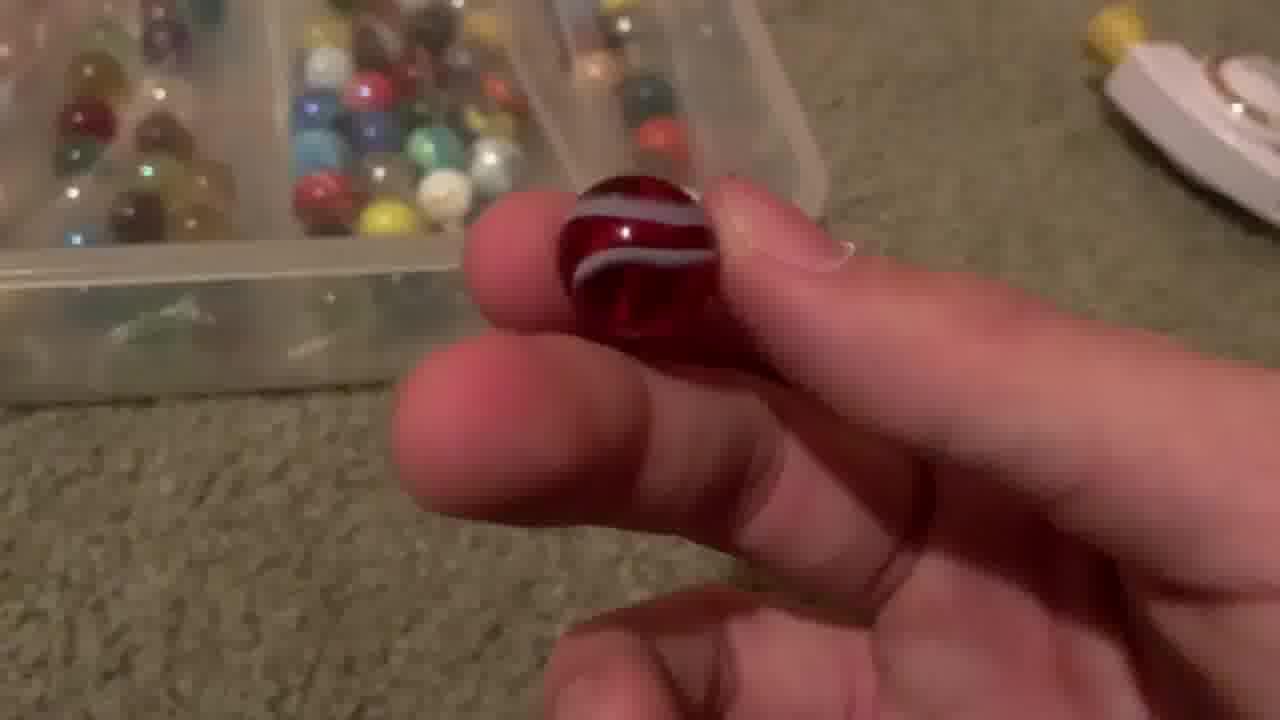}\\
And last but not least, we have rooster which is a clear red marble with white swirls.\\
\includegraphics[width=0.3\textwidth]{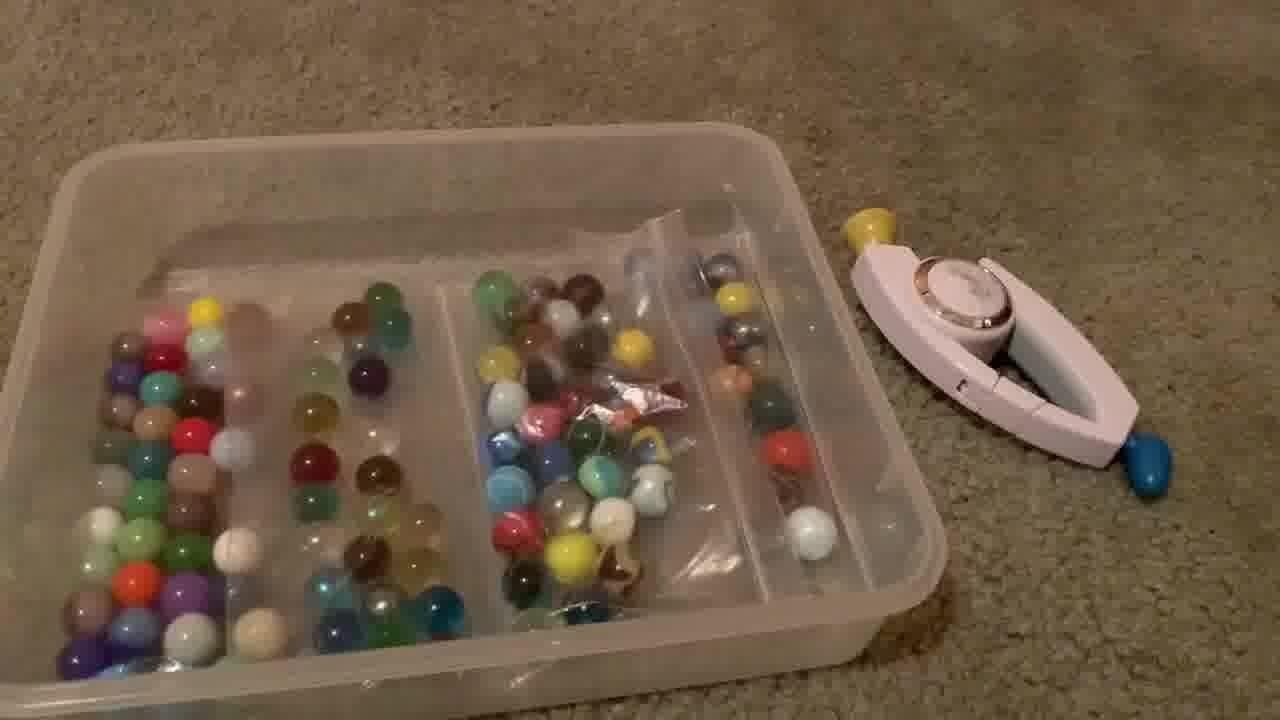}\\
So that brings me to a total of 107 marbles soon to be 108 because apparently I got another marble but that's at another house it's called Fiesta. I'll show you guys Fiesta once I'm in Florida alright guys. \\
I love you guys. Merry Christmas. And I'll see you guys on December 27 for the what we did in 2019 video. All right. I love you guys. Peace out.\rule[-10pt]{0pt}{0pt}\\
\hline
\end{longtable}

\begin{CJK}{UTF8}{gbsn} %
\begin{longtable}{p{\textwidth}}
\caption{
    Two demonstrative documents selected from {\subsetnamecn}.
}
\label{tab:demo_cw}
\\ \hline
\rule{0pt}{12pt}\textbf{\textit{Example 1:}}\rule[-5pt]{0pt}{0pt}\\
毫米波技术正广泛应用于无人驾驶\rule[-10pt]{0pt}{0pt}\\
\includegraphics[width=0.3\textwidth]{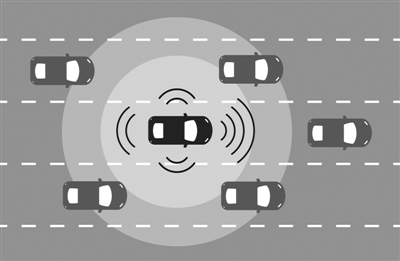}\rule[-10pt]{0pt}{0pt}\\
毫米波雷达指工作在毫米波波段的雷达，是测量被测物体相对距离、相对速度、方位的高精度传感器，早期被应用于军事领域，随着雷达技术的发展与进步，毫米波雷达传感器开始应用于汽车电子、无人机、智能交通等多个领域。\\
同超声波雷达相比，毫米波雷达具有体积小、质量轻和空间分辨率高的特点。与红外、激光、摄像头等光学传感器相比，毫米波雷达穿透雾、烟、灰尘的能力强，具有全天候全天时的特点。另外，毫米波雷达的抗干扰能力也优于其他车载传感器。由于毫米波在大气中衰减弱，所以可以实现更远距离的探测与感知，其中远距离雷达可以实现超过200米的感知与探测。\\
目前各个国家对车载毫米波雷达分配的频段各有不同，但主要集中在24GHz和77GHz。\\
频段在24GHz左右的毫米波雷达检测距离有限，因此常用于检测近处的障碍物，常被用来实现的功能有盲点检测、变道辅助等，主要为换道决策提供感知信息。\\
而性能良好的77GHz雷达的最大检测距离可以达到160米以上，因此常被安装在前保险杠上，正对汽车的行驶方向。长距离毫米波雷达能够用于实现紧急制动、高速公路跟车等功能；同时也能满足自动驾驶领域，对障碍物距离、速度和角度的测量需求。\rule[-10pt]{0pt}{0pt}\\
\hline
\rule{0pt}{12pt}\textbf{\textit{Example 2:}}\rule[-5pt]{0pt}{0pt}\\
三彩披鬃鞍马（唐）\rule[-10pt]{0pt}{0pt}\\
\includegraphics[width=0.2\textwidth]{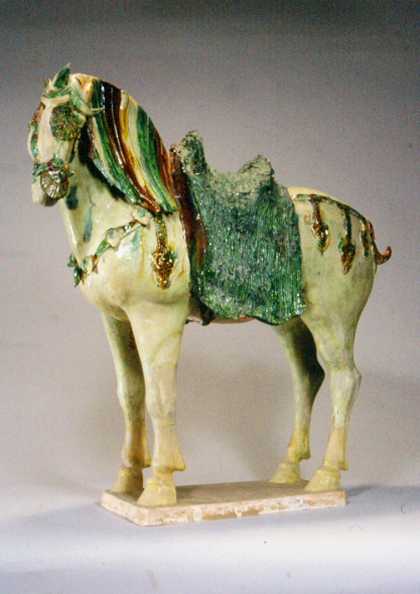}\rule[-10pt]{0pt}{0pt}\\
三彩披鬃鞍马，1990年陕西省西安市灞桥区半坡村出土，通高56.5cm，长58cm。\\
马首向左后方回望，两耳直竖，鬃毛左披，立于长方形踏板上。臀部发达，腿部强劲有力，特别是其眼睛、耳朵、筋骨、肌肉等局部雕琢精细，刀工娴熟，符合实体马的特点。马全身以白色为地釉，鬃毛为白、绿、褐三色相间；马鞍及垂于两侧腹下之毛织物为绿色釉；额前的当卢、耳鼻际的辔饰、胸前及尻上的革带及杏叶形垂饰均为黄、绿、褐三色釉；马尾为褐色。与一般唐三彩马相比，此马的釉色别具韵味，缺少大片鲜艳的红、黄、褐等色，而以素雅的白、绿色为主要色调，给人耳目一新的感觉。其造型俊洁、匀称，是唐三彩中罕见的精品。\\
唐代三彩马在造型上显示出宏大的气魄，体现着大唐王朝繁荣昌盛的景象，并从中可以看出唐人丰肥健壮的审美情趣。在形态上虽各有风采，但它们都有着共同的特征，即头小颈粗，臀圆背厚，四肢粗壮，而且骨肉匀停，线条流畅，内在的神韵在完美的造型中得到十足的体现，有力地烘托了盛世王朝的繁荣气象。\rule[-10pt]{0pt}{0pt}\\
\hline
\end{longtable}
\end{CJK}

\begin{figure}[ht]
    \centering
    \begin{subfigure}{.3\textwidth}
        \centering
        \includegraphics[width=\linewidth]{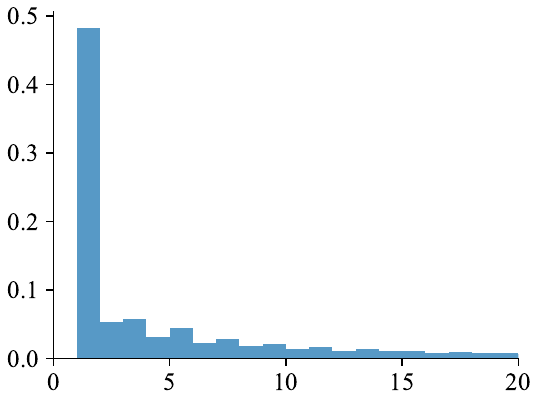}
        \caption{Line Number}
    \end{subfigure}
    \begin{subfigure}{.3\textwidth}
        \centering
        \includegraphics[width=\linewidth]{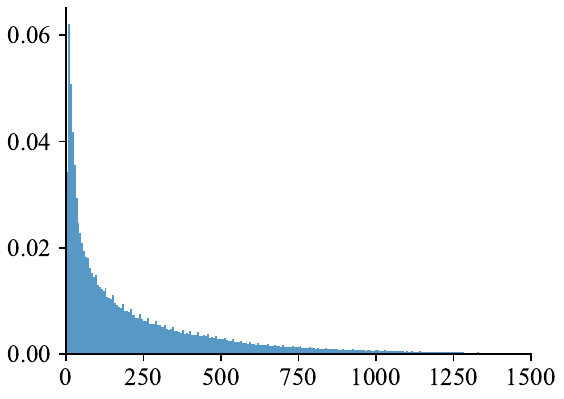}
        \caption{Line Length}
    \end{subfigure}%
    \begin{subfigure}{.3\textwidth}
        \centering
        \includegraphics[width=\linewidth]{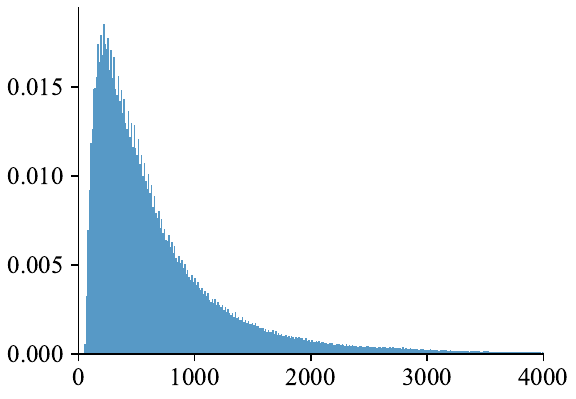}
        \caption{Token Length}
    \end{subfigure}
    \begin{subfigure}{.3\textwidth}
        \centering
        \includegraphics[width=\linewidth]{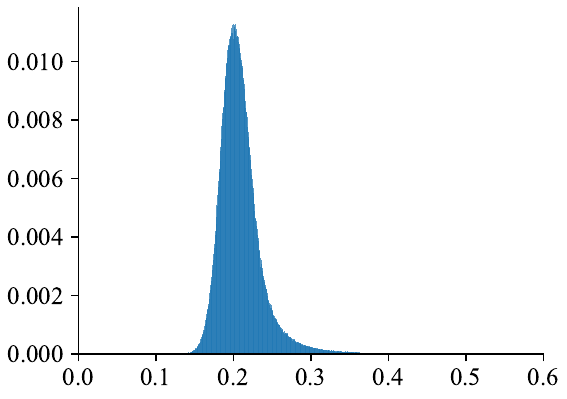}
        \caption{Non-alpha Fraction}
    \end{subfigure}%
    \begin{subfigure}{.3\textwidth}
        \centering
        \includegraphics[width=\linewidth]{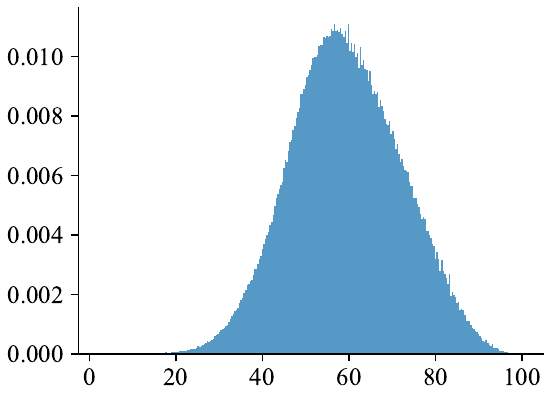}
        \caption{Unique Words Fraction}
    \end{subfigure}
    \begin{subfigure}{.3\textwidth}
        \centering
        \includegraphics[width=\linewidth]{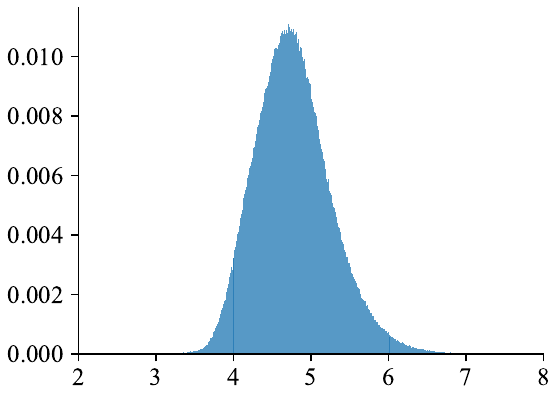}
        \caption{Mean Word Length}
    \end{subfigure}
    \begin{subfigure}{.3\textwidth}
        \centering
        \includegraphics[width=\linewidth]{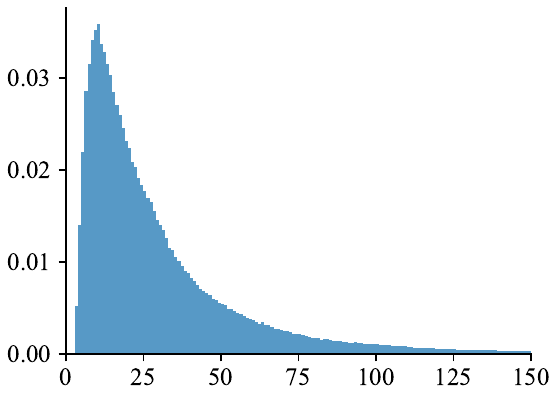}
        \caption{Sentence Number}
    \end{subfigure}%
    \begin{subfigure}{.3\textwidth}
        \centering
        \includegraphics[width=\linewidth]{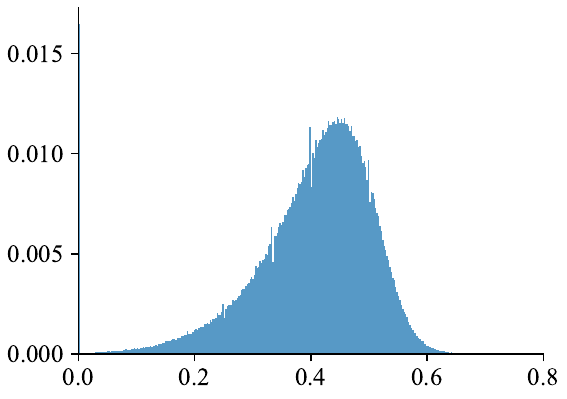}
        \caption{Stop Word Fraction}
    \end{subfigure}
    \begin{subfigure}{.3\textwidth}
        \centering
        \includegraphics[width=\linewidth]{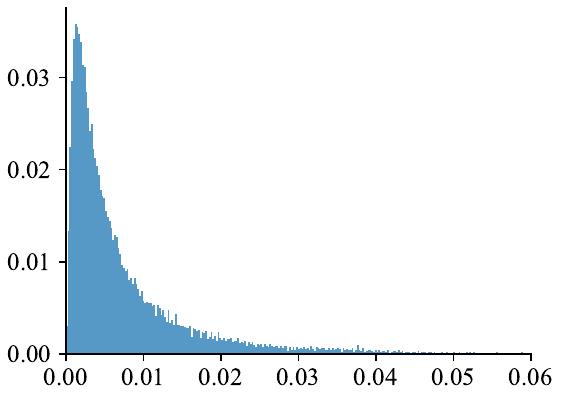}
        \caption{Symbol to Word Ratio}
    \end{subfigure}
    \caption{Percentage Statistics for Metrics on {\subsetnamecc}.}
    \label{fig:sup_statistics}
\end{figure}

\subsection{Statistics}

We follow Wanjuan-CC~\cite{qiu2024wanjuan} to compute several data quality metrics. As shown in Figure~\ref{fig:sup_statistics}, a statistical analysis is conducted on various quantitative metrics of the documents, including document length, line count, token length, percentage of non-alphabetic characters, proportion of unique words, average word length, sentence count, stop-word ratio, and symbol-to-word ratio. The distributions enable the users to have a comprehensive understanding of the various characteristics of the data.

\subsection{Topic Modeling Results}

We follow previous works~\cite{zhu2024mmc4,laurenccon2024obelics} to measure the diversity of the corpus with LDA~\cite{blei2003lda}, which presents the estimated proportions and related words for each topic. We train LDA with 20 topics on 100,000 documents randomly sampled from each partition of our dataset. The results on {\subsetnamecc}, {\subsetnamevideo}, and {\subsetnamecn} are shown in Table~\ref{tab:lda_cc_result}, Table~\ref{tab:lda_yt_result}, and Table~\ref{tab:lda_cw_result}, respectively. Figure~\ref{fig:diversity-visualization} shows T-SNE~\cite{van2008visualizing} projection of LDA topic clusters. For each document, we generate a 20-dimensional vector and then reduce it to a 2-dimensional vector using T-SNE, allowing for visualization. From the Topic Modeling Result, we can find that the MMC4 and {\subsetnamecc} have several overlapping topics because they both originate from Common Crawl, and most topics are unique in all three sources, demonstrating the large diversity of the document content in {\dsname}

\begin{longtable}{p{.20\textwidth} p{.10\textwidth} p{.60\textwidth}}
\caption{The detailed topic modeling results of {\subsetnamecc}.}
\label{tab:lda_cc_result}
\\ \hline
\rule{0pt}{12pt}Concept&Ratio&Related words\rule[-10pt]{0pt}{0pt}\\ \hline
\rule{0pt}{12pt}Legal System & 5.04\% & court, men, march, prison, trial, media, department, national, wales, ceremony, issued, tower, medal, brexit, sheriff, service, gang, khan, charity, dead\\
Political Race & 3.21\% & vote, ireland, mayor, taylor, elections, ford, usa, presidential, flag, seats, opposition, national, position, poll, circuit, lisa, dublin, colin, sox, eddie\\
Defense Tactics & 2.00\% & security, photography, laura, miller, tennis, robin, douglas, norway, clarke, defence, rebels, anime, bryan, durham, grandpa, emperor, chad, manga, cia, rafael\\
Digital Analysis & 4.40\% & trump, article, software, network, user, check, view, tool, method, device, mobile, biden, link, processing, output, resolution, document, audio, scale\\
Environmental \newline Impact& 2.68\% & paul, green, moon, waste, aircraft, toronto, wilson, bell, emissions, cemetery, tennessee, kentucky, flying, mars, crops, lawsuit, copper, belgium, idaho, violations\\
Educational \newline Initiatives & 11.74\% & program, service, national, department, schools, plan, quality, media, organization, world, teachers, campus, faculty, european, medicine, significant, innovation, clients, vision, journal\\
Creative Crafts & 4.25\% & paper, card, items, cards, challenge, drawing, letters, settings, bags, craft, pen, custom, lesson, printed, knitting, tag, check, link, supplies, ribbon\\
Historical Europe & 6.55\% & world, england, century, france, men, prayer, germany, queen, dead, stone, martin, die, italy, elizabeth, nations, view, castle, anna, powers, christians\\
Diverse Imagery & 4.09\% & space, ice, wings, layout, stamp, angel, disabilities, ram, spain, mess, owl, satan, boot, header, flames, swap, unto, cracks, milwaukee, isaac\\
Familial Bonds & 6.35\% & family, mother, young, christmas, boys, scotland, service, sam, mum, named, uncle, men, scottish, creep, picture, aunt, iowa, aged, rev, neighbors\\
Artistic Expression & 6.96\% & music, world, films, audience, play, writers, actor, young, painting, visual, media, gallery, contemporary, steve, artistic, popular, reviews, entertainment, pop, poem\\
Cooking Essentials & 4.09\% & cup, pan, milk, fresh, sauce, eggs, dish, chopped, meat, tsp, onion, salad, tbsp, tomatoes, raw, green, pasta, acid, fridge, corn\\
Weekend Fun & 9.33\% & night, weekend, sunday, picture, saturday, cake, check, plan, hit, paint, cat, awesome, summer, bike, yeah, play, dead, view, wake, shoes\\
Cultural Variety & 2.14\% & apple, foods, pakistan, davis, wire, pumpkin, mumbai, neil, hawaii, pearl, currency, louisiana, cluster, loaf, apples, moss, orleans, consultants, pudding, guild\\
Healthcare \newline Solutions & 2.85\% & skin, therapy, village, ukraine, drugs, ontario, immune, bath, tissue, lewis, ukrainian, twins, visa, infected, infections, substance, dose, respiratory, wifi, proteins\\
Seasonal Gardening & 4.11\% & summer, garden, winter, spring, trees, soil, flowers, green, quilt, harvest, peppers, kenya, deer, fresh, ice, peas, planting, batch, microwave, abortion\\
Infrastructure \newline Development & 5.48\% & museum, construction, region, national, forest, airport, birds, parking, wood, vehicles, marine, village, service, view, lane, harris, trees, streets, electric, parts\\
Current Events & 2.28\% & san, governor, com, ing, coronavirus, vaccine, bears, cable, luke, ter, rolls, bloom, cnn, moses, ash, lent, bees, chi, stitches, brunswick\\
Sports Thrills & 8.35\% & game, play, players, club, football, sports, cup, winning, hit, saturday, score, scored, night, victory, competition, sunday, baseball, young, beat, ryan\\
Housing Market & 4.11\% & price, housing, average, billion, cent, capital, period, smith, prices, estate, firm, numbers, hospitals, records, rose, commercial, significant, march, trend, latest\rule[-10pt]{0pt}{0pt}\\
\hline
\end{longtable}

\begin{longtable}{p{.20\textwidth} p{.10\textwidth} p{.60\textwidth}}
\caption{\rule{0pt}{12pt}The detailed topic modeling results of {\subsetnamevideo}.}
\label{tab:lda_yt_result}
\\ \hline
\rule{0pt}{12pt}Concept&Ratio&Related words\rule[-10pt]{0pt}{0pt}\\ \hline
\rule{0pt}{12pt}Assembly Tools & 8.06\% & side, ahead, bottom, half, tape, frame, size, slide, kit, add, tool, plug, wire, screw, holes, table, double, sides, screws, panel\\
Political Religion & 5.13\% & president, government, jesus, war, state, lord, political, father, africa, international, christ, japan, donald, korea, may, minister, truth, foreign, pray, faith\\
Sports Competition & 6.10\% & season, win, goal, league, teams, final, fans, half, round, score, competition, basketball, tonight, winner, sport, shot, plays, side, pitch, title\\
Family Routine & 8.40\% & kids, morning, girls, parents, live, beautiful, birthday, hours, yesterday, saw, tonight, dance, bathroom, table, hear, tired, waiting, coffee, lunch, makes\\
Makeup Routine & 2.82\% & skin, lip, apply, powder, blend, ahead, liquid, photo, brushes, lashes, mac, mascara, add, primer, coverage, routine, blending, vitamin, favorite, makes\\
Printed Media & 2.58\% & book, page, board, list, write, copy, printed, title, add, photo, printing, author, craft, mustang, compression, acrylic, washi, images, favorite, macros\\
Gender Education & 2.93\% & women, class, students, training, learn, schools, golf, culture, teaching, campus, state, society, industry, events, arts, sexual, youth, local, gender, role\\
Vehicle Features & 3.63\% & paint, rear, painting, side, window, fragrance, windows, hood, steering, roof, coat, storage, beautiful, transmission, horsepower, motorcycle, sport, trunk, honda, makes\\
Financial Investment & 5.50\% & money, dollars, dollar, cost, worth, value, spend, cash, buying, tax, may, income, local, businesses, marketing, spending, investment, industry, live, interest\\
Learning Methods & 8.11\% & may, question, learn, key, makes, negative, positive, ways, live, specific, computer, rather, value, results, add, function, search, creating, images, mobile\\
Medical Health & 2.82\% & blood, cancer, may, medicine, healing, emily, pregnancy, symptoms, recovery, drugs, emergency, sheriff, tissue, oxygen, trial, healthy, bacteria, labor, southwest, appointment\\
Urban Affairs & 5.98\% & city, police, morning, live, tonight, county, state, hours, bus, west, local, officer, california, valley, officers, parking, department, neighborhood, travel, clouds\\
Animal Care & 2.36\% & animals, cat, cats, cage, madrid, deer, species, hunting, euros, soccer, rescue, pets, ski, pig, trap, lion, cow, zoo, mattress, aquarium\\
Physical Exercise & 4.02\% & side, feet, leg, arm, lower, ground, shoulder, knee, flat, knees, roll, jump, core, valve, exhale, kick, swing, grip, twist, weight\\
Cooking Ingredients & 5.46\% & add, cheese, half, sauce, coffee, egg, bowl, ingredients, ahead, tastes, pour, powder, potatoes, vegetables, wine, stir, beef, onion, bacon, teaspoon\\
Fashion Preferences & 5.76\% & beautiful, favorite, size, pair, outfit, pants, pizza, comfortable, side, bottom, jeans, makes, leather, rose, saw, dollars, dollar, walmart, tag, halloween\\
Fitness Activities & 3.45\% & weight, boat, workout, exercise, fishing, fat, training, protein, foods, calories, healthy, bait, fitness, half, morning, exercises, bass, squat, rope, ups\\
Music Performance & 9.46\% & man, shot, hear, sound, saw, sounds, hell, record, live, shooting, yep, makes, kill, money, songs, guitar, nobody, album, laughter, hmm\\
Outdoor Gardening & 2.43\% & garden, winter, trail, shoe, land, soil, ground, beautiful, yarn, feet, hike, mountains, double, seed, concrete, fence, stitches, bucket, half, seeds\\
Popular Entertainment & 4.99\% & king, disney, john, favorite, scene, magic, shows, fans, films, batman, marvel, stars, deck, comic, artists, artist, role, war, ship, may\rule[-10pt]{0pt}{0pt}\\
\hline
\end{longtable}

\begin{CJK}{UTF8}{gbsn} %
\begin{longtable}{p{.25\textwidth} p{.10\textwidth} p{.60\textwidth}}
\caption{\rule{0pt}{12pt}The detailed topic modeling results of {\subsetnamecn}. The original Chinese concepts and related words are translated into English.}
\label{tab:lda_cw_result}
\\ \hline
\rule{0pt}{12pt}Concept&Ratio&Related words\rule[-10pt]{0pt}{0pt}\\ \hline
\rule{0pt}{12pt}Academic Preparation \newline (学 业 准 备) & 3.56\% & divide, open, classmate, interview, apply, review, college entrance examination (GAOKAO), prize number, question, code, subject, vocabulary, retest, balcony, foundation, resume, memorize, internship, school, capability\\
Medical Technology \newline (医 学 科 技) & 2.90\% & medical treatment, surgery, illness, discovery, experiment, substance, laser, flight, measurement, include, capability, launch, grain, drone, tumor, rocket, cat, crowd, medication, appointment\\
Regional Affiliation \newline (地 域 关 联) & 4.92 \% &Japan, Shanghai, Hong Kong, Hangzhou, Taiwan, Zhejiang, Nanjing, army, Guangdong, region, Jiangsu, Xinjiang, Chen (a surname), family, dynasty, mainland, Yunnan, Tibet, combat, Shanghai City\\
Global Initiatives \newline (全 球 计 划) & 4.33\% & United States, epidemic, region, plan, India, including, president, Ukraine, billion US dollars, ten thousand US dollars, recovery, appointment, Nezha (a mythical figure in Chinese mythology), Biden, currency, Russia, blockchain, supplies, Bitcoin, Soviet Union\\
Grassroots Governance \newline (基 层 治 理) & 5.47\% & community, deputy, party, rural, village, county, secretary, director, mu (a unit of area, equal to 1/15 hectare), the 20th National Congress, guarantee, implement, province, region, railway, previous year, within the territory, characteristic, comrade, activity\\
School Education \newline (学 校 教 育) & 5.31\% & education, school, activity, college, teaching, teacher, parent, excellent, occupation, practice, primary school, capability, classmate, campus, sports, public welfare, association, classroom, student, characteristic\\
Emotional Expression \newline (情 感 表 达) & 15.40\% & very, like, discover, feel, find, finish, the other party, spot, several, not want, slowly, fast, seems like, probably, buy, that kind, capability, habit, body, unable to\\
Smart Devices \newline (智 能 设 备) & 3.86\% & mobile phone, baby, long, function, millimeter, charge, plant, wide, glass, machine, dual, screen, inch, Haier (a brand), material, select and purchase, indoor, Bluetooth, tool, leaf\\
Culinary Experience \newline (美 食 体 验) & 3.83\% & decoration, tea, delicious, fruit, restaurant, taste, texture, put in, dish, store, catering, refrigerator, breakfast, flavor, white liquor, fragrant, ingredients, free, milk tea, guest room\\
Legal Violations \newline (法 律 违 规) & 4.91\% & journalist, discover, apply, handle, spot, contract, store, illegal, clause, activity, supermarket, receive, express delivery, afternoon, crime, takeout, party involved, customer, death penalty, violate\\
Software Development \newline (软 件 开 发) & 5.52\% &functionality, software, foundation, capability, advertisement, including, website, effect, model, very, version, stage, solve, search, free, define, memory, non, discover, targeted\\
Film \& Sports \newline (影 视 体 育) & 4.06\% & movie, director, leading actor, portray, team, champion, season, ball, match, point, athlete, league, player, participate, competition, direct, World Cup, win, act, Chinese language\\
Enterprise Management \newline (企 业 管 理) & 4.94\% & enterprise, capability, employee, plan, including, register, supply chain, foundation, supervision, wisdom, solve, financing, activity, red ball, guarantee, intellectual property, high-quality, launch, Chloe (a name), stage\\
Creative Arts \newline (创 意 艺 术) & 4.17\% & music, creation, author, novel, program, song, this book, poem, artist, melody, one piece, band, lyrics, piano, literature and art, a book, copywriting, provident fund, graphics card, hard drive\\
Body Care \newline (身 体 护 理) & 4.13\% & skin, body, effect, human body, clean, absorb, traditional Chinese medicine, very, complexion, sleep, repair, synopsis, online novel, lose weight, weight, efficacy, massage, function, moisturize, serialized\\
Urban Attractions \newline (城 市 景 点) & 5.30\% & decorate, appointment, located in, tourist, area, square, subway, discover, weather, activity, cultural relic, square meter, landscape, forest, public transportation, archaeology, ancient, underground, subway line, gate\\
Insurance Products \newline(保险产品) & 2.25\% & apple, insurance, Wuhan, guarantee, buy, beef, insurance company, fodder, lottery, milk powder, unearthed, critical illness insurance, premium, Zhao (a surname), fruit, claims settlement, sum insured, Qualcomm, Wuhan City, chattel\\
Fashion Coordination \newline (时 尚 搭 配) & 9.24\% & very, buy, like, match, color, good, feel, super, good-looking, match, clothes, effect, appearance, exquisite, high-end, color, hahaha, summer, clean, texture\\
Automotive Technology \newline (汽 车 科 技) & 2.46\% & car model, driving, engine, fitness, Xi'an, new car, vehicle, charging, body, car manufacturer, rescue, recycling, store, version, driver, ideal, Toyota, mass production, torque, firefighting\\
Investment Analysis \newline (投 资 分 析) & 3.43\% & fund, estimate, return, stock, billion US dollars, approximately, point, financing, valuation, interest rate, currency, market capitalization, analyst, same period, rebound, in, futures, buy, account, holding\rule[-10pt]{0pt}{0pt}\\
\hline
\end{longtable}
\end{CJK}

\subsection{Top-Level Domains}

\begin{figure}[ht]
    \centering
    \includegraphics[width=0.9\linewidth]{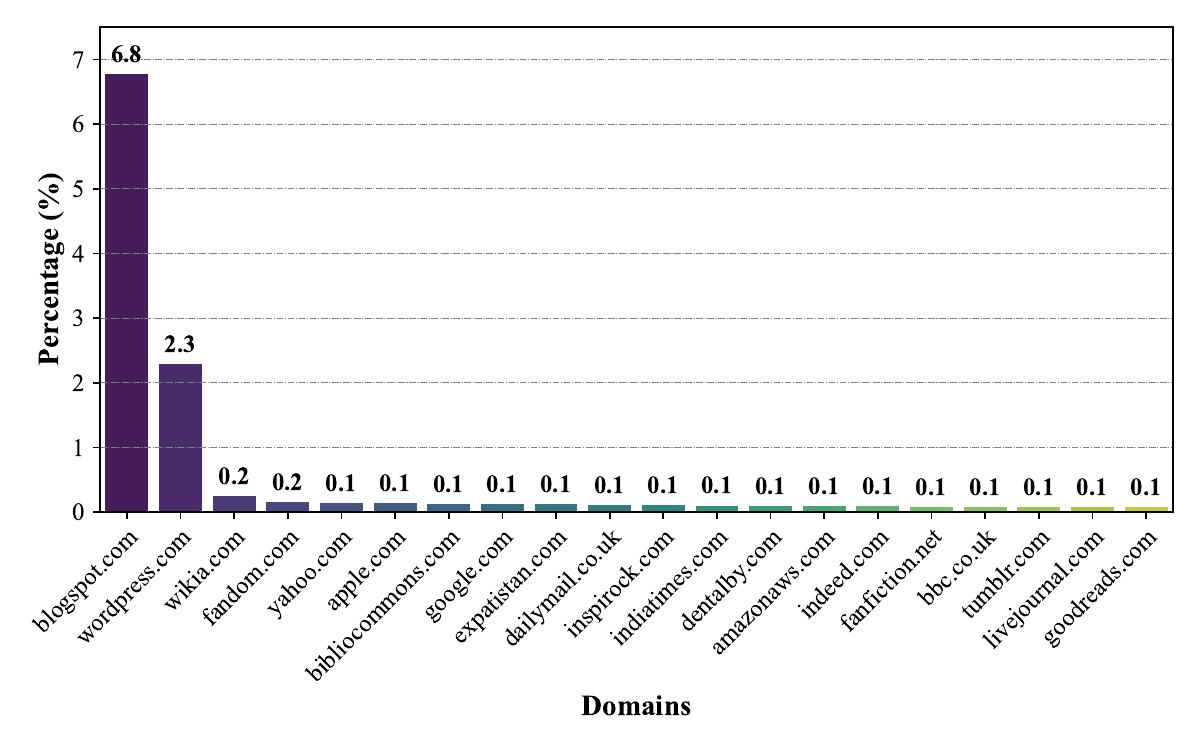}
    \caption{Top-20 Most Frequent Domains for Documents}
    \label{fig:document_domains}
\end{figure}
\begin{figure}[ht]
    \centering
    \includegraphics[width=0.9\linewidth]{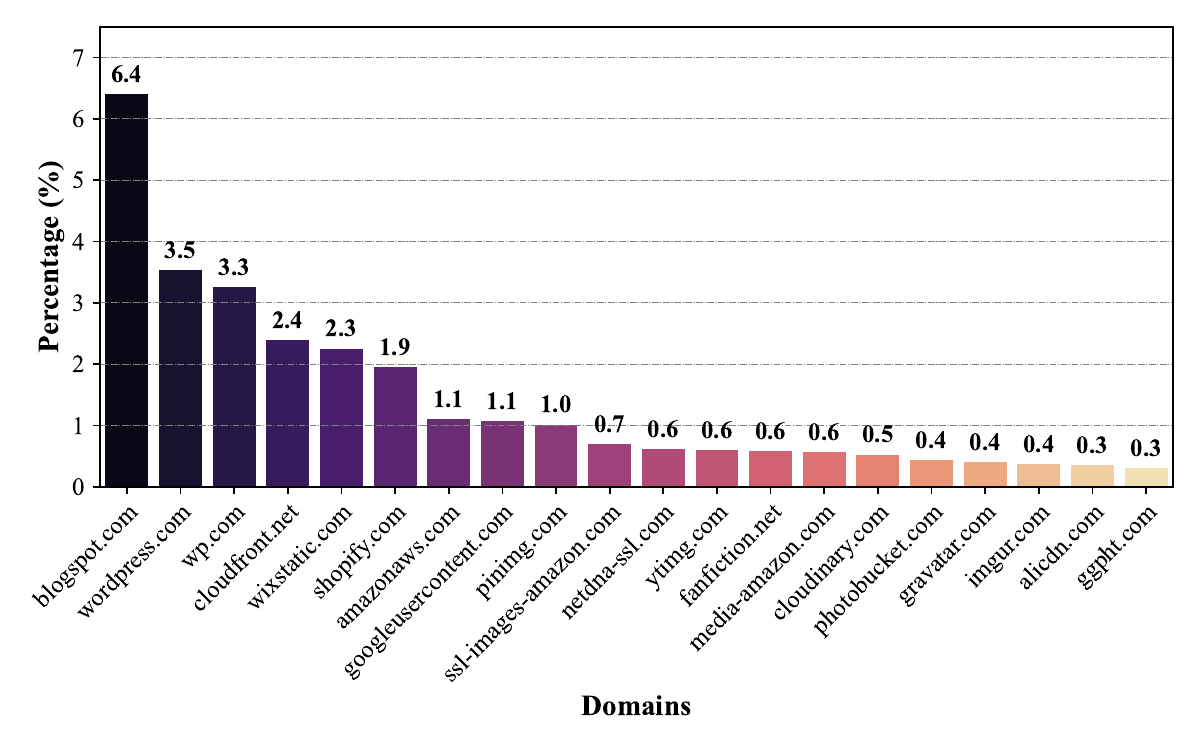}
    \caption{Top-20 Most Frequent Domains for Images}
    \label{fig:image_domains}
\end{figure}

We conduct an analysis of the top-level domains (TLDs) for the \subsetnamecc dataset. The documents are distributed across 16M domains. On average, each domain contains approximately 137 documents, with a median value of 4. As shown in Figure~\ref{fig:document_domains}, the largest sources of documents are blogging platforms, accounting for nearly 9\% of the total documents. Additionally, online encyclopedia platforms (e.g., Wikia), academic publication sites (e.g., BioRxiv), news media (e.g., Daily Mail and BBC), and e-commerce platforms (e.g., Amazon and Apple) are also prominent sources. 

Images are distributed across 14 million domains, with each domain hosting an average of 615 images and a median of 6. Figure \ref{fig:image_domains} shows that image sources are concentrated on a few major platforms, with Blogspot and WordPress accounting for over 10\% of the total images. Cloud storage and content delivery networks (e.g., CloudFront and GoogleUserContent), shopping sites (e.g., Shopify and Amazon), and image hosting platforms (e.g., Flickr and Imgur) also hold significant shares. This high concentration indicates that users prefer using a few efficient platforms for uploading and sharing images, with cloud storage and content delivery networks playing a crucial role in image hosting.

The OmniCorpus dataset shows that document sources are diverse, covering many fields and platforms, while image sources are concentrated, dominated by a few platforms. Blogging platforms are key for both, indicating their importance in user-generated content. The presence of online encyclopedias and academic sites underscores knowledge sharing, and the dominance of cloud storage highlights reliance on efficient services.

\section{License and Author Statement}

We release the dataset under a CC-BY license and Terms of Use that require disclosure of when the dataset is used for the purpose of training models. This license is not intended to replace the licenses of the source content, and any use of the content included in the dataset must comply with the original licenses and applicable rights of its data subjects.

The purpose of this statement is to clarify the responsibilities and liabilities associated with the use of this dataset. While we have made every effort to ensure the accuracy and legality of the data contained within this dataset, we cannot guarantee its absolute completeness or correctness.

Therefore, if any rights, legal or otherwise, are violated through this dataset, including but not limited to copyright infringement, privacy violations, or misuse of sensitive information, we, the authors, assume no liability for such violations.

By utilizing this dataset, you agree that any consequences, legal or otherwise, arising from using this dataset will be the user's sole responsibility. You acknowledge that you will exercise due diligence and adhere to all applicable laws, regulations, and ethical guidelines when using the dataset.

By accessing, downloading, or using this dataset, you signify your acceptance of this statement and your commitment to abide by the terms and conditions of the CC-BY license.

If you disagree with the terms of this statement or the CC-BY license, you are not authorized to use this dataset.

The dataset will be hosted and maintained on GitHub and Hugging Face Hub.

\end{document}